\documentclass[final,3p,times]{elsarticle}

\usepackage{amsmath}  %
\usepackage{amssymb}  %
\usepackage{bm}       %
\usepackage{graphicx} %
\usepackage{hyperref} %
\usepackage{siunitx}  %
\usepackage{float}    %
\usepackage{caption}  %
\usepackage{subcaption} %
\usepackage{enumitem} %
\usepackage{booktabs} %
\usepackage{verbatim} %
\usepackage{multimedia}
\usepackage{xcolor}
\usepackage{amsthm} 
\usepackage{algorithm}
\usepackage{algorithmic}
\usepackage{multirow}
\usepackage{stmaryrd}
\usepackage[size=tiny]{todonotes}

\usepackage{pifont} %

\DeclareMathOperator{\Law}{Law}

\DeclareMathOperator*{\argmin}{arg\,min}

\renewcommand{\hat}{\widehat}
\renewcommand{\bar}{\overline}

\newcommand{\normal}{\mathsf{N}}

\newcommand{\sA}{\mathsf{A}}

\newcommand{\fF}{\mathfrak{F}}

\newcommand{\fC}{\mathfrak{C}}
\newcommand{\fFNN}{\mathfrak{F}^{\textrm {NN}}}
\newcommand{\fFLN}{\mathfrak{F}^{\textrm {LN}}}

\newcommand{\sS}{\mathsf{S}}
\newcommand{\sC}{\mathsf{C}}
\newcommand{\sF}{\mathsf{F}}
\newcommand{\sFNN}{\mathsf{F}^{\textrm {NN}}}
\newcommand{\sFLN}{\mathsf{F}^{\textrm {LN}}}

\newcommand{\sFST}{\mathsf{F}^{\textrm {ST}}}
\newcommand{\sFDec}{\mathsf{F}^{\textrm {Dec}}}
\newcommand{\sFEnc}{\mathsf{F}^{\textrm {Enc}}}
\newcommand{\sFCat}{\mathsf{F}^{\textrm {Cat}}}
\newcommand{\sSST}{\mathsf{S}^{\textrm {ST}}}
\newcommand{\sCST}{\mathsf{C}^{\textrm {ST}}}
\newcommand{\sAST}{\mathsf{A}^{\textrm {ST}}}

\newcommand{\fB}{\mathfrak{B}}
\newcommand{\fA}{\mathfrak{A}}

\newcommand{\fBNO}{\fB_\textrm{MNM}}

\newcommand{\sP}{\mathsf{P}}
\newcommand{\sQ}{\mathsf{Q}}
\newcommand{\sB}{\mathsf{B}}
\newcommand{\sBNO}{\sB_\textrm{MNM}}

\renewcommand{\epsilon}{\varepsilon}

\renewcommand{\hat}{\widehat}

\newcommand{\mP}{{\mathcal P}}

\newcommand{\vd}{v^{\dagger}}
\newcommand{\yd}{y^{\dagger}}

\newcommand{\cP}{\mathcal{P}}
\newcommand{\bbR}{\mathbb{R}}
\newcommand{\bbP}{\mathbb{P}}

\newcommand{\bbN}{\mathbb{N}}
\newcommand{\bbE}{\mathbb{E}}
\newcommand{\bbZ}{\mathbb{Z}}

\newcommand{\placeholder}{\mathord{\color{black!33}\bullet}}%

\definecolor{light}{rgb}{0.5, 0.5, 0.5}

\definecolor{mypink1}{rgb}{0.858, 0.188, 0.478}

\definecolor{darkred}{rgb}{0.5,0,0}
\definecolor{darkgreen}{rgb}{0,0.75,0}
\definecolor{darkblue}{rgb}{0,0,.5}

\newtheorem{definition}{Definition}
\newtheorem{remark}{Remark}
\newtheorem{proposition}{Proposition}
\newtheorem{lemma}{Lemma}
\newtheorem{dataassumption}{Data Assumption}

\newcommand{\hv}{\hat{v}}

\newcommand{\hw}{\hat{w}}
\newcommand{\hy}{\hat{y}}
\newcommand{\hz}{\hat{z}}
\newcommand{\hvn}{\hat{v}^{(n)}}
\newcommand{\hun}{\hat{u}^{(n)}}
\newcommand{\hwn}{\hat{w}^{(n)}}
\newcommand{\hyn}{\hat{y}^{(n)}}
\newcommand{\hzn}{\hat{z}^{(n)}}

\newcommand{\vn}{v^{(n)}}

\newcommand{\Kt}{K_\theta}
\newcommand{\Lt}{L_\theta}

\newcommand{\dst}{d_\text{ST}}
\newcommand{\hg}{\hat{g}}

\newcommand{\bfv}{\mathbf{v}}

\numberwithin{equation}{section}

\journal{Journal of Computational Physics}

\begin{document}

\begin{frontmatter}

\title{Learning Enhanced Ensemble Filters}

\author[uormet,uormath,nceo]{Eviatar Bach} %
\author[utor,vector]{Ricardo Baptista} %
\author[caltech]{Edoardo Calvello} %
\author[caltech]{Bohan Chen%
   \footnote[1]{Corresponding author. Email: bhchen@caltech.edu}}%
\author[caltech]{Andrew Stuart} %

\affiliation[uormet]{organization={Department of Meteorology, University of Reading},
            addressline={ Brian Hoskins Building},
            city={Reading},
            postcode={RG6 6ET},
            country={UK}}

\affiliation[uormath]{organization={Department of Mathematics and Statistics, University of Reading},
            addressline={ Pepper Lane},
            city={Reading},
            postcode={RG6 6AX},
            country={UK}}

\affiliation[nceo]{organization={National Centre for Earth Observation},
            addressline={ Brian Hoskins Building, University of Reading},
            city={Reading},
            postcode={RG6 6ET},
            country={UK}}

\affiliation[utor]{organization={Department of Statistical Sciences, University of Toronto},
            addressline={Ontario Power Building, 700 University Ave.},
            city={Toronto},
            postcode={M5G 1Z5},
            state={ON},
            country={Canada}}

\affiliation[vector]{organization={Vector Institute for Artificial Intelligence},
            addressline={661 University Ave.},
            city={Toronto},
            postcode={M5G 1M1},
            state={ON},
            country={Canada}}
            
\affiliation[caltech]{organization={The Computing + Mathematical Sciences Department, California Institute of Technology},
addressline={1200 E California Blvd},
city={Pasadena},
postcode={91125},
state={CA},
country={USA}}

\begin{abstract}

The filtering distribution in hidden Markov models evolves according to the
law of a mean-field model in state--observation space.
The ensemble Kalman filter (EnKF) approximates this mean-field model with an ensemble of interacting particles, employing a Gaussian ansatz for the joint distribution of the state and observation at each observation time.
These methods are robust, but the Gaussian ansatz limits accuracy. Here this shortcoming is addressed by using machine learning to map the joint predicted state and observation to the updated state estimate. The derivation of methods from a mean field formulation of the true filtering distribution suggests a single parametrization of the algorithm that can be deployed at different ensemble sizes. And we use a mean field formulation of the ensemble Kalman filter as an inductive bias for our architecture.

To develop this perspective, in which the mean-field limit of the algorithm
and finite interacting ensemble particle approximations share a common set of parameters,
a novel form of neural operator is introduced, taking probability distributions as input: a \emph{measure neural mapping} (MNM). A MNM is used to design a novel approach to filtering, the \emph{MNM-enhanced ensemble filter} (MNMEF), which is defined in both the mean-field limit and for interacting ensemble particle approximations. The ensemble approach uses empirical measures as input to the MNM and is implemented using the set transformer,  which is invariant to ensemble permutation and allows for different ensemble sizes. In practice fine-tuning of a small number of  parameters, for specific ensemble sizes, further enhances the accuracy of the scheme. The promise of the approach is demonstrated by its superior root-mean-square-error performance relative to leading methods in filtering the Lorenz ‘96 and Kuramoto–Sivashinsky models.
\end{abstract}

\begin{highlights}
\item We present a framework for learning ensemble filtering algorithms, based on a mean-field state-space formulation of the evolution of the filtering distribution.

\item We introduce a novel generalization of neural operators to maps defined to act between metric spaces of probability measures. We call such maps \textit{measure neural mappings} (MNM). We identify an architecture to implement such maps, based on the attention mechanism.

\item  The MNM architecture is used to define approximations of the mean-field formulation of the filtering distribution. This leads to a novel learning-based framework for ensemble methods: the \emph{measure neural mapping enhanced ensemble filter} (MNMEF). 

\item Because of the mean-field underpinnings, the basic learned parameters are shared between filters with different ensemble sizes. As a consequence, parameterizations learned at one ensemble size can be deployed at different ensemble sizes, leading to
efficient training at small ensemble sizes. To enhance this approach, we introduce a fine-tuning methodology to incorporate a small number of parameters, such as localization and inflation, which intrinsically depend on ensemble size, into the resulting filters.

\item We demonstrate that the resulting filtering algorithm outperforms an optimized local ensemble transform Kalman filter (LETKF, a leading ensemble filter) on Lorenz '96, Kuramoto--Sivashinsky, and Lorenz '63 models, for both small and larger ensemble sizes.
\end{highlights}


\end{frontmatter}

\section{Introduction}
\label{sec:INT}

Filtering, determining the conditional distribution of the state of a partially and noisily observed dynamical system given data collected sequentially in time, constitutes a longstanding computational challenge, particularly when the state is high-dimensional. Sequential data assimilation encompasses a broad range of methods aimed at finding the filtering distribution, or simply estimating the state, in an online fashion. The focus of this work is on the use of sequential data assimilation methods for filtering and state estimation, and in particular the methodologies developed primarily in the geophysical sciences, and weather prediction especially, which are relevant in high dimensions. The overarching goal is to introduce a new framework for the \emph{machine learning} of data assimilation (DA) algorithms, and specifically ensemble filters, which facilitates the sharing of parameters between implementations with different ensemble sizes; we may, for example, train and deploy the same model with different ensemble sizes. To achieve this we work in the context of mean-field state-space formulations of filtering. This perspective enables us to consider the idea of learning algorithms which share a common set of parameters, regardless of ensemble size, by viewing large ensembles as approximating a common mean field limit. In practice, small ensemble sizes behave differently from large ensemble sizes, an issue that has heretofore been addressed by use of inflation and localization. Here we show that an efficient fine-tuning strategy can be used to transfer
parameters trained at one ensemble size to  another ensemble size, at which the algorithm can be deployed. The resulting algorithms are based on stochastic dynamical systems for a state variable whose evolution depends on its own law; particle approximation leads to implementable algorithms. We show how the attention mechanism, adapted from the transformer methodology at the heart of large language models, can be used to define probability measure-dependent transport maps, i.e. mappings that move probability mass from one distribution to another, that can be approximated by empirical measures using arbitrary ensemble sizes. We refer to the resulting class of neural network architectures as \textit{measure neural mappings}. 

In Subsection \ref{ssec:LR}, we set our work in the context of pre-existing literature. We introduce notation used throughout the paper in Subsection \ref{ssec:NO}. Then, we describe the specific contributions we make to achieve our overarching goal, and we overview the contents of the paper in Subsection \ref{ssec:CO}.

\subsection{Literature Review}
\label{ssec:LR}

Data assimilation (DA) is a fundamental framework for integrating observational data with numerical models for the purpose of state estimation and forecasting. DA methods systematically combine observations with prior model forecasts, accounting for their respective uncertainties, to produce optimal estimates of the system state or to characterize a probability distribution over the state. Several textbooks provide comprehensive overviews of this field \cite{jazwinski1970stochastic, law2015data, asch2016data, reich2015probabilistic, evensen2022data, bach2024inverse}.

The Kalman filter \cite{kalman1960new} is a foundational sequential data assimilation algorithm for linear systems with Gaussian errors. The extended Kalman filter, using linearization, adapts the Kalman methodology to nonlinear systems, but its direct application to high-dimensional geophysical problems is computationally prohibitive. The ensemble Kalman filter (EnKF) \cite{evensen2003ensemble, burgers1998analysis, anderson2001ensemble} addresses this limitation through a Monte Carlo approach, representing statistics via a limited ensemble of model states. Square-root variants, including the ensemble transform Kalman filter and the ensemble adjustment Kalman filter \cite{anderson2001ensemble,tippett_ensemble_2003}, reduce both storage requirements and sampling errors through deterministic formulations.  Stochastic EnKF variants update each ensemble member using the Kalman filter equation with random perturbations in the innovation; \cite{van2020consistent} provides a consistent derivation that favors perturbing the modeled observations.
However, EnKF-based methods commonly encounter challenges in practical implementations and, in particular, suffer
from sampling errors caused by finite ensemble sizes. To address these limitations, covariance inflation \cite{anderson1999monte, anderson2007adaptive} and localization \cite{hunt2007efficient} techniques have been developed. Approaches such as the iterative EnKF \cite{sakov2012iterative} and the maximum likelihood ensemble filter \cite{zupanski2005maximum} further improve performance in nonlinear regimes.

The integration of machine learning methodologies into DA has recently emerged as a powerful strategy to enhance predictive accuracy and scalability beyond traditional approaches constrained by Gaussian assumptions. Such machine learning approaches are reviewed in \cite{chengMachineLearningData2023,bach2024inverse}. A key direction has been the direct learning of filters from data. Learning a fixed-gain filter was considered in \cite{hoang_simple_1994,mallia-parfitt_assessing_2016,levine_framework_2022,luk_learning_2024}. McCabe et al.~\cite{mccabe_learning_2021} developed an ensemble filter that uses the mean and covariance matrix as input to a recurrent neural network. Bocquet et al.~\cite{bocquet_accurate_2024} further demonstrated the efficacy of neural network-based filtering schemes without relying on an ensemble, achieving comparable accuracy to ensemble methods by identifying key dynamical perturbations directly from forecast states.

Complementary to these direct learning strategies, a distinct class of approaches formulates ensemble filtering as a transport problem. These methods explicitly construct transformations that map forecast distributions onto analysis distributions. Representative techniques include Knothe--Rosenblatt rearrangements~\cite{spantini2022coupling}, normalizing flows~\cite{chipilski2025exact}, bridge matching~\cite{shi2022conditional}, and optimal transport frameworks~\cite{al2023nonlinear}. These methods offer rigorous probabilistic interpretations and enable accurate sampling from the filtering distribution at each assimilation step. Typically these approaches find transformations acting on each individual particle, rather than operating directly at the distributional level, and are not designed to directly transform or update the ensemble members collectively.

Beyond per-particle transformations, a complementary line of research updates the ensemble collectively via particle flows \citep{van2019particle}, with applications to atmospheric models \cite{hu2024implementation}. Early particle-flow methods cast the analysis as a continuous ordinary differential equation that drives particles toward the posterior \citep{daum2011particle}, with related continuous-time ensemble formulations such as the EnKF--Bucy perspective \citep{bergemann2012ensemble}. Variational or Stein-based mappings are considered to push forward the prior through learned maps, e.g., the variational mapping particle filter \citep{pulido2018kernel}. 

Probabilistic formulations provide another promising direction for learning-based filters. These approaches aim to directly approximate the Bayesian inference problem of obtaining the filtering distribution \cite{brocker_probabilistic_2009,boudier_data_2023,luk_learning_2024,bach2024inverse}. In particular, recent works have proposed frameworks to jointly learn the forecast and analysis steps~\cite{boudier_data_2023}, or assume knowledge of the dynamics and learn parameterized analysis maps via variational inference~\cite{luk_learning_2024}.

Building on the use of deep learning architectures for ensemble filtering, recent efforts have incorporated permutation-invariant neural networks to respect the unordered nature of ensembles. Such architectures enable models to learn interactions among ensemble members without being sensitive to their ordering, which is essential for consistent filter behavior across runs. These designs have been applied to directly learn ensemble update rules~\cite{zhou_bi-eqno_2024} and to refine ensemble forecasts in post-processing settings~\cite{hohlein_postprocessing_2024}.

Transformers and the underlying attention mechanism \cite{vaswani2017attention} are now ubiquitous in machine learning. Their success at modeling global, long-range correlations has made this architecture an attractive methodology for operator learning; see for example \cite{cao2021choose, li2023transformer, calvello2024continuum, wang2025cvit}. In this article we are interested in the attention mechanism as a map between metric spaces of probability measures, thus going beyond the definition on Euclidean spaces from \cite{vaswani2017attention} and the generalization to Banach spaces in \cite{calvello2024continuum}. Recent developments in the mathematical analysis of transformers have led to the formulation
of the continuum limit of self-attention layers \cite{geshkovski2024mathematical}. In \cite{geshkovski2024mathematical} the authors describe the evolution under this continuum limit architecture as the solution of a continuity equation; see \cite{castin2025unified} for an overview on this perspective. This formulation is also employed in \cite{geshkovski2024measure} to analyze self-attention dynamics as a measure-to-measure map. However, this analysis is restricted to measures defined on the sphere and does not include cross-attention. In this article we present a general methodological framework for neural network architectures on the metric space of probability measures and a general definition of attention as a measure-to-measure map. We will leverage this formulation to build implementable methodologies effecting data assimilation.

\subsection{Notation}
\label{ssec:NO}

Throughout the paper we use $\bbN=\{1,2,\ldots\}$, $\bbZ^+=\{0,1,2,\ldots\}$, $\bbR=(-\infty, \infty)$, and $\bbR^+=[0,\infty)$ to denote the set of positive integers, non-negative integers, real numbers, and non-negative real numbers, respectively. We define $[N]=\{1,2,\ldots,N\}$. We use $\mathcal{F}(A)$ to denote the set of all nonempty finite subsets of a set $A$.

We denote by $\cP(\Omega)$ the space of probability measures defined on set $\Omega$.
Given a sigma-algebra on $\Omega$, we define a probability measure $\bbP$ on $\Omega$
and we let $\bbE$ denote expectation under this probability measure. We denote by $\Law(v)$ the law of a random variable $v$, and denote by $T_\#\pi$ the pushforward measure of $\pi$ by the map $T$: if $u \sim \pi$ then $T(u) \sim T_\#\pi.$ For $\tau\in\bbR^d$, we let $\delta_\tau$ denote the Dirac mass on $\bbR^d$ centered at point $\tau$. We use the notation $\normal (m,C)$ for the Gaussian distribution with mean $m$ and covariance $C$. We use the font $\mathsf{mathsf}$ for operators acting on the space of probability measures or the space of functions; we also use this convention for operations on the space of vector-valued sequences over $[N]$, denoted by $\mathcal{U}([N];\bbR^{d})$. If an operator $\sB$ on probability measures is defined as the pushforward by a map, we denote the associated transport map with the font $\mathfrak{mathfrak}$, i.e. $\sB(\cdot) = \fB_\sharp (\cdot)$. 

In the remainder of this paper we assume that all probability measures are absolutely continuous with respect to the Lebesgue measure so that they have probability density functions, or that they comprise a convex combination of Dirac masses. We will refer to probability measures and probability density functions interchangeably, depending on the setting. 

In the context of DA, we use $v$ to denote the system states and $y$ to denote the observations. The superscript $\dagger$ indicates true values and the subscript $j\in\bbZ^+$ denotes the time step. For ensemble methods, we use the parenthesized superscript $(n)$ for the ensemble index. 

\subsection{Contributions and Overview}
\label{ssec:CO}

Our five primary contributions are as follows: 

\begin{enumerate}

\item We present a framework for learning ensemble filtering algorithms, based on a mean-field state-space formulation of the evolution of the filtering distribution.

\item We introduce a novel generalization of neural operators to maps defined to act between metric spaces of probability measures. We call such maps \textit{measure neural mappings} (MNM). We identify an architecture to implement such maps, based on the attention mechanism.

\item  The MNM architecture is used to define approximations of the mean-field formulation of the filtering distribution. This leads to a novel learning-based framework for ensemble methods: the \emph{measure neural mapping enhanced ensemble filter} (MNMEF). 

\item Because of the mean-field underpinnings, the basic learned parameters are shared between filters with different ensemble sizes. As a consequence, parameterizations learned at one ensemble size can be deployed at different ensemble sizes, leading to
efficient training at small ensemble sizes. To enhance this approach, we introduce a fine-tuning methodology to incorporate a small number of parameters, such as localization and inflation, which intrinsically depend on ensemble size, into the resulting filters.

\item We demonstrate that the resulting filtering algorithm outperforms an optimized local ensemble transform Kalman filter (LETKF, a leading ensemble filter) on Lorenz '96, Kuramoto--Sivashinsky, and Lorenz '63 models, for both small and larger ensemble sizes.

\end{enumerate}

In Section \ref{sec:MFF} we describe filtering from the perspective of mean-field dynamics and define a learning framework to approximate the true filter, addressing Contribution 1. Section \ref{sec:AFM} is devoted to the introduction of \textit{measure neural mappings}, a generalization of neural operators to maps taking probability measures as inputs; we show in this section that the attention mechanism can be used to build a transformer architecture defined on probability measures, thereby tackling Contribution 2. Contributions 3 and 4 are addressed in Section~\ref{sec:LIE}, where we detail the algorithm for finite particle sizes. The numerical experiments in Section \ref{sec:NUM_EXP} support Contribution 5. We conclude in Section \ref{sec:CON}. We also note that a reader focused on methodological contributions alone can skip the self-contained Section~\ref{sec:AFM}.

\section{Learning Mean-field Filters}
\label{sec:MFF}

In this section, we formulate the filtering problem (Subsection~\ref{ssec:fsu}), describing sequential DA from the perspective of nonlinear evolution of probability densities (Subsection~\ref{ssec:PE}) and from the mean-field state-space evolution perspective (Subsection~\ref{ssec:SE}). We then present the general idea of learning an analysis map which has a probability measure as an input (Subsection~\ref{ssec:prob_map}). Finally, we introduce our learning-based mean-field filter approach, which enhances traditional ensemble Kalman filtering by incorporating trainable correction terms in the mean-field equations (Subsection~\ref{ssec:enkf_ins_map}). This allows us to extend and, as will show in numerical results presented in the penultimate section, improve upon the standard EnKF methodology, which relies on a Gaussian ansatz; in so-doing we maintain the interpretable structure of Kalman filtering.

\subsection{Filtering Set-Up}
\label{ssec:fsu}

Consider the stochastic dynamics model given by
\begin{subequations}\label{eq:dynamic}
\begin{align}
    v_{j+1}^\dag &= \Psi(v_j^\dag) + \xi_j^\dag, \quad j \in \bbZ^+,  \\
    v_0^\dag &\sim \normal(m_0, C_0), \quad \xi_j^\dag \sim \normal(0, \Sigma) \, \text{i.i.d.},
\end{align}
\end{subequations}
where $\Psi:\bbR^{d_v}\to\bbR^{d_v}$ is the forward dynamic model, and $v^\dag_j,\xi_j^\dag\in \bbR^{d_v}$ for $j\in \bbZ^+$. We assume that the sequence $\{\xi_j^\dag\}_{j \in \bbZ^+}$ is independent of initial condition $v_0^\dag$; this is often written as $\{\xi_j^\dag\}_{j \in \bbZ^+} \perp v_0^\dag$. The data model is given by
\begin{subequations}\label{eq:observation}
\begin{align}
    y_{j+1}^\dag &= h(v_{j+1}^\dag) + \eta_{j+1}^\dag, \quad j \in \bbZ^+, \\
    \eta_j^\dag &\sim \normal(0, \Gamma) \, \text{i.i.d.},
\end{align}
\end{subequations}
where $h:\bbR^{d_v}\to\bbR^{d_y}$ is the observation operator and $y^\dag_j,\eta_j^\dag\in \bbR^{d_y}$ for $j\in \bbZ^+$. We assume that the sequence $\{\eta_j^\dag\}_{j \in \bbN} \perp v_0^\dag$ and that the two sequences in the dynamics and data models are independent of one another: $\{\xi_j^\dag\}_{j \in \bbZ^+} \perp \{\eta_j^\dag\}_{j \in \bbN}$. The states $v_j^\dag$ lie in $\mathbb{R}^{d_v}$, while the observations $y_j^\dag$ lie in $\mathbb{R}^{d_y}$. 
Note that there is a hidden parameter $\Delta t$ specifying the time interval between steps $j$ and $j+1$. It is implicitly embedded in \eqref{eq:dynamic} through $\Psi$, which represents the discrete-time map obtained by integrating the continuous dynamics over an interval of length $\Delta t$. 

\begin{remark} We use $\dagger$ for all variables defining the state and observation, and the noises that define them. This is to distinguish them from other similar
variables, appearing without $\dagger$, in the algorithms that follow. In this paper, we adopt autonomous dynamics with additive Gaussian noises according to \eqref{eq:dynamic} and \eqref{eq:observation}. There are several extended settings, briefly outlined in Remark~\ref{rem:extensions}. These extensions are important in applications but introduce substantial technical challenges. We do not consider those settings in this paper for clarity of exposition. \end{remark} 

\begin{remark}[Possible extensions beyond the baseline model]
\label{rem:extensions}
Beyond the autonomous, additive-Gaussian setting in~\eqref{eq:dynamic}--\eqref{eq:observation}, one may consider \cite{bain2009fundamentals, stuart2010inverse, reich2015probabilistic}:
\begin{itemize}
\item \textbf{Time-dependent dynamics/observations.} Allow $\Psi$, $\Sigma$, $h$, and $\Gamma$ to vary with $j$. This can destroy stationarity and leads to non-homogeneous transition/likelihood kernels.
\item \textbf{Multiplicative or non-Gaussian noises.} Replace additive Gaussian errors by state-dependent or heavy-tailed noises. This can induce heteroscedastic innovations and non-quadratic likelihoods.
\item \textbf{Infinite-dimensional state spaces.} Model $v_j$ in a function space (e.g., a separable Hilbert space) for PDE-governed systems. One must address well-posedness (e.g., trace-class noise/covariances), operator-valued updates, and consistency under spatial discretization/mesh refinement.
\end{itemize}
\end{remark}

There are two primary types of problems in data assimilation: the filtering problem and the smoothing problem. In this paper, we focus on the filtering problem. To this end we define the accumulated data up to time $j$ by $Y_j^\dagger := \{y_1^\dagger, \dots, y_j^\dagger\}.$ Using this notation we state the filtering problem in Definition~\ref{def:filtering_problem}, and the state estimation problem that stems from it in Definition~\ref{def:state_estimation}.

\begin{definition}[The Filtering Problem]\label{def:filtering_problem}
The filtering problem refers to identifying and sequentially updating the probability densities $\pi_j(v_j^\dagger) := \bbP(v_j^\dagger | Y_j^\dagger)$ in $\cP(\mathbb{R}^{d_v})$ for $j \in [J]$. These densities $\pi_j$ 
are referred to as the filtering distributions at time $j$.
\end{definition}

\begin{definition}[State Estimation]\label{def:state_estimation}
The state estimation problem refers to estimating the state ${v_j^\dagger}$, based on the given observations ${Y_j^\dagger}$, and to updating the estimate sequentially over a series of time indices $j \in [J]$. 
\end{definition}

State estimation can be viewed as a subproblem of the filtering problem. Given the filtering distribution $\pi_j$, a natural state estimate is the conditional expectation $\mathbb{E}[v_j^\dagger | Y_j^\dagger]$ as this delivers the minimum mean-squared error estimate of $v_j^\dagger$.
In this paper we explicitly focus on \emph{state estimation}: our objective is to recover the hidden state $v_j^\dagger$ at each time step rather than to approximate the full filtering distribution $\pi_j$. Accordingly, our loss functions are defined with respect to the true state. We note that, in high-dimensional settings, the conditional mean is generally inaccessible in practice—accurate computation would require particle filters with prohibitively large ensembles—so it is standard to evaluate and tune filters against the true state. As outlined in Section~\ref{sec:CON}, our framework can be extended to the filtering problem by adopting distribution-matching losses to learn the filtering distribution (and hence its mean).

\subsection{Filtering: Probability Evolution}
\label{ssec:PE}

We consider the filtering problem, which consists of identifying the filtering distribution given by
\begin{equation}
    \pi_j(v_j^\dagger) := \bbP(v_j^\dagger | Y_j^\dagger).
\end{equation}
We define the following two measures
\begin{subequations}
\begin{align}
    &\hat{\pi}_{j+1}(v_{j+1}^\dagger) = \bbP(v_{j+1}^\dagger|Y_j^\dagger),\label{eq:measure_hpi}\\
    &\rho_{j+1}(v_{j+1}^\dagger, y_{j+1}^\dagger) = \bbP(v_{j+1}^\dagger, y_{j+1}^\dagger|Y_j^\dagger).\label{eq:measure_rho}
\end{align}
\end{subequations}
The update $\pi_j\rightarrow \pi_{j+1}$ can be written in the following three steps:
\begin{subequations}
\begin{align}
    &\text{Predict (linear):} &\quad \hat{\pi}_{j+1} &= \sP \pi_j, \label{eq:pred}\\
    &\text{Extend to data/state space (linear):} &\quad \rho_{j+1} &= \sQ \hat{\pi}_{j+1}, \label{eq:extend}\\
    &\text{Analysis (nonlinear):} &\quad \pi_{j+1} &= \sB(\rho_{j+1}; y_{j+1}^\dagger). \label{eq:analysis}
\end{align}
\end{subequations}
$\sP$ is a linear operator that transforms the filtering distribution at time $j$ to the forecast distribution at time $j+1$, and is given for the stochastic dynamics model in~\eqref{eq:dynamic} by:
\begin{equation}\label{eq:operator_P}
    \sP \pi(v) = \frac{1}{\sqrt{(2\pi)^{d_v} \det \Sigma}} 
    \int \exp\left(-\frac{1}{2} \|v - \Psi(u)\|_{\Sigma}^2 \right) \pi(u) \, du.
\end{equation}
$\sQ$ is also a linear operator, one that multiplies the forecast distribution by the observation likelihood, and is given by:
\begin{equation}\label{eq:operator_Q}
    \sQ \pi(v,y) = \frac{1}{\sqrt{(2\pi)^{d_y} \det \Gamma}} 
\exp\left(-\frac{1}{2} \|y - h(v)\|_\Gamma^2 \right) \pi(v).
\end{equation}
Then, the nonlinear operator $\sB$ can be written in the form:
\begin{equation}\label{eq:operator_B}
    \sB(\rho_{j+1};\yd_{j+1})(v) = \frac{\rho_{j+1}(v,\yd_{j+1})}{\int_{\bbR^{d_v}} \rho_{j+1}(u, y_{j+1}^\dagger) \, du}.
\end{equation}

\subsection{Filtering: State-Space Evolution}
\label{ssec:SE}

Directly evolving the filtering distribution $\pi_j$ is generally intractable since the nonlinear operator $\sB$ \eqref{eq:operator_B} cannot be computed explicitly for most practical applications. Alternatively, we consider a random state evolution process governed by its own law, given by
\begin{subequations}
\label{eq:meta_state}
\begin{align}
    &\text{Predict:} &\quad \hat{v}_{j+1} &= \Psi(v_j) + \xi_j, \quad\xi_j\sim\normal (0,\Sigma),\label{eq:state_pred}\\
    &\text{Extend to observation:} &\quad \hy_{j+1} &= h(\hv_{j+1}) + \eta_{j+1},\quad \eta_{j+1}\sim\normal (0,\Gamma),\label{eq:state_extend}\\
    &\text{Analysis:} &v_{j+1} &= \fB(\hv_{j+1},\hy_{j+1};\yd_{j+1},\rho_{j+1}), \label{eq:state_analysis}
\end{align}
\end{subequations}
where $\rho_{j+1}$ is the joint measure defined by \eqref{eq:measure_rho}. From the definitions of $\sP$ \eqref{eq:operator_P} and $\sQ$ \eqref{eq:operator_Q}, and from equations \eqref{eq:state_pred} and \eqref{eq:state_extend}, it follows that $(\hv_{j+1},\hy_{j+1})$ is distributed as $\rho_{j+1}$, i.e., $\rho_{j+1} = \Law\bigl((\hv_{j+1},\hy_{j+1})\bigr)$. Now, using the theory of transport, we choose $\fB$ so that the pushforward of $\rho_{j+1}$
under this map delivers the desired conditioning on the observed data $\yd_{j+1}:$
$$\sB(\rho_{j+1};\yd_{j+1}) = \fB(\,\placeholder\,,\,\placeholder\,,;\yd_{j+1},\rho_{j+1})_\sharp (\rho_{j+1}).$$
It follows that if $\pi_{j} = \Law(v_{j})$ then $\pi_{j+1} = \Law(v_{j+1})$.

The preceding construct is mathematically appealing but it leaves open the question of how to identify an appropriate choice of $\fB.$ The mean-field ensemble Kalman filter \cite{calvello2022ensemble} leaves (\ref{eq:meta_state}a) and (\ref{eq:meta_state}b) intact but replaces the mapping $\fB$ in (\ref{eq:meta_state}c) by the affine (in $(\hv_{j+1},\hy_{j+1})$) approximation
\begin{equation}\label{eq:mf_kf0}
v_{j+1} = \fB_\mathrm{EnKF}(\hv_{j+1},\hy_{j+1};\yd_{j+1},\rho_{j+1}),\\
\end{equation}
defined by
\begin{subequations}\label{eq:mf_kf}
\begin{align}
    v_{j+1} &= \hv_{j+1} + K_{j+1}\left(\yd_{j+1} - \hy_{j+1}\right),\\
    K_{j+1} &= \hat{C}^{vh}_{j+1}\left(\hat{C}^{hh}_{j+1} + \Gamma\right)^{-1}. \label{eq:EnKF_gain}
\end{align}
\end{subequations}
Here, $K_{j+1}$ is called the Kalman gain, $\hat{C}^{vh}$ is the covariance between $\hat{v}_{j+1}$ and $h(\hat{v}_{j+1})$, and $\hat{C}^{hh}_{j+1}$ is the covariance for $h(\hat{v}_{j+1})$. Both $\hat{C}^{vh}$ and $\hat{C}^{hh}_{j+1}$ are computed under $\hat{\pi}_{j+1}$:
\begin{subequations}
\label{eq:cov_enkf0}
    \begin{align}
        \hat{C}_{j+1}^{vh} &= \bbE^{\hv\sim\hat{\pi}_{j+1}}\left[\left(\hv - \bar{v}_{j+1} \right) \otimes \left(h(\hv) - \bar{h}_{j+1} \right)\right],\\
        \hat{C}_{j+1}^{hh} &= \bbE^{\hv\sim\hat{\pi}_{j+1}}\left[\left(h(\hv) - \bar{h}_{j+1} \right) \otimes \left(h(\hv) - \bar{h}_{j+1} \right)\right].
    \end{align}
\end{subequations}
Here, $\bar{v}_{j+1}$ and $\bar{h}_{j+1}$ are the mean of states and observations respectively, under $\hat{\pi}_{j+1}$:
\begin{equation}
\label{eq:sample_means0}
\bar{v}_{j+1} = \bbE^{\hv\sim\hat{\pi}_{j+1}}\hv, 
\quad \bar{h}_{j+1} = \bbE^{\hv\sim\hat{\pi}_{j+1}} h(\hv).
\end{equation}
We note that in computation of these covariances
we have simply used expectations under $\hat{\pi}_{j+1}$, the marginal of $\rho_{j+1}$ on the state. It is possible to use the replacement
$$\hat{C}^{yy}_{j+1}=\hat{C}^{hh}_{j+1} + \Gamma,$$
where $\hat{C}^{yy}_{j+1}$ is the covariance for $\hat{y}_{j+1}.$
Hence we write the map $\fB_\mathrm{EnKF}$ as dependent on $\rho_{j+1}$, which covers both cases.

What is sometimes termed the stochastic ensemble Kalman filter is implemented in practice by employing an interacting particle system approximation of the mean-field model. The methodology uses the empirical measure defined by the set of $N$ particles $\{v_j^{(\ell)}\}_{\ell=1}^N$ to approximate the filtering distribution $\pi_j = \Law(v_j)$, and hence the empirical measure defined by $\left\{\hat{v}_{j+1}^{(\ell)} \right\}_{\ell=1}^N$ and $\left\{\left(\hat{v}_{j+1}^{(\ell)}, \hat{y}_{j+1}^{(\ell)}\right)\right\}_{\ell=1}^N$, to approximate 
$\hat{\pi}_{j+1}$ and $\rho_{j+1}$, respectively. 
We write the resulting approximations as
\begin{equation}
    \hat{\pi}_{j+1}^{(N)}:=\frac{1}{N}\sum_{n=1}^N \delta_{\hat{v}_{j+1}^{(n)}}, \quad 
    \rho_{j+1}^{(N)}:=\frac{1}{N}\sum_{n=1}^N \delta_{(\hat{v}_{j+1}^{(n)}, \hat{y}_{j+1}^{(n)})}.
\end{equation}
Making this particle approximation in \eqref{eq:meta_state}, and replacing
$\fB$ by $\fB_\mathrm{EnKF}$ leads to the interacting particle system
\begin{subequations}
\begin{align}
    \hvn_{j+1} &= \Psi(\vn_j) + \xi^{(n)}_j,\\
    \hyn_{j+1} &= h(\hvn_{j+1}) + \eta^{(n)}_{j+1},\\
    \vn_{j+1} &= \fB_\mathrm{EnKF}\bigl(\hvn_{j+1},\hyn_{j+1};\yd_{j+1},
    \rho_{j+1}^{(N)}\bigr).
   \end{align}
\end{subequations}
This is a particular instance of the ensemble Kalman filter (EnKF), where use of the empirical approximation $\rho_{j+1}^{(N)}$ means
that the covariances $\hat{C}^{vh}_{j+1}$ and $\hat{C}^{hh}_{j+1}$ given by \eqref{eq:cov_enkf0} are now computed by use of empirical approximation $\hat{\pi}_{j+1}^{(N)}$ of $\hat{\pi}_{j+1}$:
\begin{subequations}\label{eq:cov_enkf}
\begin{align}
    \hat{C}_{j+1}^{vh} &=  \bbE^{\hv\sim\hat{\pi}^{(N)}_{j+1}}\left[\left(\hv - \bar{v}_{j+1} \right) \otimes \left(h(\hv) - \bar{h}_{j+1} \right)\right] = \frac{1}{N} \sum_{n=1}^{N} \left(\hvn_{j+1} - \bar{v}_{j+1} \right) \otimes \left(h(\hvn_{j+1}) - \bar{h}_{j+1} \right), \label{eq:cov_vh_enkf} \\
    \hat{C}_{j+1}^{hh} &=  \bbE^{\hv\sim\hat{\pi}^{(N)}_{j+1}}\left[\left(h(\hv) - \bar{h}_{j+1} \right) \otimes \left(h(\hv) - \bar{h}_{j+1} \right)\right] = \frac{1}{N} \sum_{n=1}^{N} \left(h(\hvn_{j+1}) - \bar{h}_{j+1} \right) \otimes \left(h(\hvn_{j+1}) - \bar{h}_{j+1} \right). \label{eq:cov_hh_enkf}
\end{align}
\end{subequations}
Here $\bar{v}_{j+1}$ and $\bar{h}_{j+1}$, originally defined in \eqref{eq:sample_means0}, are also computed by use of an empirical approximation of $\hat{\pi}_{j+1}$:
\begin{subequations}
\label{eq:sample_means}
\begin{align}
\bar{v}_{j+1} = \bbE^{\hv\sim\hat{\pi}^{(N)}_{j+1}}\hv = \frac{1}{N} \sum_{n=1}^{N} \hvn_{j+1}, \qquad
\bar{h}_{j+1} = \bbE^{\hv\sim\hat{\pi}^{(N)}_{j+1}}h(\hv) = \frac{1}{N} \sum_{n=1}^{N} h(\hvn_{j+1}).
\end{align}
\end{subequations}

\begin{remark}
The ensemble Kalman filter exhibits permutation invariance with respect to the ensemble members. This important property stems from the fact that both the sample means $\bar{v}_{j+1}$, $\bar{h}_{j+1}$ in \eqref{eq:sample_means} and the empirical covariance matrices $\hat{C}_{j+1}^{vh}$, $\hat{C}_{j+1}^{hh}$ in \eqref{eq:cov_enkf}  are calculated as averages over all ensemble members, making them invariant to the ordering of particles in the ensemble. This permutation invariance is a crucial property that motivates the design of our methodology.
\end{remark}

\subsection{Learning Probability Measure--Dependent Analysis Maps}
\label{ssec:prob_map}
Machine learning--based approaches can be applied to approximate the analysis step \eqref{eq:analysis}. 
Indeed, one may seek an operator $\sBNO$ (\emph{measure neural mappings, or MNMs, will be introduced in Section \ref{sec:AFM}) acting on probability measures so that}
\begin{equation}
\label{eq:BNO_measures}
\sBNO(\rho_{j+1}; y_{j+1}^\dagger)\approx\sB(\rho_{j+1}; y_{j+1}^\dagger).
\end{equation}
This aim motivates a novel and significant extension of neural operators \cite{kovachki_neural_2023,kovachki2024operator}: designing operator architectures that act on the space of probability measures. We refer to these neural operators that act on probability measures as \emph{measure neural mappings} (MNM). A natural way to construct such an approximation is by neural operator $\fBNO$, acting on the joint space of state and data and parameterized by the observation and by the law of the state, so that
\begin{subequations}
\begin{align}
    &\fBNO(\hv_{j+1},\hy_{j+1};\yd_{j+1},\rho_{j+1}) \approx \fB(\hv_{j+1},\hy_{j+1};\yd_{j+1},\rho_{j+1}) ,\label{eq:BNO_transport0}\\
    &\sBNO(\rho_{j+1}; y_{j+1}^\dagger) = \fBNO(\,\placeholder\,,\,\placeholder\,;\yd_{j+1},\rho_{j+1})_\sharp \rho_{j+1} \approx \sB(\rho_{j+1}; y_{j+1}^\dagger).\label{eq:BNO_transport}
\end{align}
\end{subequations}
The map $\fBNO$ in \eqref{eq:BNO_transport0} acts on the joint state--observation space; in contrast the map $\sBNO$ in \eqref{eq:BNO_transport}, defined through pushforward, acts on the space of probability measures on the state space. The quantities
after the semi-colon parameterize these maps.
To use this neural operator approximation in the state-space evolution,
we replace \eqref{eq:state_analysis} by the approximation
\begin{equation}
v_{j+1} = \fBNO(\hv_{j+1},\hy_{j+1};\yd_{j+1},\rho_{j+1}).
\end{equation}

\subsection{EnKF-Inspired Probability Measure--Dependent Analysis Maps}
\label{ssec:enkf_ins_map}

We now propose a specific novel filtering methodology, which we coin as the \emph{measure neural mapping enhanced ensemble filter} (MNMEF). In this section, we will discuss the mean-field formulation of this approach. We consider a specific form of $\fBNO$ inspired by the success of the EnKF and describe it here in the mean-field limit. We generalize \eqref{eq:mf_kf} to take the form
\begin{subequations}\label{eq:our_mf}
\begin{align}
\begin{split}
    v_{j+1} &= \hat{v}_{j+1} + (\Kt)_{j+1}\left(\yd_{j+1} - \hat{y}_{j+1}\right),
\end{split}\label{eq:our_BNO}\\
    (\Kt)_{j+1} &= \Kt\left(\rho_{j+1},\yd_{j+1}\right),\label{eq:BNO_k_theta}
\end{align}
\end{subequations}
where $K_\theta\left(\rho_{j+1},\yd_{j+1}\right)$ is a parameterized $\mathbb{R}^{d_v\times d_y}$-valued function, and $\theta$ denotes the vector of trainable parameters. 
Our proposed choice for $\fBNO$ \eqref{eq:our_BNO} introduces inductive bias by mimicking the affine EnKF update. The following proposed form for $\Kt(\placeholder,\placeholder)$ is identical at each time step; for simplicity we omit the subscript $j+1$ in its definition and, in particular, use $\hat{\pi}$ as the distribution of the predicted state. We assume the following form for $\Kt$, generalizing the Kalman gain \eqref{eq:EnKF_gain}:
\begin{equation}\label{eq:k-theta}
    \Kt = \Kt^{(1)} \left( \Kt^{(2)} + \Gamma \right)^{-1}.
\end{equation}
Here $\Kt^{(1)}$ and $\Kt^{(2)}$ are modified versions of $\hat{C}^{vh}$ and $\hat{C}^{hh}$ from \eqref{eq:cov_enkf0}. Specifically we have
\begin{subequations}\label{eq:full_K}
\begin{align}
    \Kt^{(1)} &= \bbE^{\hv\sim\hat{\pi}} \left[ \left(\hv - \bar{v} + \hw_\theta\right) \otimes \left(h(\hv) - \bar{h} + \hz_\theta\right)\right], \label{eq:full_K1}\\
    \Kt^{(2)} &= \bbE^{\hv\sim\hat{\pi}} \left[ \left(h(\hv) - \bar{h} + \hz_\theta\right) \otimes \left(h(\hv) - \bar{h} + \hz_\theta\right)\right], \label{eq:full_K2}
\end{align}
\end{subequations}
where $\hw_\theta$ and $\hz_\theta$ are trainable correction terms that depend on the state $\hv$, observation $h(\hv)$, true observation $\yd$, and the joint distribution $\rho_h$.  We apply a neural operator $\fF(\cdot;\theta):\bbR^{d_v}\times \bbR^{d_y}\times \bbR^{d_y} \times \mathcal{P}(\bbR^{d_b}\times \bbR^{d_y}) \rightarrow \bbR^{d_v}\times \bbR^{d_y}$ to define the correction terms as 
\begin{equation}\label{eq:FNO}
    \fF\left(\hv, h(\hv), y^\dagger, \rho_h;\theta\right) = \left(\hw_\theta, \hz_\theta\right),
\end{equation}
where $\rho_h = \Law\bigl((\hat{v},h(\hat{v}))\bigr)$ is the joint distribution of $(\hat{v},h(\hat{v}))$. As well as being defined in the mean-field limit, this form of the model can both be learned and deployed through interacting particle system approximations, just as for the original ensemble Kalman filter. These interacting particle systems, as well as details of the architecture employed for $\fF$ in the particle approximation, are defined in Section~\ref{sec:LIE}.

\begin{remark}[Factoring Our $\Gamma$]
We note that $\Kt$ \eqref{eq:BNO_k_theta} depends on the probability measure $\rho$ while $\fF$ \eqref{eq:FNO} depends on $\rho_h$. The relationship between $\rho$ and $\rho_h$ is given by
\begin{equation}
    (\hat v,\hat y)\sim\rho
\;\Leftrightarrow\;
\begin{cases}
(\hat v,\,h(\hat v))\sim\rho_h,\\
\eta\sim\mathcal N(0,\Gamma),\quad\eta\perp \hat v,\\
\hat y = h(\hat v) + \eta.
\end{cases}
\end{equation}
By factoring out the noise covariance in the term $(\Kt^{(2)} + \Gamma)^{-1}$ in \eqref{eq:k-theta}, we use the explicit knowledge of $\Gamma$ in derivation of $\Kt$. Therefore, $\fF$ uses the input $\rho_h$ which does not depend on $\Gamma.$
\end{remark}

\begin{remark}[Beyond Gaussian Assumptions]
    The mean-field map $\fBNO$ \eqref{eq:our_BNO} is constrained to be affine in both inputs $\hv$ and $\hy$. Indeed, 
    \[
    \fBNO(\hv,\hy;\yd,\rho) = \hat{v} + \Kt\left(\yd - \hat{y}\right).
    \]
    We note that this is the same constraint satisfied by the ensemble Kalman filter; see \cite{calvello2022ensemble}. However, as shown in \cite{calvello2022ensemble}, the gain $K$ in the ensemble Kalman filter is such that the resulting state estimate has law with first and second-order moments matching a Gaussian approximation of the true filter. Our approach is more general. Indeed, since $K_\theta$ is trainable, unlike in the EnKF, the gain is not constrained to satisfy any moment-matching with a Gaussian approximation of the true filter. The ensemble Kalman filter possesses statistical guarantees for problems involving filtering distributions that are close to Gaussian (given, for example, by dynamics and observation operators that are close to linear) \cite{calvello2024accuracy}. Further work will involve investigating whether this more general methodology possesses theoretical guarantees for a broader class of filtering problems than the EnKF.
\end{remark}

The core of our proposed approach is to learn a neural operator $\fF$, as given in \eqref{eq:FNO}, which takes both probability measures, as well as vectors, as inputs.  Section~\ref{sec:AFM} is a self-contained description of a novel methodology for training neural operators that take probability measures as inputs; it leverages the attention mechanism \cite{vaswani2017attention}.
In Section~\ref{sec:LIE} we will build on this novel
neural operator framework to provide a concrete methodology for enhanced data assimilation when actioned using interacting particle approximations of the mean-field limit.

\section{Learning Maps on the Space of Probability Measures}
\label{sec:AFM}

In this section we introduce a general framework to define neural network architectures acting on the space of probability measures. Let $\Omega_1, \Omega_2$ denote two sets,
equipped with sigma algebras so that we may assign probabilities, and consider maps of the form $\sF:\mP(\Omega_1)\times\Theta\to \mP(\Omega_2)$; here $\Theta \subseteq \bbR^p.$
Our goal is to define the parametric class of functions so that, by choice of $\theta^* \in \Theta$, we may approximate a given map 
$\sF^\dagger: \mP(\Omega_1)\to \mP(\Omega_2)$
by $\sF(\cdot;\theta^*):  \mP(\Omega_1)\to \mP(\Omega_2).$
As the desired size of the approximation error shrinks we will typically require
$\Theta$, and in particular the number of parameters $p$, to grow. Such methodology
is now well-established for maps between Banach spaces; see the review \cite{kovachki2024operator}. Defining such maps between metric spaces in general,
and probability spaces in particular, is an unexplored research question which we
address here. We refer to neural networks on the space of probability measures as \textit{measure neural mappings} (MNM).

In what follows, we show that the popular transformer architecture may be formulated as a mean-field mapping on the space of probability measures; this mapping is mean-field because it not only acts on but is also parametrized by probability measures. We call such a neural network architecture the \textit{transformer measure neural mapping} (TMNM). This architecture will serve as a parametric family of operators of the form $\sF:\mP(\bbR^{d_u})\times \mP(\bbR^{d_w})\times\Theta\to\mP(\bbR^{d_v})$ defined via action on $\mu \in \mP(\bbR^{d_u})$ and $\nu\in \mP(\bbR^{d_w})$ as 
\begin{equation}
\label{eq:general}
    \sF(\mu,\nu; \theta) = \mathfrak{F}(\placeholder\,;\nu, \theta)_\sharp \mu.
\end{equation}
The transport map in \eqref{eq:general} is a mapping of the form $\mathfrak{F}: \bbR^{d_u}\times \mP(\bbR^{d_w}) \times \Theta \to \bbR^{d_v}$. For ease of exposition we will henceforth omit the $\theta$ parameterizations of the operators.
In Section \ref{sec:LIE} we will show how this transformer measure neural mapping, in particular the transport map that defines it, may be used to implement \eqref{eq:FNO} in the context of learning an ensemble-based data assimilation
algorithm. For this section, however, we work quite generally, without specific reference to data assimilation.

\begin{remark}[Training a Transformer MNM]
    \label{rm:MNM_training}
    In practice, implementing MNM architectures such as that in \eqref{eq:general} involves working with samples from the measures $\mu \in \mP(\bbR^{d_u})$ and $\nu \in\mP(\bbR^{d_w})$ rather than the measures themselves. This finite sample approximation introduces the challenge of selecting an appropriate loss function  that is readily computed from samples. For the purposes of this article, we will employ a loss function based on matching the mean of the output measure under the map. Future work will optimize based on the use of metrics on probability measures and scoring rules, not just the mean; see Section \ref{sec:LIE} for more details. 
    
    In what follows we assume access to samples $u(i)\sim\mu$ for $i=1,\ldots,N$ and similarly $w(j)\sim\nu$ for $j=1,\ldots,M$. We note the following useful identity,
    which follows from \eqref{eq:general} (dropping explicit $\theta-$dependence):
    \begin{equation}
    \label{eq:equivalence}
    \sF\Bigl(\frac{1}{N}\sum_{i=1}^N\delta_{u(i)},\nu\Bigr) = \frac1N\sum_{i=1}^N\delta_{\mathfrak{F}\bigl(u(i);\nu\bigr)}.
    \end{equation}
    This characterization will be used to describe implementable algorithms.
\end{remark}
In view of Remark \ref{rm:MNM_training} we proceed by describing the TMNM architecture in detail as a mapping between spaces of probability measures. This mean-field perspective is more general and justifies the deployment of MNM architectures trained with a finite collection of samples to ensembles of different sizes. Nonetheless, in the development that follows, we provide examples illustrating the finite sample empirical setting such as in Remark \ref{rm:MNM_training}, which will be the focus of later sections.

We proceed in Subsection \ref{ssec:TMNM_block} by describing the TMNM architecture as a composition of specific maps on probability measures, which we call attention blocks and which involve the application of the attention map. In Subsection \ref{ssec:attention_measures} we define 
attention as a map on measures and show how, via application of this map to empirical measures, it is possible to recover the formulation of attention on Euclidean spaces from \cite{vaswani2017attention}.

\subsection{Transformer Measure Neural Mapping}
\label{ssec:TMNM_block}

The TMNM architecture as an operator of the form $\sF:\mP(\bbR^{d_u})\times \mP(\bbR^{d_w})\to\mP(\bbR^{d_u})$ consists of a composition of $N_e \in \bbN$ number of operators $\sS_w: \mP(\bbR^{d_w}) \to \mP(\bbR^{d_w})$, $N_d \in \bbN$ number of operators $\sS_u: \mP(\bbR^{d_u}) \to \mP(\bbR^{d_u})$, and $\sC:  \mP(\bbR^{d_u})\times \mP(\bbR^{d_w}) \to \mP(\bbR^{d_u})$, which we call the self-attention and cross-attention blocks, respectively. We define $\sF$ via action on inputs $\mu \in \mP(\bbR^{d_u})$ and $\nu\in \mP(\bbR^{d_w})$ as 
\begin{equation}
\label{eq:composition}
    \sF(\mu,\nu) = \sS_{u,N_d+N_e}(\placeholder)\circ\cdots\circ\sS_{u,1+N_e}(\placeholder)\circ \sC(\mu,\placeholder) \circ \sS_{w,N_e}(\placeholder)\circ\cdots \circ\sS_{w,2}(\placeholder)\circ\sS_{w,1}(\nu).
\end{equation}
For ease of exposition, for operators $\sS$ we will henceforth drop the notation $\sS_{\placeholder}$, as the dimension of the input state space will be clear from context. We note that the cross-attention block is intimately related to the pooling multihead attention block from the set transformer architecture \citep{lee2019set}. In Remark \ref{rm:ST} and in Section \ref{sec:LIE} we make explicit the connection between the TMNM and the set transformer. Both operators $\sS$ and $\sC$ will be defined via a pushforward operation under a transport map $\fC$: we define
\begin{subequations}
    \begin{align}
    \label{eq:Smeasure}
        \sS(\mu) &= \fC(\placeholder; \mu)_\sharp\mu,\\
    \label{eq:Cmeasure}
        \sC(\mu,\nu) &= \fC(\placeholder; \nu)_\sharp\mu.
    \end{align}
\end{subequations}
We now outline the construction of the transport map appearing in \eqref{eq:Smeasure} and \eqref{eq:Cmeasure}. Indeed, we may write the map $\fC:\bbR^{d_u}\times\mP(\bbR^{d_w})\to \bbR^{d_u}$ acting on the inputs $u\in\bbR^{d_u}$ and $\pi\in\mP(\bbR^{d_w})$ as the composition of the maps
\begin{subequations}
\label{eq:self_block_transport}
\begin{align}
    u &\mapsfrom \fFLN\bigl(u +\fA(u;\pi)\bigr),\\
    u &\mapsfrom \fFLN\bigl(u + \fFNN(u)\bigr).
\end{align}
\end{subequations}
We note that each of the maps appearing in the composition in \eqref{eq:self_block_transport} may be used to define a pushforward operator on measures. 
\begin{remark}[Set Transformer Architecture]
\label{rm:ST}
    We note that the composition in \eqref{eq:composition} is a generalization of the set transformer architecture from \cite{lee2019set} to the continuum. Indeed, the set transformer acts on a collection of vectors $\{u(1),\ldots,u(N)\}\subset\bbR^{d_u}$ which we may assume to be drawn from an underlying reference measure $\mu$. The set transformer relies on learning a vector $w\in \bbR^{d_w}$ with which cross-attention is applied to the ensemble. The set transformer architecture may thus be viewed as the map $\sF$ in \eqref{eq:composition} when acting on the empirical measure $\mu^N= \frac1N\sum_{i=1}^N\delta_{u(i)}$ and a learned dirac $\delta_w$.
\end{remark}

We will now turn our focus to defining each of the maps in \eqref{eq:self_block_transport}. The map $\fFLN:\bbR^{d_u}\to \bbR^{d_u}$ defines layer normalization and is such that
\begin{equation}
\label{eq:LN}
\Bigl(\fFLN({u};\gamma,\beta)\Bigr)_k = \gamma_k\cdot\frac{u_k - m(u)}{\sqrt{\sigma^2(u)+\epsilon}} +\beta_k,
\end{equation}
for $k=1,\dots,d_u$, any $u \in \bbR^{d_u}$, where the subscript notation $(\placeholder)_k$ is used to denote the $k$'th entry of the vector. In equation \eqref{eq:LN}, $\epsilon\in\bbR^+$ is a fixed parameter, $\gamma_k,\beta_k\in\bbR$ are learnable parameters and $m,\sigma$ are defined as
\begin{equation}
\label{eq:LN_meanvar}
\begin{aligned}
    m(u) = \frac1{d_u}\sum_{k=1}^{d_u}u_k,\qquad
    \sigma^2(u) = \frac1{d_u}\sum_{k=1}^{d_u}\bigl(u_k-m(u) \bigr)^2 ,
\end{aligned}
\end{equation}
for any $u\in \bbR^{d_u}.$ The operator $\fFNN: \bbR^{d_u} \to \bbR^{d_u}$ is defined such that
\begin{equation}
\label{eq:NN}
    \fFNN(u;W_1,W_2,b_1,b_2) = W_2f\bigl(W_1u+b_1\bigr) + b_2,
\end{equation}
for any $u \in \bbR^{d_u}$, where $W_1,W_2\in\bbR^{d_u\times d_u}$ and $b_1,b_2 \in \bbR^{d_u}$ are learnable parameters and where $f$ is a nonlinear activation function. 

The map $\fA:\bbR^{d_u}\times\mP(\bbR^{d_w})\to \bbR^{d_u}$ is the attention map: Subsection \ref{ssec:attention_measures} which follows is dedicated to its formulation as a transport defining the attention map on measures $\sA$.

\subsection{Attention as a Map on Measures}
\label{ssec:attention_measures}

We present the definition of the attention operator as a map between a product of the spaces of probability measures $\mP(\bbR^{d_u})$ and $\mP(\bbR^{d_w})$ into another space of probability measures $\mP(\bbR^{d_v})$ \footnote{In this self-contained subsection we use the notation $d_v$ to denote the dimension associated with the output space of probability measures. This is not to be confused with the dimension of the state of the dynamical system \eqref{eq:dynamic} that is subject of the majority of the remainder of the paper.}. If considering the map as acting on measures defined on the unbounded domain $\bbR^{d_w}$ we must make assumptions on the input measures themselves, i.e., tail decay assumptions, so that the attention operator is well-defined. Hence, for ease of exposition we first present attention as an operator $\sA: \mP(\Omega_u)\times \mP(\Omega_w)\to \mP(\bbR^{d_v})$, where $\Omega_u,\Omega_w$ are bounded open subsets of $\bbR^{d_u}, \bbR^{d_w}$ respectively; we will then extend the definition to the general case $\mathsf{A}:\mP(\mathbb{R}^{d_u})\times \mP_{\delta}(\mathbb{R}^{d_w})\to \mP(\mathbb{R}^{d_v}) $, where the subset of measures $\mP_{\delta}(\mathbb{R}^{d_w})\subset \mP(\mathbb{R}^{d_w})$ consisting of all probability measures whose tails decay strictly faster than exponential, will be appropriately defined.

\begin{definition}[Attention on Measures: Bounded State Space]
We define the operator $\sA: \mP(\Omega_u)\times \mP(\Omega_w)\to \mP(\bbR^{d_v})$ via its action on inputs $(\mu,\nu) \in \mP(\Omega_u)\times\mP(\Omega_w)$ as a pushforward of the first input measure $\mu\in\mP(\Omega_u)$ by a transport map $\fA:\Omega_u\times\mP(\Omega_w) \to \bbR^{d_v} $ which is parametrized by the second input measure $\nu \in \mP(\Omega_w)$, hence
\begin{equation}
\label{eq:attn_measures}
    \sA(\mu,\nu) = \fA(\cdot;\nu)_\sharp \mu.
\end{equation}
The transport map $\mathfrak{A}:\Omega_u\times \mP(\Omega_w)\to \mathbb{R}^{d_v}$ takes as input a vector $u\in \Omega_u$ and a probability measure $\nu\in \mP(\Omega_w)$ and computes an expectation under a rescaling of the measure $\nu$, which is parametrized by $u$. Namely for $(u,\nu)\in\Omega_u\times\mP(\Omega_w) $ it holds that
\begin{equation}
\label{eq:transport}
    \mathfrak{A}(u;\nu) = \mathbb{E}_{w\sim p(\cdot;u,\nu)}Vw,
\end{equation}
where
\begin{equation}
\label{eq:prob_measure}
    p(dw;u,\nu) = \frac{\exp\bigl(\langle Qu,Kw\rangle\bigr)\nu(dw)}{\int_{\Omega_w}\exp\bigl(\langle Qu,Kz\rangle\bigr)\nu(dz)}.
\end{equation}
\end{definition}
Dimensions of the learnable matrices $Q,K,V$ are determined from the above, provided
$Q,K$ have the same output dimension, which is a free parameter to be specified.

\begin{remark}[$p(\placeholder;u)$ is a probability measure]
\label{rm:measure_bounded}
    Clearly, the normalization constant in \eqref{eq:prob_measure} is finite, as the integral is taken over $\Omega_w$, a bounded domain. Furthermore $p$ integrates to $1$ over $\Omega_w$, hence it defines a probability measure.
\end{remark}
In view of Remark \ref{rm:measure_bounded}, in order to extend the definition of the attention operator expressed in \eqref{eq:attn_measures} to a map on probability measures defined on an unbounded domain, extra care is required to ensure the normalization constant in \eqref{eq:prob_measure} is finite. Indeed, we must assume that the measure $\nu \in \mP(\bbR^{d_w})$ has sufficiently fast tail decay in order to compensate for the exponential growth in the integrand. To this end, we introduce the following subset of $\mP(\bbR^{d_w})$, on which the attention operator will be shown to be well-defined. Intuitively, the condition that defines the subset is satisfied by all measures whose tails decay faster than exponentially.

\begin{definition}[Set of Measures with Exponential Tail Decay]
\label{def:tail_decay}
    For $\delta > 0$ we define $\mathcal{P}_{\delta}(\mathbb{R}^{d_u})\subset \mathcal{P}(\mathbb{R}^{d_u})$ as a set of measures satisfying the following exponential tail decay condition: 
\begin{equation*}
\label{eq:tail}
    \mathcal{P}_{\delta} (\mathbb{R}^{d_u}) = \left\{\mu\in\mathcal{P}(\mathbb{R}^{d_u}):\,\, 
    \mu\{\omega\in\Omega:|X(\omega)|\geq t\}\leq 2\exp\bigl(-t^{1+\delta}\bigr) \right\}.
\end{equation*}
\end{definition}

Given Definition \ref{def:tail_decay}, we may now define the attention operator as a map on probability measures defined over unbounded state spaces.
\begin{definition}[Attention on Measures: Unbounded State Space] \label{def:wnc}
We define the operator $\sA: \mP(\bbR^{d_u})\times \mP_{\delta}(\mathbb{R}^{d_w})\to \mP(\bbR^{d_v})$ via its action on inputs $(\mu,\nu) \in \mP(\bbR^{d_u})\times\mP_{\delta}(\mathbb{R}^{d_w})$ as a pushforward of the first input measure $\mu\in\mP(\bbR^{d_u})$ by a transport map $\fA:\bbR^{d_u}\times\mP_{\delta}(\mathbb{R}^{d_w}) \to \bbR^{d_v} $ which is parametrized by the second input measure $\nu \in \mP_{\delta}(\mathbb{R}^{d_w})$, hence
\begin{equation}
    \sA(\mu,\nu) = \fA(\cdot;\nu)_\sharp \mu.
\end{equation}
The transport map $\mathfrak{A}:\bbR^{d_u}\times \mP_{\delta}(\mathbb{R}^{d_w})\to \mathbb{R}^{d_v}$ takes as input a vector $u\in \bbR^{d_u}$ and a probability measure $\nu\in \mP_{\delta}(\mathbb{R}^{d_w})$ and involves computation of an expectation under a rescaling of the measure $\nu$ which is parametrized by $u$. Namely for $(u,\nu)\in\bbR^{d_u}\times\mP_{\delta}(\mathbb{R}^{d_w}) $ it holds that
\begin{equation}
\label{eq:transport2}
    \mathfrak{A}(u;\nu) = \mathbb{E}_{w\sim p(\cdot;u,\nu)}Vw,
\end{equation}
where
\begin{equation}
\label{eq:prob_measure2}
    p(dw;u,\nu) = \frac{\exp\bigl(\langle Qu,Kw\rangle\bigr)\nu(dw)}{\int_{\bbR^{d_w}}\exp\bigl(\langle Qu,Kz\rangle\bigr)\nu(dz)}.
\end{equation}
\end{definition}

Definition \ref{def:tail_decay} allows to state and prove the following lemma regarding the finiteness of the normalization constant appearing in \eqref{eq:prob_measure2}.
\begin{lemma}
    Let $Q, K, V$ be fixed matrices of appropriate dimensions. The measure $p(du;s)$ is a well-defined probability measure for $\nu\in\mathcal{P}_{\delta}(\mathbb{R}^{d_u})$. The measure $p(dw;u,\nu)$ is a well-defined probability measure for $\nu\in\mathcal{P}_{\delta}(\mathbb{R}^{d_w})$.
\end{lemma}
\begin{proof}
    It suffices to show $\int_{\bbR^{d_w}} \exp\bigl(\langle Qu,Kz\rangle\bigr)\nu(dz)<\infty$. Using the Cauchy--Schwarz and Young's inequalities we obtain that,
\begin{equation}
    \exp\bigl(\langle Qu,Kz\rangle\bigr) \leq \exp\bigl(\|Q\|\|K\||u||z|\bigr) \leq \exp\left(\frac{(\|Q\|\|K\||u|)^{\frac{1+\delta}{\delta}}}{(1+\delta)/\delta}\right)\exp\left(\frac{|z|^{{1+\delta}}}{(1+\delta)}\right).
\end{equation}
    We note that 
    \begin{subequations}
        \begin{align*}
            \int_{\bbR^{d_w}} \exp\bigl(\langle Qu,Kz\rangle\bigr)\nu(dz) & \leq C(u)\int_{\bbR^{d_w}} \exp\left(\frac{|z|^{{1+\delta}}}{(1+\delta)}\right) \nu(dz)\\
            &=C(u)\int_0^\infty \nu\left(\exp\left(\frac{|z|^{{1+\delta}}}{(1+\delta)}\right) >t \right)dt\\
            &= C(u)\int_0^1 \nu\left(\exp\left(\frac{|z|^{{1+\delta}}}{(1+\delta)}\right) >t \right)dt+C(u)\int_1^\infty \nu\left(\exp\left(\frac{|z|^{{1+\delta}}}{(1+\delta)}\right) >t \right)dt\\
            &\leq C(u) + C(u)\int_1^\infty \nu(|z|>\tau)\exp\left(\frac{\tau^{{1+\delta}}}{(1+\delta)}\right) \tau^\delta d\tau\\
            &\leq C(u)+ C(u)\int_1^\infty \exp\left(\frac{-\delta}{(1+\delta)}\cdot\tau^{{1+\delta}}\right)\tau^\delta d\tau <\infty,
    \end{align*}
    \end{subequations}
    where $C(u)$ is a constant depending on $u$ that changes from line to line, and where in the fourth line we used that the first integral is less than or equal to $1$ and applied the substitution $t = \exp \bigl(\tau^{1+\delta}/(1+\delta) \bigr)$ to the second integral; furthermore, we used $\nu\in \mathcal{P}_{\delta}(\mathbb{R}^{d_w})$ for the final inequality.
\end{proof}

For implementable algorithms it is useful to consider the finite sample case, where we assume we have access to empirical approximations of the measures $\mu\in\mP(\bbR^{d_u})$ and $\nu \in \mP_{\delta}(\bbR^{d_w})$ via a collection of samples, or ensembles, $\{u(1),\ldots,u(N)\}\subset \bbR^{d_u}$ and $\{w(1),\ldots,w(M)\}\subset \bbR^{d_w}$. 
The following proposition provides a description of the attention operator when applied to empirical measures. The resulting expressions will be useful more generally to formulate the trainable algorithms we will use for data assimilation.  

\begin{proposition}[Attention on Empirical Measures]
\label{prop:cross_sequences}
      Assume that, for $\mu\in \mP(\bbR^{d_u}), \nu \in \mP_{\delta}(\bbR^{d_w})$,  samples $u(1),\cdots,u(N)$ are drawn i.i.d. from $\mu$ and samples $w(1),\cdots,w(M)\sim\nu$ are drawn i.i.d. from $\nu.$  Then for $\mu^N=\frac1N\sum_{j=1}^N\delta_{u(j)}$ and $\nu^M=\frac1M\sum_{j=1}^M\delta_{w(j)}$
    it holds that
    \[\sA(\mu^N,\nu^M) = \frac1N\sum_{j=1}^N\delta_{{\fA\bigl(u(j);\,\nu^M\bigr)}},\]
    where 
    \begin{equation}
\label{eq:empirical}
 \fA\bigl(u(j);\nu^M\bigr) =\frac{\sum_{k=1}^M\exp\bigl(\langle Qu(j),Kw(k)\rangle\bigr)Vw(k)}{\sum_{\ell=1}^M\exp\bigl(\langle Qu(j),Kw(\ell)\rangle\bigr)}. 
\end{equation}
\end{proposition}
\begin{proof}
This follows by application of identity \eqref{eq:equivalence} to Definition \ref{def:wnc}.
\end{proof}

We note that the expression \eqref{eq:empirical} corresponds to the definition of attention on discrete sequences as highlighted in \cite{vaswani2017attention}.

\section{Particle Approximation of Learned Mean-field Filters}\label{sec:LIE}
In this section, we detail our proposed machine-learning-based DA method, the \emph{measure neural mapping enhanced ensemble filter (MNMEF)}, for the state estimation problem. In particular we build on the mean-field formulation introduced in Section~\ref{sec:MFF}, and in Subsection \ref{ssec:enkf_ins_map} in particular, by introducing an interacting particle system approximation.
The resulting machine-learned ensemble approach is based on updating 
$\{\vn_j\}_{n=1}^N$ to $\{\vn_{j+1}\}_{n=1}^N$ through the following steps:
\begin{subequations}\label{eq:ourens}
\begin{align}
    &\text{Predict:} & \hvn_{j+1} &= \Psi(\vn_j) + \xi^{(n)}_j, \label{eq:ourens_pred}\\
    &\text{Extend to observation space:} & \hyn_{j+1} &= h(\hvn_{j+1}) + \eta^{(n)}_{j+1}, \label{eq:ourens_extend}\\
    &\text{Analysis:} &\vn_{j+1} &= \hvn_{j+1} + (\Kt)_{j+1} \left( y^\dagger_{j+1} - \hyn_{j+1} \right). \label{eq:ourens_analysis}
\end{align}
\end{subequations}
After the prediction steps \eqref{eq:ourens_pred} and \eqref{eq:ourens_extend}, we obtain a set of particles $\{(\hvn_{j+1},h(\hvn_{j+1}))\}_{n=1}^N$.
The computation of $\Kt$ is elaborated through equations \eqref{eq:k-theta}, \eqref{eq:full_K}, and \eqref{eq:FNO} in a mean-field perspective, now extended
to also include this particle setting. As in Section~\ref{sec:MFF}, we drop time index $j$ in what follows and recall that $\rho_h = \Law\bigl((\hat{v},h(\hat{v}))\bigr)$ is the joint distribution of $(\hat{v},h(\hat{v}))$ in the mean-field limit. In the particle case, under their implied empirical distribution, we have 
\begin{equation}
\label{eq:rhoN}
     \rho_h^{(N)}:=\frac{1}{N}\sum_{n=1}^N \delta_{\bigl(\hat{v}_{j+1}^{(n)}, h(\hat{v}_{j+1}^{(n)})\bigr)}
\end{equation}
and use this as an input measure to $\Kt.$
The core idea is to train a parameterized neural operator $\fF\left(\hv, h(\hv), y^\dagger, \rho_h;\theta\right)$ that outputs the correction terms used to compute $\Kt$. 
This can be done using the transformer measure neural mapping (TMNM) architecture $\sF$ \eqref{eq:composition} introduced in Section \ref{sec:AFM}. When applied to empirical measures such as \eqref{eq:rhoN}, this TMNM scheme is equivalent to the set transformer architecture \cite{lee2019set} acting on sequences, as outlined in Remark \ref{rm:ST}. In this section, which is concerned with implementation, we will work with the formulation of the scheme on sequences; we will also show that the resulting scheme is invariant to permutation of the elements of the sequence. To distinguish the resulting operator on sequences to the one acting on measures from Section \ref{sec:AFM}, we will employ the notation $\sFST$ opposed to $\sF$; see \eqref{eq:st_architecture} for the definition of $\sFST$. Due to the permutation invariance property of the set transformer, we can treat the input as an unordered set: we define the corresponding operator acting on finite sets using the notation $F^\mathrm{ST}$ ; see \eqref{eq:FST}. 

The parameters $\theta$ learned in $\Kt$ implicitly depend on $N$, the number of ensemble members used in training. However, the relation between the set transformer and its mean-field limit from Section \ref{sec:AFM} allows us to deploy the model parameters $\theta$, trained using an ensemble size $N$, to an ensemble of a different sizes $N'$. This is because the $N$ and $N'$ systems may both be thought of as approximating the same mean-field limit. Similarly to the standard use of the EnKF, our learning framework also incorporates techniques such as inflation and localization to enhance the performance of the filter \cite{bach2024inverse} as ensemble sizes vary. We view this as a fine-tuning: fixing the majority of the parameters in $\theta$ as ensemble sizes vary, but allowing a small subset of parameters—including those related to inflation and localization—to vary with ensemble size.

We present the complete implementation details of our method in the following subsections. In Subsection~\ref{ssec:discrete_ST}, we review the set transformer architecture. In Subsection~\ref{ssec:architecture}, we present our architecture that learns a parameterized gain matrix, generalizing the EnKF formulation, and enhanced by use of adaptively learned inflation and localization parameters. Subsection~\ref{ssec:training_parms} explains the loss function and training process. In Subsection~\ref{ssec:fine-tuning}, we describe how to efficiently fine-tune our method to allow for ensemble size dependent inflation and localization. In \ref{app:framework_overview} we provide a comprehensive summary of our learning framework with detailed pseudo-code algorithms.

Note that all aspects are based on a discrete setting, where our approach evolves an ensemble of particles, as in the EnKF. Consequently, all measure-related operations are transformed into computations based on the empirical measure, or equivalently, the ensemble, viewed as a set of points in state space. When the time-step subscript is omitted, this indicates that the analysis applies to any time step $j$.

\subsection{The Set Transformer}\label{ssec:discrete_ST}
In this section, we review the set transformer architecture \citep{lee2019set} used to process ensembles $\{(\hvn)\}_{n=1}^N$ or $\{(h(\hvn))\}_{n=1}^N$ into fixed-dimensional feature vectors. While Section~\ref{sec:AFM} introduced the set transformer from a measure-theoretic perspective, we now consider our input as a sequence that, due to the architecture's design, can be treated either as a set or an empirical measure.

For two sequences $u\in \mathcal{U}([N];\bbR^{d_u})$ and $w\in \mathcal{U}([M];\bbR^{d_w})$, we provide the definition for the sequence-based attention mechanism, which is given by
\begin{equation}\label{eq:seq_fA}
    \fA\bigl(u(j);w\bigr) =\frac{ \sum_{k=1}^M \exp\bigl(\langle Qu(j),Kw(k)\rangle\bigr)Vw(k)}{\sum_{\ell=1}^M\exp\bigl(\langle Qu(j),Kw(\ell)\rangle\bigr)}.
\end{equation}
We recall that this is equivalent to the formulation  \eqref{eq:empirical} in Proposition \ref{prop:cross_sequences} obtained by applying the measure theoretic definition of attention to empirical measures. Defining the set of sequences of finite length in $\bbR^{d}$ by,
\begin{equation}
    \mathcal{U}_F(\bbR^d) = \cup_{N=1}^\infty\mathcal{U}([N];\bbR^{d}),
\end{equation}
we can define an attention operator $\sAST$ (where the superscript $\mathrm{ST}$ indicates it is used in the set transformer architecture) as a sequence-to-sequence operator, analogous to the attention $\sA$ acting on measures \eqref{eq:attn_measures}:
\footnote{As in Section \ref{sec:AFM}, here $d_v$ does not necessarily refer to the state space dimension in the dynamical system; the exposition in this subsection is quite general, defining the set transformer and linking it to the MNM in Section \ref{sec:AFM}.}
\begin{equation}
\label{eq:seq_sA}
\sAST: \mathcal{U}_F(\bbR^{d_u})\times\mathcal{U}_F(\bbR^{d_w})\rightarrow \mathcal{U}_F(\bbR^{d_v}).
\end{equation}
In view of \eqref{eq:seq_fA}, the output sequence length is the same as the length of the first input. Specifically, for $u\in \mathcal{U}([N];\bbR^{d_u})$ and $w\in \mathcal{U}([M];\bbR^{d_w})$,
\begin{equation}
    \sAST(u, w)(j) = \fA\bigl(u(j);w\bigr),\quad \forall j\in [N].
\end{equation}
The learnable parameters in $\sAST$ are denoted by
\begin{equation}\label{eq:att_parameters}
    \theta(\sAST) = \{Q,K,V\}
\end{equation}
In the context of practical implementations, $\sAST$ is commonly instantiated as a multihead attention mechanism, which allows the model to jointly attend to information from different representation subspaces \cite{vaswani2017attention}.
The detailed formulation of multihead attention is presented in \ref{app:multiheadatt}.

The set transformer architecture is composed of several key building blocks \citep{lee2019set}. The fundamental computation unit is the Cross-Attention Block (CAB) given in the following definition.

\begin{definition}[CAB for Sequences]\label{def:CAB}
The Cross-Attention Block (CAB) $\sCST$ is an operator that maps two sequences into one sequence:
\begin{equation}
    \sCST:\quad \mathcal{U}_F(\mathbb{R}^{d_u})\times \mathcal{U}_F(\mathbb{R}^{d_w})\rightarrow \mathcal{U}_F(\mathbb{R}^{d_u}).
\end{equation}
For $u\in \mathcal{U}([N];\bbR^{d_u})$ and $w\in \mathcal{U}([M];\bbR^{d_w})$, $\sCST$ is defined as 
\begin{subequations}\label{eq:CAB}
\begin{align}
\sCST(u,w) &= \sFLN_1(u_1 + \sFNN(u_1))\in \mathcal{U}([N];\mathbb{R}^{d_u}),\label{eq:CAB_1}\\
u_1 &= \sFLN_2\bigl(u +\sAST(u,w)\bigr)\in \mathcal{U}([N];\mathbb{R}^{d_u})\label{eq:CAB_2}.
\end{align}
\end{subequations}
Here $\sAST$ is the sequence-to-sequence attention \eqref{eq:seq_sA}; $\sFLN$
denotes application of the layer normalization operator $\fFLN$ \eqref{eq:LN},
elementwise in the sequence; $\sFNN$ denotes application of the feedforward operator $\fFNN$ \eqref{eq:NN}, also elementwise in the sequence. The $+$ operator acting on sequences is interpreted as adding elements with the same index, i.e.,
\begin{equation}
    (u + v)(j) = u(j) + v(j).
\end{equation}
The trainable parameters in $\sCST$ include the parameters from $\sFLN_1$, $\sFLN_2$, $\sFNN$, and $\sAST$, i.e.,
\begin{equation}\label{eq:para_CAB}
   \theta(\sCST) = \theta(\sFLN_1) \cup \theta(\sFLN_2) \cup \theta(\sFNN) \cup \theta(\sAST),
\end{equation}
where the layer normalization layers $\sFLN_1, \sFLN_2$ have trainable parameters according to \eqref{eq:LN}\eqref{eq:LN_meanvar}, the feedforward layer $\sFNN$ has trainable parameters according to \eqref{eq:NN}, and the attention layer $\sAST$ has trainable parameters according to \eqref{eq:att_parameters}. We note that practically the cross-attention block on sequences is the multi-head attention block from \cite{vaswani2017attention,lee2019set} and may be viewed as the sequence-to-sequence analog of the cross-attention block $\sC$ on measures from \eqref{eq:Cmeasure}.
\end{definition}

The set transformer architecture is composed of several self-attention blocks (SAB) (Definition~\ref{def:SAB}) and a pooling by multi-head attention block (PMA) (Definition~\ref{def:PMA}), which are defined in the following using CAB.

\begin{definition}[SAB]\label{def:SAB}
    The self-attention block (SAB) $\sSST$, also called the set attention block \cite{lee2019set}, maps a sequence to another sequence with the same length and dimension, so that
    \begin{equation}
        \sSST: \quad\mathcal{U}_F(\mathbb{R}^{d_u})\rightarrow \mathcal{U}_F(\mathbb{R}^{d_u}).
    \end{equation}
    The application of $\sSST$ is calculated by setting both inputs of $\sCST$ to be the same, i.e.,
    \begin{equation}\label{eq:SAB}
        \sSST(u) = \sCST(u,u).
    \end{equation}
    Here, the trainable parameters satisfy
    \begin{equation}
        \theta(\sSST) = \theta(\sCST).
    \end{equation}
    See~\eqref{eq:Smeasure} for the measure theoretic analog $\sS$.
\end{definition}

\begin{definition}[PMA]\label{def:PMA}
    The pooling by multi-head attention block (PMA) $\sCST(s,\placeholder)$ is defined by replacing the first input in the CAB $\sCST(\placeholder,\placeholder)$ by a trainable parameter $s$, called the seed. The seed $s\in\mathcal{U}([N_s], \bbR^{d_s})$ is a sequence, where $N_s$ and $d_s$ are fixed hyperparameters. Therefore $\sCST(s,\placeholder)$ maps a sequence of any finite length to a sequence with fixed length,
    \begin{equation}\label{eq:PMA}
        \sCST(s,\placeholder): \quad\mathcal{U}_F(\mathbb{R}^{d_u})\rightarrow \mathcal{U}([N_s];\mathbb{R}^{d_s}).
    \end{equation}
    The trainable parameters in PMA are the parameters in the CAB and the seed $s$,
    \begin{equation}
        \theta\bigl(\sCST(s,\placeholder)\bigr) = \theta(\sCST)\cup\{s\}.
    \end{equation}
    See Remark \ref{rm:ST} for the measure theoretic analog $\sC(s,\placeholder)$.
\end{definition}

The set transformer maps a sequence of any finite length to a fixed-dimensional feature vector:
\begin{equation}\label{eq:st_seq}
    \sFST:\mathcal{U}_F(\mathbb{R}^{d_u})\rightarrow\bbR^{\dst}.
\end{equation}
To fully specify the detailed architecture of $\sFST$, we will utilize two multilayer perceptrons \eqref{eq:NN} $\sFNN_\mathrm{in}:\bbR^{d_u}\rightarrow\bbR^{d_\mathrm{latent}}$, and $\sFNN_\mathrm{out}:\bbR^{N_s\times d_s}\rightarrow\bbR^{\dst}$ (that act elementwise on sequences), and a parameter-free concatenation layer $\sFCat:\mathcal{U}([N_s],\bbR^{d_s})\rightarrow \bbR^{N_s\times d_s}$ that concatenates all elements in a sequence into a vector.

For a sequence $u\in\mathcal{U}([N];\mathbb{R}^{d_u})$, the set transformer is hence given by:
\begin{subequations}\label{eq:st_architecture}
\begin{align}
    \sFST(u) &= \sFDec(\placeholder)\circ\sCST(s,\placeholder)\circ\sFEnc(u),\\
    \sFEnc(u) &=\bigl(\sSST_{N_\mathrm{e}}(\placeholder)\circ\cdots\circ\sSST_{1}(\placeholder)\bigr)\circ\sFNN_\mathrm{in}(u),\\ \sFDec(u)&=\sFNN_\mathrm{out}(\placeholder)\circ\sFCat(\placeholder)\circ\bigl(\sSST_{N_\mathrm{d} + N_\mathrm{e}}(\placeholder)\circ\cdots\circ\sSST_{1+ N_\mathrm{e}}(\placeholder)\bigr)(u).
\end{align}
\end{subequations}
The subscripts $\ell\in [N_\mathrm{e} + N_\mathrm{d}]$ emphasize that the trainable parameters of these SABs are different.
Integer $N_\mathrm{e}$ is the number of SABs, $\sSST_\ell, \ell=1,2,\ldots,N_\mathrm{e}$, in the encoder $\sFEnc$, while $N_\mathrm{d}$ is the number of SABs, $\sSST_\ell, \ell=1+N_\mathrm{e},2+N_\mathrm{e},\ldots,N_\mathrm{d}+N_\mathrm{e}$, in the decoder $\sFDec$. Specifically, we have
\begin{subequations}
\begin{align}    &\sSST_\ell:\mathcal{U}_F(\bbR^{d_\mathrm{latent}})\rightarrow \mathcal{U}_F(\bbR^{d_\mathrm{latent}}),\quad &&\ell=1,2,\ldots,N_\mathrm{e},\\
&\sSST_\ell:\mathcal{U}_F(\bbR^{d_s})\rightarrow \mathcal{U}_F(\bbR^{d_s}),\quad &&\ell=1+N_\mathrm{e},2+N_\mathrm{e},\ldots,N_\mathrm{d}+N_\mathrm{e}.
\end{align}
\end{subequations}
Further practical details about our set transformer architecture, including the dimensions of latent features and the choices of layers, are provided in \ref{app:nn_details}.

The trainable parameters in $\sFST$ are given by:
\begin{equation}
\begin{split}
    \theta(\sFST) &= \theta\bigl(\sFDec\bigr)\cup\theta\bigl(\sCST(s,\placeholder)\bigr)\cup\theta\bigl(\sFEnc\bigr)\\ &=\theta\bigl(\sFNN_\mathrm{in}\bigr)\cup\theta\bigl(\sFNN_\mathrm{out}\bigr)\cup\bigl(\cup_{\ell=1}^{N_\mathrm{e}+N_\mathrm{d}}\theta\bigl(\sSST_\ell\bigr)\bigr)\cup\theta\bigl(\sCST(s,\placeholder)\bigr).
\end{split}
\end{equation}

The set transformer architecture exhibits two key properties when processing an input sequence, namely, the permutation invariance (Proposition~\ref{prop:permutation_invariance}) and adaptation to variable input lengths (Proposition~\ref{prop:variant_input_length}). These two propositions are proved in \cite{lee2019set} under a slightly different setting: in the original work a length-$N$ sequence in $\bbR^d$ is represented as an $N\times d$ matrix, whereas here it is a mapping $u\in\mathcal{U}([N],\mathbb{R}^d)$.

\begin{proposition}[Permutation Invariance]\label{prop:permutation_invariance}
A permutation $\sigma$ is a bijective function from $[N]$ to $[N]$. We extend the definition of $\sigma$ to sequences in $u\in\mathcal{U}([N], \bbR^{d_u})$ so that
\begin{equation}
    \sigma(u)(n) = u(\sigma(n)), \quad n\in [N].
\end{equation}
Then for any permutation $\sigma:[N]\mapsto[N]$ and any input sequence $u\in\mathcal{U}([N], \bbR^{d_u})$, we have
\begin{equation}
    \sFST(u) = \sFST(\sigma(u)).
\end{equation}
\end{proposition}

\begin{proposition}[Adaptation to Variable Input Length]\label{prop:variant_input_length}
A set transformer architecture with fixed parameters takes input sequences with different lengths while the output dimension is fixed:
\begin{equation}
\label{eq:setref}
    \sFST(u)\in \bbR^{\dst},\quad \forall u\in\mathcal{U}([N], \bbR^{d_u}), \quad\forall N\in\bbN.
\end{equation}
\end{proposition}

Based on these two properties, we can bridge the sequence input and the probability measure input for the set transformer architecture. The permutation invariance property (Proposition \ref{prop:permutation_invariance}) enables us to abstract away from the sequential nature of the input and treat it as an unordered set. More fundamentally, this allows us to interpret the input as an empirical distribution, where each element in the sequence represents a sample from some underlying distribution. Furthermore, Proposition \ref{prop:variant_input_length} demonstrates that we can accommodate varying input lengths without architectural modifications, effectively allowing us to adjust the number of samples in the empirical distribution. Together, these properties (Propositions \ref{prop:permutation_invariance} and \ref{prop:variant_input_length}) provide theoretical foundation for analysis of the main components of the set transformer in the context of probability measures (Section~\ref{sec:AFM}).

\begin{remark}[Common Notation for Operations on Measures and on Sequences]
In this section, much of the notation coincides with that in Section~\ref{sec:AFM}. This is because the underlying definitions are almost identical, with the key distinction being that in Section~\ref{sec:AFM} the operators act on probability measures, whereas in this section they act on sequences. The preceding developments in this subsection explain the connection. For example, in Section~\ref{sec:AFM}, $\sS$ and $\sC$ denote the measure-based self-attention \eqref{eq:Smeasure} and cross-attention \eqref{eq:Cmeasure} blocks, while in this section, $\sSST$ and $\sCST$ represent the sequence-based self-attention \eqref{eq:SAB} (Definition~\ref{def:SAB}) and cross-attention \eqref{eq:CAB} (Definition~\ref{def:CAB}) blocks, respectively. In addition, $\sF$ refers to the transformer measure neural mapping \eqref{eq:composition} in Section~\ref{sec:AFM} while $\sFST$ refers to the set transformer \eqref{eq:st_architecture} acting on sequences.
\end{remark}

\subsection{Our Architecture}\label{ssec:architecture}
In this section we provide a detailed description the architecture in our proposed learnable DA method, MNMEF. The core of our approach is to learn a parameterized gain matrix (Subsection~\ref{sssec:learning_gain}), generalizing how the classical EnKF utilizes a Gaussian-inspired gain. Additionally, to enhance performance, we incorporate inflation and localization techniques (Subsections~\ref{sssec:inflation} and \ref{sssec:localization}), akin to their role in EnKF. 

In our architecture, we use a set transformer (Subsection~\ref{ssec:discrete_ST}) to process the ensemble $\{(\hvn,h(\hvn))\}_{n=1}^N$ into a fixed-dimensional feature vector. We define the set of finite subsets in $\bbR^d$ by,
\begin{equation}
    \mathcal{F}(\mathbb{R}^d) = \{ A \subseteq \mathbb{R}^d \mid A \text{ contains a finite set of elements} \}.
\end{equation}
Because of the permutation invariance property (Proposition~\ref{prop:permutation_invariance}), we may reinterpret the set transformer $\sFST$ as a map $F^\mathrm{ST}$ that acts on sets rather than sequences: $F^\mathrm{ST}: \mathcal{F}(\bbR^{d_v}\times\bbR^{d_y})\rightarrow \bbR^{\dst}$, given by
\begin{equation}\label{eq:FST}
    F^\mathrm{ST}\left(\{(\hvn,h(\hvn))\}_{n=1}^N;\theta_\mathrm{ST}\right) = f_v\in\bbR^{\dst}
\end{equation}
The feature vector $f_v$ can be viewed as a representation of the empirical distribution $\rho_h^{(N)}$ given in \eqref{eq:rhoN} and similarly as an approximate summary of $\rho_h$. This representation is subsequently utilized to learn the gain matrix, inflation, and the localization.

\begin{remark}[Choice of $\dst$]
By Proposition~\ref{prop:variant_input_length}, $F^\mathrm{ST}$ outputs a fixed dimension $\dst$ regardless of ensemble size $N$. Therefore, we do not need to adjust $\dst$ based on $N$. Larger $\dst$ values offer more expressiveness, but increase computational cost. Our tests with the Lorenz '96 model (Subsection~\ref{ssec:L96}) using $\dst=32,64,128$ showed only marginal improvements with increasing $\dst$, thus we selected $\dst=64$ for all experiments.
\end{remark}

\subsubsection{Learning the Gain Matrix}\label{sssec:learning_gain}
From the analysis step \eqref{eq:ourens_analysis} and
the proposed form \eqref{eq:k-theta}, we have the following form for the learnable analysis step:
\begin{subequations}
\begin{align}
    \vn &= \hvn + \Kt \left( y^\dagger - \hyn \right), \label{eq:kalman_gain_IB}\\
    \Kt &= \Kt^{(1)} \left( \Kt^{(2)} + \Gamma \right)^{-1}. \label{eq:k-theta-sec4}
    \end{align}
\end{subequations}
Recall that, in the mean-field limit, we impose an inductive bias on our architecture by defining $\Kt^{(1)}$ and $\Kt^{(2)}$ in the form \eqref{eq:full_K}, allowing for a learned correction through \eqref{eq:FNO}. Recall that $\hat{m}$ and $\hat{h}$ are the mean of $\{\hvn\}_{n=1}^N$ and $\{h(\hvn)\}_{n=1}^N$, respectively. The empirical version of \eqref{eq:full_K}, used in the practical implementation, takes the form
\begin{subequations}\label{eq:full_K-sec4}
\begin{align}
    \Kt^{(1)} &= \frac{1}{N} \sum_{n=1}^N \left(\hvn - \hat{m} + \hwn_\theta\right) \otimes \left(h(\hvn) - \hat{h} + \hzn_\theta\right), \label{eq:full_K1-sec4}\\
    \Kt^{(2)} &= \frac{1}{N} \sum_{n=1}^N \left(h(\hvn) - \hat{h} + \hzn_\theta\right) \otimes \left(h(\hvn) - \hat{h} + \hzn_\theta\right), \label{eq:full_K2-sec4}
\end{align}
\end{subequations}
where $\hwn_\theta$ and $\hzn_\theta$ are trainable correction terms that depend on the state $\hvn$, the observation $h(\hvn)$, the true observation $\yd$, and the ensemble $\{(\hvn,h(\hvn))\}_{n=1}^N$.

The correction terms are defined by a neural operator $\fF$ as in~\eqref{eq:FNO} for the mean-field perspective presented in Section~\ref{sec:MFF}. In our practical implementation, we replace the dependence on $\rho_h$ in $\fF$ by the ensemble $\{(\hv^{(\ell)},h(\hv^{(\ell)}))\}_{\ell=1}^N$ and use the set transformer $F^\mathrm{ST}$ \eqref{eq:FST} to process the ensemble into a feature vector $f_v$. Recall that this can be viewed as using the empirical approximation $\rho_h^{(N)} \approx \rho_h$ given by \eqref{eq:rhoN}.

To learn the correction terms for each particle, we then train a neural network $F^\mathrm{gain}: \bbR^{d_v}\times \bbR^{d_y}\times \bbR^{d_y}\times \bbR^{\dst}\rightarrow \bbR^{d_v}\times \bbR^{d_y}$ with the standard multilayer perceptron (MLP) architecture that takes the feature vector for the ensemble as an input: 
\begin{equation}\label{eq:nn_F}
    F^\mathrm{gain}\left(\hvn, h(\hvn), y^\dagger, f_v; \theta_\mathrm{gain}\right) = \left(\hwn_\theta, \hzn_\theta\right),
\end{equation}
where $\theta_\mathrm{gain}$ denotes trainable parameters.

From \eqref{eq:full_K1-sec4} and \eqref{eq:full_K2-sec4}, it is evident that if the correction terms satisfy $\hwn_\theta = \hzn_\theta = 0$, the formulation reduces to the EnKF \eqref{eq:cov_enkf}. When the dynamics and observation models are linear, or in other words, when the filtering distribution of the states $v$ is Gaussian, the Kalman filter achieves optimality \cite{kalman1960new, evensen2003ensemble, calvello2022ensemble}. Therefore, the motivation for our proposed scheme \eqref{eq:k-theta} lies in its optimality in the linear (or Gaussian) setting, while further aiming to improve the performance of the Kalman filter in nonlinear problems by learning the correction terms.

\begin{remark}[On the Invertibility in the Gain Computation]
The matrix $\Kt^{(2)}$, defined in \eqref{eq:full_K2-sec4} as an average of outer products, is positive semi-definite. Since $\Gamma$ is a fixed positive definite matrix, the sum $\Kt^{(2)} + \Gamma$ is also positive definite and thus invertible. Crucially, the minimum eigenvalue of $\Kt^{(2)} + \Gamma$ is bounded from below by the minimum eigenvalue of $\Gamma$, which is a positive constant independent of any trainable parameters. This property ensures the numerical stability of computing $(\Kt^{(2)} + \Gamma)^{-1}$ during backpropagation.
\end{remark}

\begin{remark}[Correction Approach Versus End-to-end Gain Learning]
In addition to the proposed approach with correction terms, we have implemented the end-to-end learning of the operator $\fB$ presented in Subsection~\ref{ssec:prob_map} and learning the Kalman gain $\Kt$ directly without any constraint on its form. We find that the end-to-end approach or directly learning $\Kt$ results in unstable and slower learning. A poorly initialized or an unconstrained $\Kt$ leads to substantial divergence of the resulting filtered trajectory away from the ground truth trajectory. The EnKF-like structure in \eqref{eq:k-theta-sec4} with learnable correction terms provides better stability while maintaining adaptability to nonlinear dynamics.
\end{remark}

\subsubsection{Learning the Inflation}\label{sssec:inflation}
Inflation is a technique used to increase the ensemble spread, mitigating the underestimation of analysis covariance caused by sampling errors in ensemble-based methods. Finite ensemble sizes often lead to sampling errors in the forecast covariance, causing the filter to overly trust forecasts and risk filter divergence. Inflation helps maintain the response of the filter to observations by counteracting these effects.

The most common form of inflation is multiplicative inflation, where the analysis ensemble $\{v^{(n)}\}_{n=1}^N$ that is calculated after the analysis step \eqref{eq:ourens_analysis} is modified as follows
\begin{equation}\label{eq:infl}
\begin{split}
    v^{(n)} &\gets  v^{(n)} + (\alpha - 1)\left(v^{(n)} - \frac{1}{N}\sum_{\ell=1}^Nv^{(\ell)}\right),
\end{split}
\end{equation}
where $\alpha\geq 1$ is the inflation parameter. In our learning framework, we replace the term proportional to $(\alpha - 1)$ by a learned correction term $\hun_\theta$,

that is output by an MLP-based neural network $F^\mathrm{infl}:\bbR^{d_v}\times \bbR^{\dst}\rightarrow \bbR^{d_v}$ given by
\begin{equation}\label{eq:f_infl}
    F^\mathrm{infl}\bigl(\vn, f_v;\theta_\mathrm{infl}\bigr) = \hun_\theta,
\end{equation}
where $\vn$ is the estimated state given by \eqref{eq:ourens_analysis}, and $\theta_\mathrm{infl}$ denotes trainable parameters. 
The learned inflation step is processed after the analysis step \eqref{eq:ourens_analysis}, resulting in the particle update:
\begin{equation}\label{eq:our_infl}
    v^{(n)} \gets v^{(n)} + \hun_\theta.
\end{equation}

\begin{remark}[Dependence and Implications of the Learned Inflation Term]
The common form of inflation is a function of the ensemble states, i.e., not the observations, in~\eqref{eq:infl}.
Similarly, the learned inflation term $\hun_\theta$ in \eqref{eq:f_infl} depends on the ensemble states $\hat{v}^{(n)}$ and the feature vector $f_v$, without information of the true observation $y^\dagger$ or the observation noise covariance $\Gamma$. We avoid the risk that the inflation term would evolve into an end-to-end correction term, ensuring the learning performed by other components in our framework remains meaningful and effective.
\end{remark}

\subsubsection{Learning the Localization}\label{sssec:localization}
In spatially extended systems, sampling errors can lead to inaccurate covariance estimation, particularly when the sample size is small. True correlations typically decay with distance, but spurious long-range correlations may arise, distorting the covariance structure. Localization addresses this issue by suppressing artificial correlations while preserving meaningful local correlations.

Localization is implemented by damping covariances in an empirical covariance matrix based on distance. This is typically achieved by applying a Hadamard product (elementwise product, denoted by $\circ$) between the empirical covariance matrix and a predefined localization weight matrix $L$. For the dynamic model \eqref{eq:dynamic} with the state in $d_v$-dimensional space (i.e., $v\in\bbR^{d_v}$), $L\in\bbR^{d_v\times d_v}$ is calculated by
\begin{equation}\label{eq:loc_weight}
    (L)_{k\ell} = g(D_{k\ell}),\quad \forall k,\ell \in [d_v],
\end{equation}
where $g:\bbR\rightarrow\bbR$ is a function that maps distance values to localization weights (e.g., the Gaspari--Cohn function \cite{gaspari1999construction}), and $D_{k\ell}$ is the distance between index $k$ and $\ell$.
To incorporate the localization technique in the EnKF, the Kalman gain formula \eqref{eq:EnKF_gain} is modified as
\begin{equation}\label{eq:EnKF_loc}
    K = \left(\hat{C}^{vh}\circ L^{vh}\right)\left(\hat{C}^{hh} \circ L^{hh} + \Gamma\right)^{-1},
\end{equation}
where $L^{vh}\in \bbR^{d_v\times d_y}$ and $L^{hh} \in \bbR^{d_v\times d_y}$ are localization weight matrices corresponding to $\hat{C}^{vh}$ and $\hat{C}^{hh}$, respectively.

In our learning-based formulation of localization we proceed as follows: we incorporate localization into our parameterized Kalman gain by modifying \eqref{eq:k-theta-sec4} to take the form
\begin{equation}\label{eq:loc_k-theta}
    \Kt = \left(\Kt^{(1)}\circ \Lt^{(1)}\right) \left( \Kt^{(2)}\circ \Lt^{(2)} + \Gamma \right)^{-1},
\end{equation}
The localization weight matrices $\Lt^{(1)}$ and $\Lt^{(2)}$ follow the structure in~\eqref{eq:loc_weight},
where $g$ is replaced by a parameterized distance-to-weight function $g_\theta:\bbR\rightarrow\bbR$.

In practice, it is not necessary to learn the entire function $g_\theta$ since we only have a finite number of different distances, defined by the spatial grid for discretized PDEs, and time indices in the setting of ODEs. We denote the unique distance values in $\{D_{k\ell}\}_{k,\ell=1}^{d_v}$ as $\{D_1, D_2, \ldots, D_{N_D}\}$, where $N_D$ is the number of distinct distances. Given a vector $\hg = [\hg_1, \hg_2, \ldots, \hg_{N_D}] \in \mathbb{R}^{N_D}$, we can define the function $g_\theta$ by
\begin{equation}
g_\theta(D_i) = \hg_i, \quad\forall i\in [N_D].
\end{equation}
We employ a MLP-based neural network $F^\mathrm{loc}: \bbR^{\dst}\rightarrow \bbR^{N_D}$ given by:
\begin{equation}\label{eq:f_loc}
F^\mathrm{loc}\left(f_v; \theta_\mathrm{loc}\right) = \hg.
\end{equation}
The parameterized localization weight matrices $\Lt^{(1)}$ and $\Lt^{(2)}$ are calculated based on the output of this learned function $g_\theta$. In \ref{app:learn_loc}, we provide a step-by-step example illustrating the computation of the parameterized localization weight matrices.

\begin{remark}[Implementation Insights for Learning Localization Weights]
Here we provide more details for the learning process of the localization weight matrices and make some comments on our proposed approach. 
\begin{itemize}
    \item The final layer of $F^\mathrm{loc}$ is a $\mathrm{softmax}$ function scaled by 2, ensuring all outputs lie within the interval $[0,2]$. Although amplifying underestimated covariances is typically the responsibility of  inflation alone, our architecture jointly performs localization and inflation. Thus, we allow the weights to range in $[0,2]$ rather than $[0,1]$ to grant additional flexibility in the localization component.
    \item Instead of learning a continuous parametric form for $g_\theta$, we learn its values only at discrete distances $\{D_i\}_{i=1}^{N_D}$. This reduces model complexity while preserving localization expressivity.
    
    \item While EnKF-based localization methods typically use fixed weight matrices across time steps (e.g., using the Gaspari--Cohn function \cite{gaspari1999construction}), there is much interest in adaptive approaches \cite{vishny_high-dimensional_2024}. Our learning-based approach adapts these matrices for each time step $j$ using neural networks via its dependence on the ensemble. As shown in \eqref{eq:f_loc} (time index $j$ omitted for simplicity), the network processes time-step-specific inputs, allowing the localization scheme to dynamically adjust based on the current system state.
\end{itemize}
\end{remark}

\begin{remark}[On sharing the distance-to-weight function across $L^{(1)}_{\theta}$ and $L^{(2)}_{\theta}$]
In forthcoming experiments in this paper, the observation operator $h$ subsamples state components, so the distances between observation indices are induced by the same state-space metric used for state indices. Accordingly, we learn a single distance-to-weight function $g_{\theta}$ and use it consistently to build $L^{(1)}_{\theta}$ and $L^{(2)}_{\theta}$. For more general $h$ (e.g., nonlinear or indirect observations), one may instead learn two distinct functions, $g^{(1)}_{\theta}$ and $g^{(2)}_{\theta}$, tailored to state–observation pairs for $L^{(1)}_{\theta}$ and observation–observation pairs for $L^{(2)}_{\theta}$, respectively.
\end{remark}

\subsection{Loss Function}\label{ssec:training_parms}

This section outlines the procedure for training the neural network parameters, including the gain, inflation and localization (Subsection~\ref{ssec:architecture}). Based on the stochastic dynamic model \eqref{eq:dynamic} and the data observation model \eqref{eq:observation}, we make the follow assumption for the training:

\begin{dataassumption}\label{da:fix_params}
Our training is conducted for fixed dynamics $\Psi$, observation operator $h$, and noise covariance matrices $\Sigma$ and $\Gamma$. 
The time step $\Delta t$ is also fixed (and incorporated in $\Psi$). The initial condition is a Gaussian with a fixed covariance $C_0$, i.e. $v_0\sim \normal (\placeholder, C_0)$.
For some $M,J\in\mathbb{N}$, the training data consists of $M$ trajectories of length $J + 1$, extracted from a single long trajectory generated by propagating the dynamic model \eqref{eq:dynamic} for $M(J + 1)$ steps. Each sub-trajectory $\{v^\dagger_j\}_{j=0}^J$ has corresponding observations 
$\{y^\dagger_j\}_{j=1}^J$ generated according to \eqref{eq:observation}.
\end{dataassumption}

For each trajectory, we generate estimated states from our ensemble-based data assimilation model to compare with the true states from the dynamic model, enabling us to train the parameters $\theta_\mathrm{ST}$, $\theta_\mathrm{gain}$, $\theta_\mathrm{infl}$, and $\theta_\mathrm{loc}$, for the set transformer $F^\mathrm{ST}$ \eqref{eq:FST}, and MLPs $F^\mathrm{gain}$ \eqref{eq:nn_F}, $F^\mathrm{infl}$ \eqref{eq:f_infl} and $F^\mathrm{loc}$ \eqref{eq:f_loc} respectively.

We initialize an ensemble $\{v_0^{(n)}\}_{n=1}^N$ at the initial time step $j = 0$ for each sub-trajectory, with members sampled i.i.d. from a Gaussian distribution with the mean $\vn_0$ and the covariance $C_0$:
\begin{equation}\label{eq:initial_ensemble}
v_0^{(n)} \sim \normal(v^\dagger_0, C_0), \quad n = 1, \ldots, N.
\end{equation}
For $j=0,1,\ldots,J-1$ and trainable parameters $\theta = \{\theta_\mathrm{ST}, \theta_\mathrm{gain}, \theta_\mathrm{infl}, \theta_\mathrm{loc}\}$, we update the ensemble $\{v_j^{(n)}(\theta)\}_{n=1}^N$ to $\{v_{j+1}^{(n)}(\theta)\}_{n=1}^N$ according to
\begin{subequations}\label{eq:train_evolve}
\begin{align}
    \hvn_{j+1}(\theta) &= \Psi(v_j^{(n)}(\theta)) + \xi_j^{(n)}, \quad \xi_j^{(n)}\sim\normal(0,\Sigma),\label{eq:ours_ens_pred}\\
    \hyn_{j+1}(\theta) &= h(\hvn_{j+1}(\theta)) + \eta_{j+1}^{(n)}, \quad \eta_{j+1}^{(n)}\sim\normal(0,\Gamma), \label{eq:ours_ens_extend}\\
    \vn_{j+1}(\theta) &= (\hvn_{j+1}(\theta) + \hun_\theta) + (\Kt)_{j+1} \left( y^\dagger_{j+1} - \hyn_{j+1}(\theta) \right), \label{eq:ours_ens_analysis}
\end{align}
\end{subequations}
where $(\Kt)_{j+1}$ is our learned parameterized gain matrix, including localization, and $\hun_\theta$ is the inflation term according to Subsection~\ref{ssec:architecture}. For all parameters $\theta$ we choose $v_0^{(n)}(\theta) = v_0^{(n)}$. The ensembles $\{v_j^{(n)}(\theta)\}_{n=1}^N$ depend on the trainable parameters $\theta$ for $j=1,2,\ldots,J$.

In this paper we focus on state estimation, given by the mean of the estimated ensemble $\{v_j^{(n)}(\theta)\}_{n=1}^N$, i.e., 
\begin{equation}\label{eq:ensemble_mean}
    \bar{v}_j(\theta) = \frac{1}{N}\sum_{n=1}^N v_j^{(n)}(\theta).
\end{equation}
To train our neural network parameters effectively, we introduce a loss function that quantifies the discrepancy between the estimated states and the true states across the $m$-th sub-trajectory $V_m = \{v^\dagger_j\}_{j=1}^J$, given by:
\begin{equation}\label{eq:loss}
    \mathcal{L}_m(\theta) = \frac{1}{J}\sum_{j=1}^{J}\frac{\|\bar{v}_j(\theta)- v_j^\dagger\|_2^2}{\|v_j^\dagger\|_2^2},
\end{equation}
which measures the relative accuracy of our ensemble mean predictions compared to the true trajectory. Our loss \eqref{eq:loss} is a relative squared $L_2$ loss, which is preferred over the standard $L_2$ loss for two main reasons. First, using the squared $L_2$ norm avoids computing square roots, which can lead to numerical instabilities when calculating gradients, especially when the error approaches zero. Second, normalizing by the true state magnitude ($\|v_j^\dagger\|_2^2$) makes the loss scale-invariant across different state variables and trajectory segments, ensuring balanced parameter updates regardless of the absolute magnitudes of the state components. This relative formulation is particularly important in dynamical systems where state variables may span different orders of magnitude.

In the training process, the loss is averaged across $M$ training trajectories indexed by $m\in [M]$, 
\begin{equation}\label{eq:batch_loss}
    \mathcal{L}_{[M]}(\theta) = \frac{1}{M}\sum_{m\in [M]}\mathcal{L}_m(\theta).
\end{equation}
The trainable parameter $\theta$ is optimized based on the above-defined loss $\mathcal{L}_{[M]}$. In practice, the loss $ \mathcal{L}_{[M]}$ is approximated by a mini-batch as a subset of $[M]$.

\begin{remark}[Gradient Truncation in Sequential Learning]\label{rmk:gradient-detach}
Computing the loss \eqref{eq:loss} requires evaluating ensemble estimates $\{\{v_j^{(n)}(\theta)\}_{n=1}^N\}_{j=1}^J$ across the entire trajectory. For the ensemble at step $j$, this involves nested evaluations of $\Kt$ up to $j$ times. These nested evaluation arise from the sequential nature of the trajectory, where each step depends on the computing of the Kalman gain from the previous steps, requiring backpropagation through the entire computational history. With large trajectory lengths $J$, this causes slow training and numerical instability since it is nearly equivalent to training a very deep neural network.

To address this issue, we introduce a gradient-detach hyperparameter $J_0 \in \bbN$. This hyperparameter limits the gradient computation for the loss related to time step $j$ to only the steps $j, j-1, \ldots, j-J_0+1$. This approach assumes that the trajectory at time step $j$ is not significantly affected by parameter changes in the Kalman gain for earlier time steps, which is reasonable given the diminishing influence of distant past states in many dynamical systems. This technique significantly improves training speed and stability.
\end{remark}

\begin{remark}[Training Targets and Probabilistic Losses]\label{rmk:prob-loss}
In the loss \eqref{eq:loss}, the ground-truth states used for supervised training are obtained by simulating the known dynamical model (forward integration of $\Psi$ with prescribed noise model). Thus, the availability of training data hinges on access to an explicit and tractable dynamic model. In high-dimensional settings where such a model is unavailable or impractical, a surrogate simulator (e.g., a reduced-order or data-driven one) can be used to synthesize training trajectories and observations. Furthermore, while the proposed loss \eqref{eq:loss} focuses on state estimation, a promising direction for future work is to adopt probabilistic loss functions that target the full filtering distribution $\bbP(v_j\mid Y_j)$ rather than only the mean-based point estimator; see Subsection~\ref{ssec:LR}.
\end{remark}

\subsection{Fine-Tuning Inflation \& Localization}\label{ssec:fine-tuning}
Following the developments in Subsection~\ref{ssec:architecture}, we may apply a set transformer \eqref{eq:FST} to process ensembles of any size into fixed-dimensional feature vectors. This allows for pretraining our MNMEF with ensemble size $N$ and perform inference with $N' \neq N$. However in classical EnKF formulations, hyperparameters such as inflation and localization depend on the ensemble size, and allowing for this
dependence significantly enhances performance. Thus, we consider fine-tuning a subset of the parameters for inference at ensemble size $N' \ne N$. We now detail this methodology.

The architecture defined in Subsection~\ref{ssec:architecture} is based on trainable parameters  $\theta_\mathrm{ST}$, for the set transformer $F^\mathrm{ST}$ \eqref{eq:FST}, and  $\theta_\mathrm{gain}$, $\theta_\mathrm{infl}$, and $\theta_\mathrm{loc}$ for the MLPs $F^\mathrm{gain}$ \eqref{eq:nn_F}, $F^\mathrm{infl}$ \eqref{eq:f_infl} and $F^\mathrm{loc}$ \eqref{eq:f_loc} respectively. For efficient fine-tuning when transitioning from ensemble size $N$ to $N'$, we freeze $\theta_\mathrm{ST}$ and only fine-tune $\theta_\mathrm{gain}$, $\theta_\mathrm{infl}$, and $\theta_\mathrm{loc}$. This specific choice of fine-tuning is effective because:
(1) freezing $\theta_\mathrm{ST}$ significantly reduces computational costs, as the set transformer contains the majority of network parameters; (2) the remaining parameters, though small in number, measurably improve performance when they are adapted to the ensemble size.

The training loss for the $m$-th sub-trajectory $V_m = \{v^\dagger_j\}_{j=1}^J$ is defined by \eqref{eq:ensemble_mean}, \eqref{eq:loss}:
\begin{equation}\label{eq:ft_loss}
    \mathcal{L}_m^{(N)}(\theta_\mathrm{ST}, \theta_\mathrm{gain}, \theta_\mathrm{infl}, \theta_\mathrm{loc}) = \frac{1}{J}\sum_{j=1}^{J}\frac{\|\bar{v}_j(\theta)- v_j^\dagger\|_2^2}{\|v_j^\dagger\|_2^2},
\end{equation}
where we add the superscript $N$ to emphasize dependence on the ensemble size. In pretraining with size $N$ we solve
\begin{equation}\label{eq:pretrain_N_problem}
\theta_\mathrm{ST}^{(N)}, \theta_\mathrm{gain}^{(N)}, \theta_\mathrm{infl}^{(N)}, \theta_\mathrm{loc}^{(N)} = \argmin_{\theta_\mathrm{ST}, \theta_\mathrm{gain}, \theta_\mathrm{infl}, \theta_\mathrm{loc}}\; \mathcal{L}_{[M]}^{(N)}(\theta_\mathrm{ST}, \theta_\mathrm{gain}, \theta_\mathrm{infl}, \theta_\mathrm{loc}).
\end{equation}
During fine-tuning for size $N'\ne N$, using the same training trajectories we solve
\begin{equation}\label{eq:ft_problem}
    \theta_\mathrm{gain}^{(N')}, \theta_\mathrm{infl}^{(N')}, \theta_\mathrm{loc}^{(N')} = \argmin_{\theta_\mathrm{gain}, \theta_\mathrm{infl}, \theta_\mathrm{loc}}\; \mathcal{L}^{(N')}_{[M]}(\theta_\mathrm{ST}^{(N)}, \theta_\mathrm{gain}, \theta_\mathrm{infl}, \theta_\mathrm{loc})
\end{equation}
with fixed $\theta_\mathrm{ST}^{(N)}$. In practice, we use small mini-batches of $[M]$ to estimate the losses in~\eqref{eq:pretrain_N_problem} and~\eqref{eq:ft_problem}. 

With the updated parameters, we proceed with ensemble size $N'$ using:
\begin{subequations}
\begin{align}
    &\text{Feature Vector: }&&f_v^{(N')} = F^\mathrm{ST}\left(\{(\hvn,h(\hvn))\}_{n=1}^{N'};\theta_\mathrm{ST}^{(N)}\right), \\
    &\text{Gain: }&&F^\mathrm{gain}\left(\placeholder, f_v^{(N')}; \theta_\mathrm{gain}^{(N')}\right),\\
    &\text{Inflation: }&&F^\mathrm{infl}\left(\placeholder, f_v^{(N')}; \theta_\mathrm{infl}^{(N')}\right),\\
    &\text{Localization: }&&F^\mathrm{loc}\left(f_v^{(N')}; \theta_\mathrm{loc}^{(N')}\right).
\end{align}
\end{subequations}
Although $\theta_\mathrm{ST}^{(N)}$ remains unchanged after the fine-tuning, the feature vector $f_v^{(N')}$ differs from the pretraining features $f_v$ since the input ensemble size changes from $N$ to $N'$.

\begin{remark}[Efficient Training Strategy]
While the per-particle update in the analysis step has linear complexity with respect to the ensemble size $N$ for fixed gain matrix, the overall cost of one forward pass is dominated by the attention operations used to compute correction terms. These attention blocks scale quadratically in $N$. In practice, we adopt an efficient training strategy by pretraining on smaller ensemble sizes $N$ and applying fine-tuning for larger ensembles $N' > N$, which maintains comparable performance while substantially reducing computational cost. Further improvements are possible by employing localized or linear-attention variants that can reduce the scaling in $N$.
\end{remark}

In Subsection~\ref{ssec:Exp_FT}, we demonstrate this fine-tuning method's effectiveness, achieving comparable performance to full parameter fine-tuning with significantly improved efficiency.

\section{Numerical Experiments}
\label{sec:NUM_EXP}
In this section we present numerical experiments to demonstrate the improved numerical benefit of our proposed method, MNMEF, in comparison with leading existing ensemble methods; the experiments also serve to validate the effectiveness of several specific details within our approach. First, in Subsections~\ref{ssec:L96}, \ref{ssec:KS}, and \ref{ssec:L63}, we compare our method with several classical approaches across several nonlinear dynamical systems -- Lorenz '96, Kuramoto--Sivashinsky, and Lorenz '63.
In Subsection~\ref{ssec:cost}, we compare the computational time between our MNMEF and other benchmarks. In Subsection~\ref{ssec:stab_analysis}, we show the robustness of our method to the inherent randomness in test trajectories. In Subsection~\ref{ssec:Exp_inf_loc}, we conduct experiments to study the learning-based inflation and localization methods proposed in Subsections~\ref{sssec:inflation} and \ref{sssec:localization}. Subsequently, in Subsection~\ref{ssec:Exp_FT}, we perform experiments
to illustrate the efficiency and effectiveness of our proposed fine-tuning method, introduced in Subsection~\ref{ssec:fine-tuning}. The code for our method is publicly available\footnote{\url{https://github.com/wispcarey/DALearning}} and can be used to reproduce all of our experiments. 

We reiterate that all reported experiments
are for nonlinear problems. Our methodology is based on learning corrections to the Kalman filter structure inherent in the EnKF. For nonlinear problems there
is bias in the EnKF which can be effectively learned and corrected for, and this is the essence of our approach. For linear problems there is no bias in the mean-field limit (although there will still be a bias with a finite ensemble \cite{sacher_sampling_2008}), so our
methodology would attempt to fit neural networks to what is largely noise. Without any regularization or early stopping, this can lead to filter divergence for large enough training epochs. \ref{app:lin} reports a linear–Gaussian experiment to illustrate this.

\subsection{Experimental Set-Up}
\label{ssec:ESU}
In the following three Subsections~\ref{ssec:L96}, \ref{ssec:KS}, and~\ref{ssec:L63}, we compare our method against several benchmark approaches: 
\begin{enumerate}
\item \textbf{Localized EnKF Perturbed Observation (EnKF)} \cite{burgers1998analysis, anderson2001ensemble}: This is the classic implementation of EnKF that employs a stochastic update approach, where the observations are perturbed by adding random noise samples drawn from the observation noise distribution. 
    
\item \textbf{Ensemble Square Root Filter (ESRF)} \cite{tippett_ensemble_2003, sakov2008deterministic}: This method is a deterministic variant of the EnKF. It directly transforms the ensemble using matrix square root operations to match the second order statistics of the EnKF, but without requiring stochastic perturbations.
    
\item \textbf{Local Ensemble Transform Kalman Filter (LETKF)} \cite{hunt2007efficient}: This method builds upon the Ensemble Transform Kalman Filter (ETKF) \cite{bishop2001adaptive} by incorporating spatial localization. ETKF is also a deterministic version of EnKF. Under linear–Gaussian assumptions and in the mean-field limit, ESRF and ETKF are theoretically equivalent, but they differ in practical implementation. 

\item \textbf{Iterative Ensemble Kalman Filter (IEnKF)} \cite{sakov2012iterative}: This method enhances the EnKF (perturbed observation) by incorporating a variational scheme to better handle strongly nonlinear systems. It iteratively solves a minimization problem between adjacent time steps, progressively refining the solution for the analysis step through multiple iterations. In our experiments, we set the maximum number of iterations to $10$ and use a convergence tolerance of $10^{-5}$.

\end{enumerate}

\begin{remark}
Inflation and localization are important for DA in high-dimensional systems. In our experiments, all benchmark filters (EnKF, ESRF, LETKF, and IEnKF) employ post–analysis inflation as in~\eqref{eq:infl}. For the higher-dimensional Lorenz–96 (Subsection~\ref{ssec:L96}) and Kuramoto–Sivashinsky (Subsection~\ref{ssec:KS}) systems, we apply localization to the \textbf{EnKF} and \textbf{LETKF}. We employ the Gaspari--Cohn (GC) function \cite{gaspari1999construction} as the distance-to-weight function for the localization according to the DAPPER package \cite{raanes_dapper_2024}. We do \emph{not} localize ESRF or IEnKF for the following reasons: 
\begin{enumerate}
    \item LETKF can be viewed as ETKF with localization, and ETKF is theoretically equivalent to ESRF in the mean-field/linear–Gaussian regime; adding localization to ESRF would thus be essentially redundant with LETKF.
    \item For IEnKF, a localized implementation is omitted because it requires multiple iterations with the dynamic model integrations at each step, making a joint grid search over inflation and localization computationally prohibitive for Lorenz '96 and KS, and because without careful radius tuning the method exhibited numerical instability (frequent NaNs due to filter divergence). 
\end{enumerate}
The hyperparameters for inflation and localization are optimized separately for each ensemble size via grid search; detailed settings and results are reported in \ref{app:grid_search}.
\end{remark}

In addition to the methods described above, we evaluate our MNMEF approach at various ensemble sizes. Unless otherwise specified, our MNMEF is pretrained with ensemble size $N=10$ as described in Subsection~\ref{ssec:training_parms}. We then examine our model's performance across different ensemble sizes $N\in \{5, 10, 15, 20, 40, 60, 100\}$ to demonstrate the generalizability of our approach.

Our evaluation includes two variants of our method:
\begin{enumerate}
\item \textbf{Pretrained MNMEF}: Trained once at $N=10$ for $1000$ epochs and applied directly to all ensemble sizes, demonstrating transfer capability across different ensemble configurations. This appears as \textbf{`Pretrain'} in our figures.
\item \textbf{Fine-tuned MNMEF}: Fine tune the pretrained model specific to each target ensemble size for $20$ epochs, as detailed in Subsection~\ref{ssec:fine-tuning}. This appears as \textbf{`Tuned'} in our figures.
\end{enumerate}

The intermediate feature dimensions in the set transformer for different dynamic models are provided in \ref{app:nn_details}. The detailed intiail conditions for different dynamic models are provided in \ref{app:initial_conditions}. The training hyperparameters including trajectory lengths, learning rates and batch sizes, are provided in \ref{app:hyperparameter_for_training}. 

For a fair comparison, all benchmark methods (EnKF, LETKF, and IEnKF) are tuned by RMSE with respect to the true state $v_j^\dagger$. Specifically, hyperparameters for inflation and localization are selected via comprehensive grid searches at multiple ensemble sizes to minimize true-state RMSE. The explored parameter spaces and heatmaps of the grid searches are reported in \ref{app:grid_search}.
This approach ensures that all benchmark methods achieve close to their optimal performance; the benchmarks thus present a robust test of our proposed methodology. Notably, we performed hyperparameter selection for the benchmarks directly on their test set performance, without using a validation set, which only strengthens the benchmark methods' advantage in our comparisons. 

The performance of the methods is evaluated using the difference between the estimated state (i.e., the mean of particles) and the ground truth state over the entire trajectory. In the continuous-time setting, let $\bfv^\dagger(t):[0,T]\rightarrow \bbR^{d_v}$ denote the ground truth trajectory between time $0$ and $T$, and let $\bfv(t):[0,T]\rightarrow \bbR^{d_v}$ denote the estimated state trajectory. Let $\|\cdot\|_2$ denote the Euclidean norm in $\mathbb{R}^{d_v}$. Then, the performance can be evaluated using the relative $L^1:=L^1([0,T];\bbR^{d_v})$ error:
\begin{equation}\label{eq:cont_error}
    \mathcal{E}(\bfv,\bfv^\dagger) = \frac{\|\bfv - \bfv^\dagger\|_{L^1}}{\|\bfv^\dagger\|_{L^1}} = \frac{\int_0^T \|\mathbf{v}(t) - \mathbf{v}^\dagger(t)\|_2 dt}{\int_0^T \|\mathbf{v}^\dagger(t)\|_2 dt}.
\end{equation}

In the discrete-time setting, we interpret the discrete time index $j$ as time $t_j = j\Delta t$, where $\Delta t = T / J$ is the time step size. The estimated trajectory is given by the sequence $V^{(N)} = \{\bar{v}^{(N)}_j\}_{j=1}^J$ where $\bar{v}^{(N)}_j = \frac{1}{N}\sum_{n=1}^{N} v^{(n)}_j$, and the ground truth trajectory is given by $V^\dagger = \{v_j^\dagger\}_{j=1}^J$. The relative discrete $L^1$ error can then be defined as
\begin{equation}
    \mathcal{E}(V^{(N)}, V^\dagger) = \frac{\sum_{j=1}^J \|\bfv(j\Delta t) - \bfv^\dagger(j\Delta t)\|_2 \, \Delta t}{\sum_{j=1}^J \|\bfv^\dagger(j\Delta t)\|_2 \, \Delta t} = \frac{\sum_{j=1}^J \|\bar{v}_j - v_j^\dagger\|_2}{\sum_{j=1}^J \|v_j^\dagger\|_2}.
\end{equation}
This error metric can be interpreted as a form of relative root-mean-square error (RMSE). At each time step $j$, the quantity $\|\bar{v}_j - v_j^\dagger\|_2$ corresponds to the root-mean-square error over the $d_v$-dimensional state vector. The overall metric thus represents a time-averaged spatial RMSE, normalized by the magnitude of the ground truth trajectory. In what follows, we refer to this error $\mathcal{E}(V^{(N)}, V^\dagger)$ as the relative RMSE (R-RMSE).

For $M_\mathrm{test}$ trajectories $\{V_m^\dagger\}_{m=1}^{M_\mathrm{test}}$ and their corresponding estimated trajectories $\{V_m^{(N)}\}_{m=1}^{M_\mathrm{test}}$, the averaged R-RMSE $\bar{\mathcal{E}}$ is given by:
\begin{equation}\label{eq:avg_r_rmse}
    \bar{\mathcal{E}} = \frac{1}{M_\mathrm{test}}\sum_{m=1}^{M_\mathrm{test}}\mathcal{E}(V_m^{(N)},V_m^\dagger).
\end{equation}
In the experiments presented in the remainder of this section, we adopt the averaged R-RMSE $\bar{\mathcal{E}}$ as the primary evaluation metric, allowing us to ameliorate the effect of randomness inherent in each individual run and providing a more reliable assessment of the overall performance of different methods. Unless otherwise specified, the number of test trajectories is fixed at $M_\mathrm{test} = 64$.

To further highlight the advantages of our method, we consider the relative improvement in terms of R-RMSE $\mathcal{E}_\mathrm{RI}$ defined as:
\begin{equation}\label{eq:relative_imp}
\mathcal{E}_\mathrm{RI} = \frac{\bar{\mathcal{E}}_\mathrm{benchmark} - \bar{\mathcal{E}}_\mathrm{ours}}{\bar{\mathcal{E}}_\mathrm{benchmark}}.
\end{equation}
This metric quantifies the percentage reduction in R-RMSE achieved by our method compared to the benchmark method. When $\mathcal{E}_{RI}<0$, our method performs worse than the benchmark.

We apply the following general hyperparameter setting in all of our experiments unless otherwise specified. The dynamic operator $\Psi$ in Equation~\eqref{eq:dynamic} is obtained by solving a differential equation using the fourth-order Runge--Kutta (RK4) \cite{stoer1980introduction} method. The time step $\Delta t$ denotes the interval between consecutive observations. For accurate numerical integration of the dynamics, we perform multiple RK4 steps within each interval $\Delta t$, utilizing an integration step size smaller than $\Delta t$. The dynamic noise $\xi \sim \mathcal{N}(0, \Sigma)$ is typically set at $\Sigma = \sigma_v^2 I$ where $I$ is the identity matrix and $\sigma_v =10^{-3}$ or $0.$ For the observation noise, $\eta \sim \mathcal{N}(0, \Gamma)$, we define its covariance matrix as $\Gamma = \sigma_y^2 I$, where we set $\sigma_y$ to either $0.7$ or $1.0$.

\subsection{Lorenz '96}
\label{ssec:L96}

The Lorenz '96 model is a widely-used set of ordinary differential equations that possess features of large-scale atmospheric dynamics \cite{lorenz96}. The model takes the form of a damped-driven linear system, with additional energy conserving quadratic nonlinearity that induces cyclic interactions (waves) among the components of the system; this makes it particularly valuable for studying atmospheric predictability. The system dynamics are governed by the following equations:
\begin{equation}\label{eq:L96_equations}
    \frac{du_i}{dt} = \bigr(u_{i+1} - u_{i-2}\bigl)u_{i-1} - u_i + F,\quad i=1,2,\ldots, d.
\end{equation}
The parameter settings for the Lorenz '96 system are shown in Table~\ref{tab:lorenz96_settings}. The forcing parameter $F=8$ places the Lorenz '96 system in a chaotic regime when $d=40.$

\begin{table}[h]
\centering
\caption{Lorenz '96 System Settings}
\label{tab:lorenz96_settings}
\begin{tabular}{ll}
    \toprule
    \textbf{Category} & \textbf{Values} \\
    \midrule
    Parameters & $F=8$ \\
    \midrule
    States & $v = (u_1,u_2,\ldots,u_{40}) \in \mathbb{R}^d, \quad d = 40$ \\
    \midrule
    Observations & $h(v)=(u_1,u_5, \cdots, u_{33}, u_{37}) \in \bbR^{d_\mathrm{obs}}, \quad d_\mathrm{obs}=10$ \\ 
    \midrule
    Time Step & Observation time step: $\Delta t=0.15$; $5$ RK4 integration steps: $\Delta t/5 = 0.03$ \\
    \bottomrule
\end{tabular}
\end{table}

\begin{figure}[htbp]
\centering
\resizebox{\textwidth}{!}{
\begin{tabular}{ccc}
\parbox[b]{1.6cm}{\centering Relative\\RMSE\\\;\\\;\\\;\\\;}
&\includegraphics[width=0.45\linewidth]{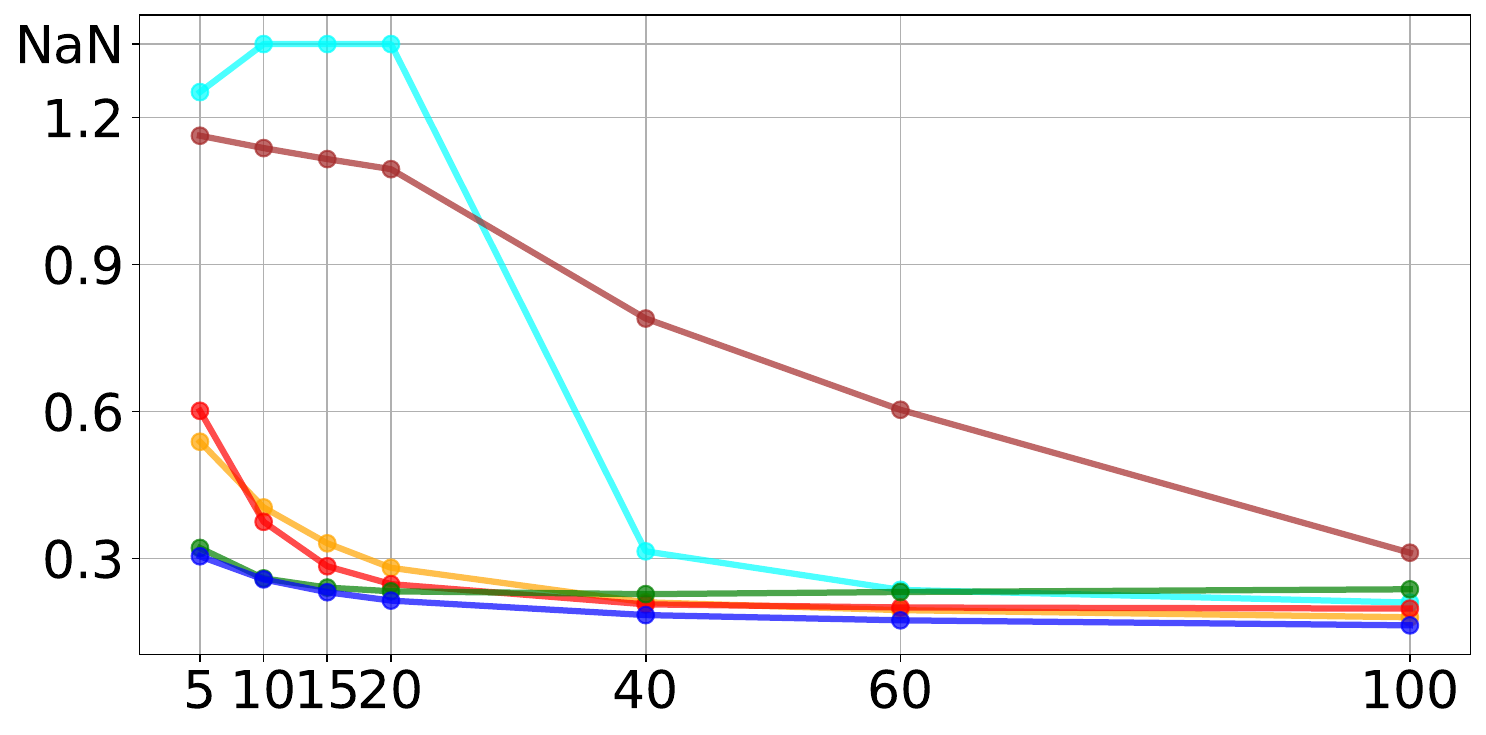}  
&\includegraphics[width=0.45\linewidth]{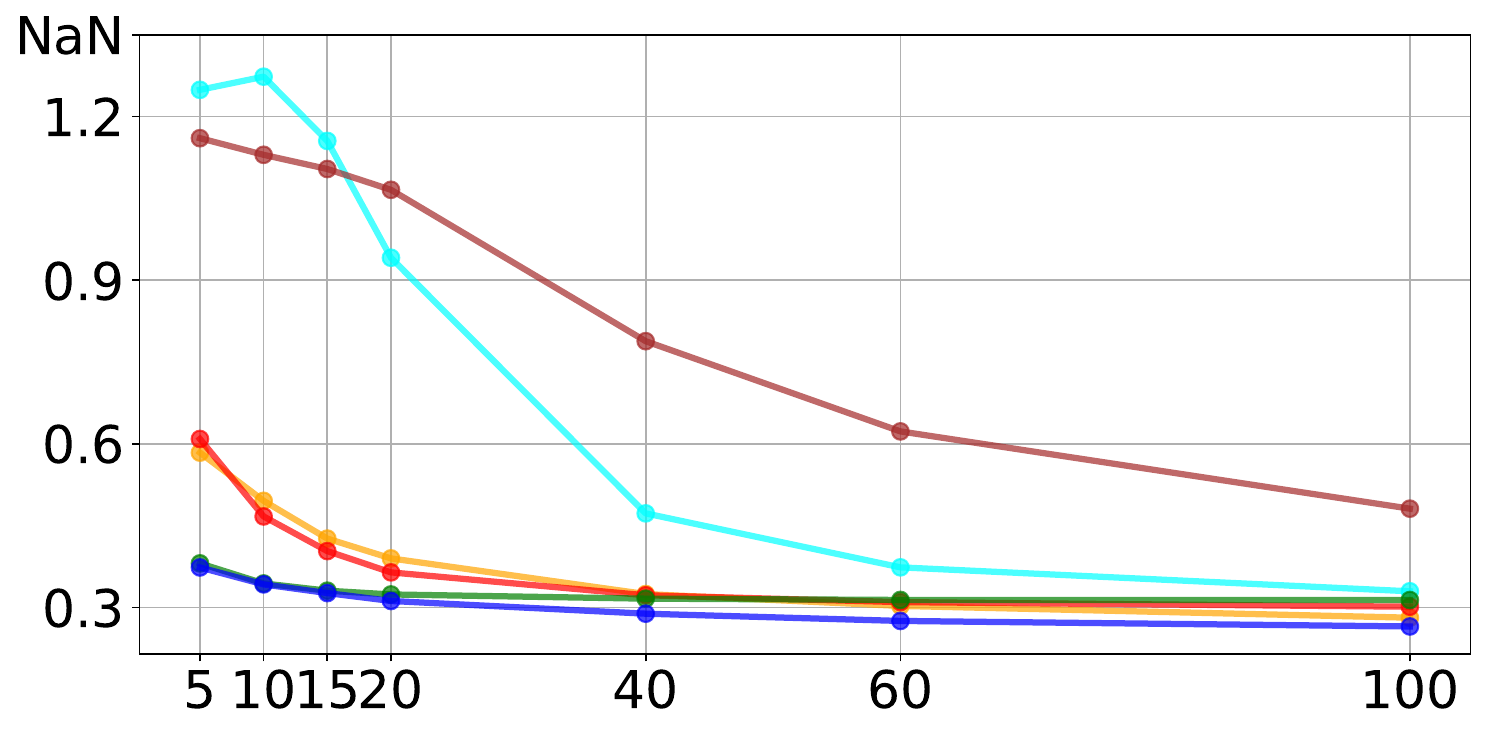}  
\\
\parbox[b]{1.6cm}{\centering Relative\\Improvement\\\;\\\;\\\;\\\;}
&\includegraphics[width=0.45\linewidth]{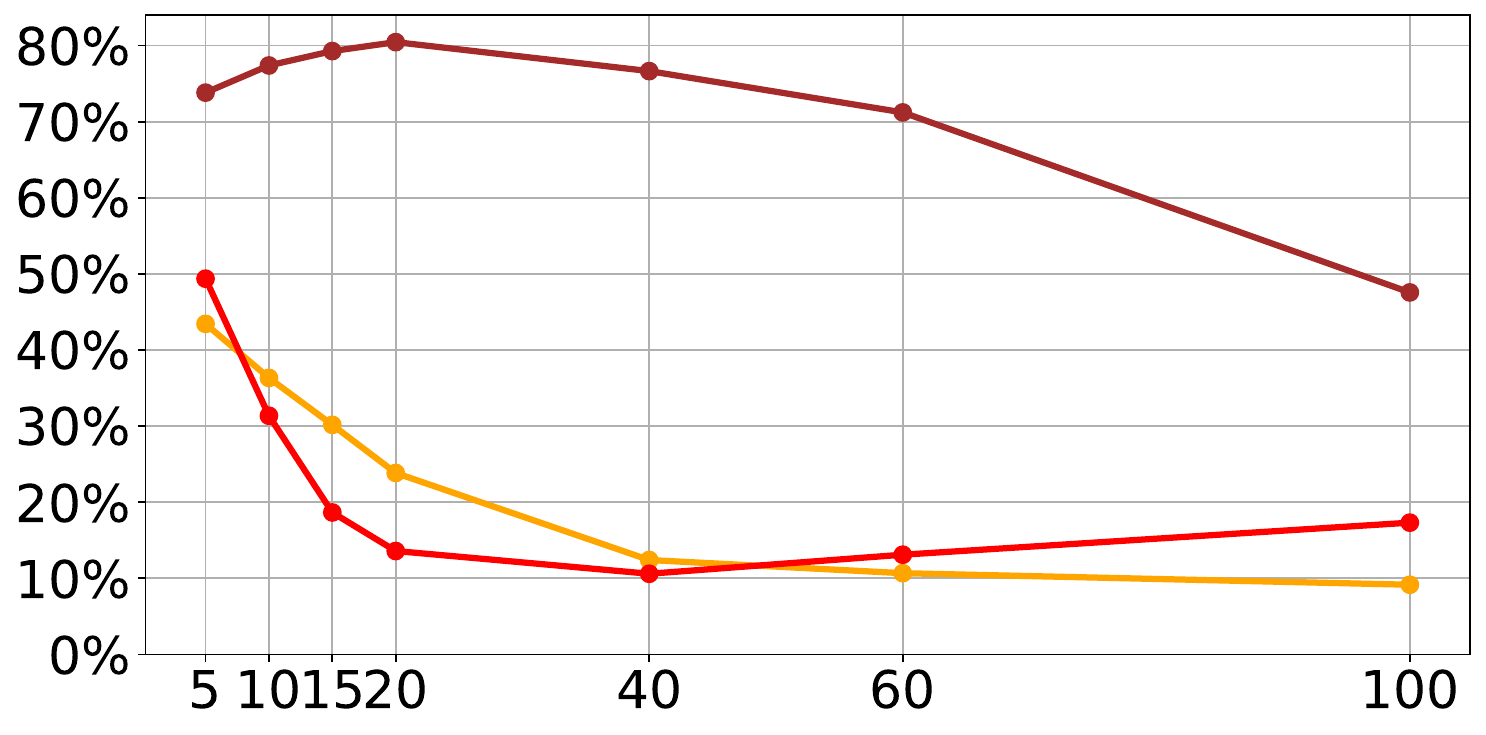}  
&\includegraphics[width=0.45\linewidth]{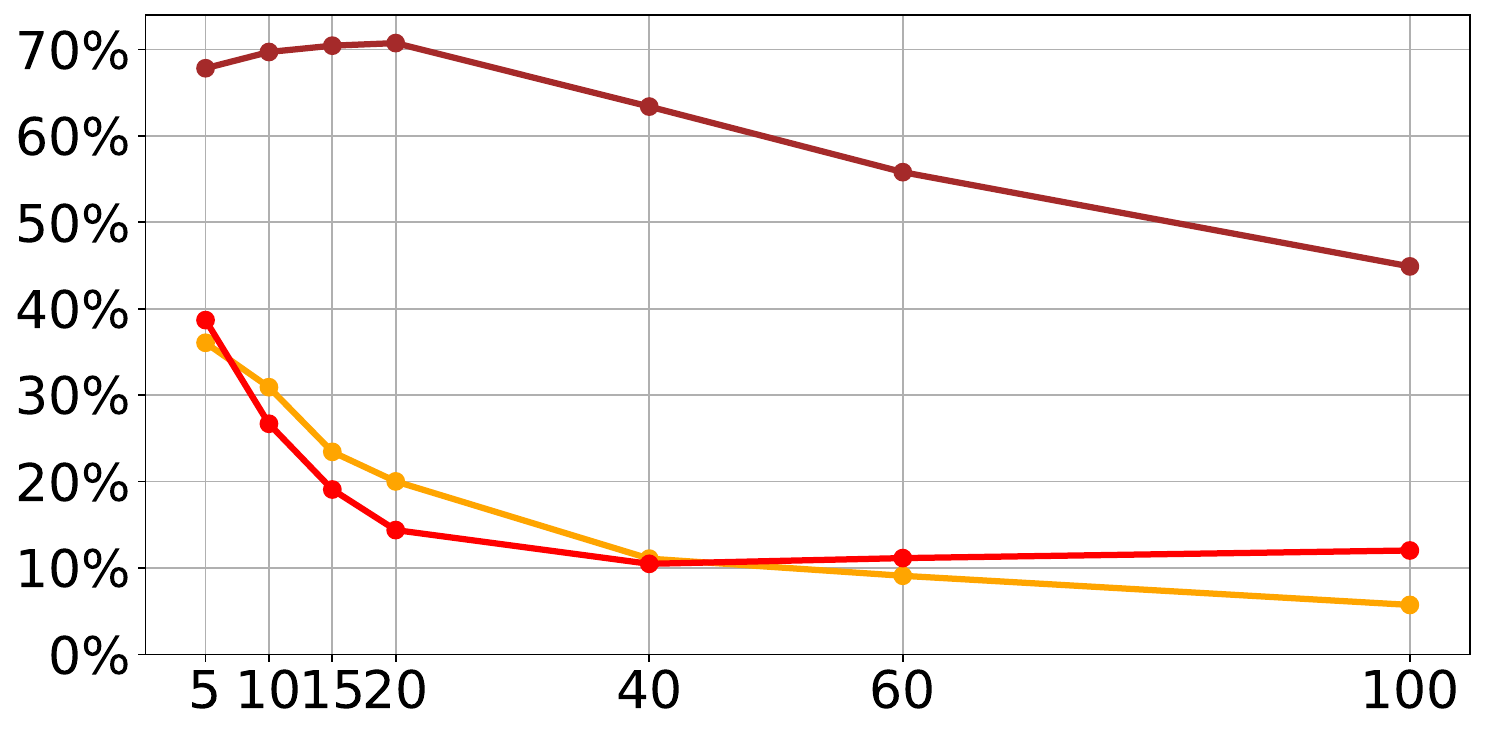}  
\\
\multirow{2}{*}{\textbf{Lorenz '96}}
& Ensemble Size
& Ensemble Size
\\
& $\sigma_y = 0.7$
& $\sigma_y = 1.0$
\\
& \multicolumn{2}{c}{\includegraphics[trim={0 1.5cm 0 1.5cm},clip, width=0.8\linewidth]{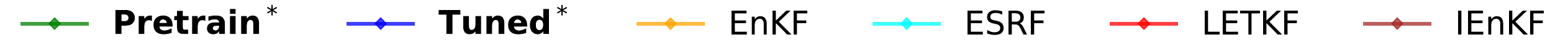}  }
\end{tabular}
}
\vspace{-0.5cm}
\caption{Comparison results on the Lorenz '96 system. The upper row of plots shows direct R-RMSE comparisons between different methods, while the lower row illustrates the relative improvement of our fine-tuning method compared to benchmarks. These comparisons are presented for observation noise levels $\sigma_y=0.7$ (left column) and $\sigma_y=1.0$ (right column). Our proposed MNMEF method is highlighted in the legends with bold font and an asterisk (e.g. \textbf{Pretrain}$^*$). In summary, our fine-tuned MNMEF model consistently outperforms benchmarks, showing substantial improvements for small ensembles and a 15-20\% advantage over LETKF at larger ensemble sizes (eg. 60, 100).}
\label{fig:L96}
\end{figure}

\begin{figure}[htb]
\centering
\resizebox{\textwidth}{!}{
\begin{tabular}{cccc}
     \parbox[b]{1.5cm}{\centering Ground\\ Truth\\\;}
     &\includegraphics[width=0.3\linewidth]{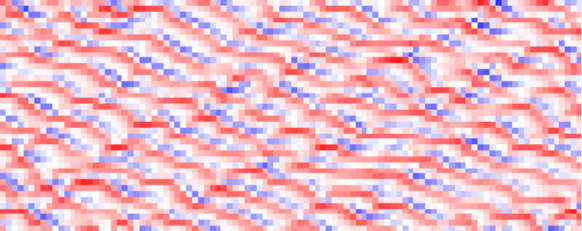}  
     & \parbox[b]{3.4cm}{\centering Observe every 4th \\dimension; $\sigma_y=1.0$\\$\xrightarrow{\hspace{3cm}}$\\\;}
     & \includegraphics[width=0.3\linewidth]{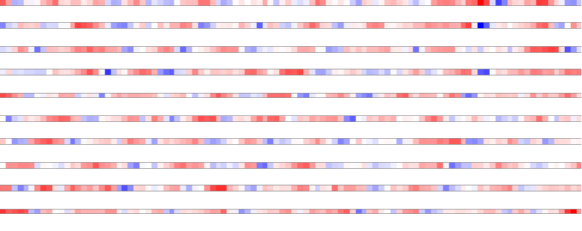}
     \\ 
     &(a) States 
     &
     &(b) Observations
     \\
     \parbox[b]{1.5cm}{\centering MNMEF \\(ours)\\$N=10$\vspace{0.15cm}}
     &\includegraphics[width=0.3\linewidth]{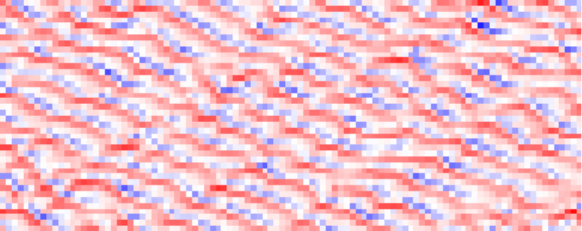}  &\includegraphics[width=0.3\linewidth]{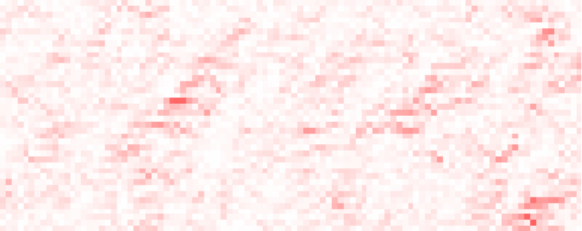}
     &\includegraphics[width=0.3\linewidth]{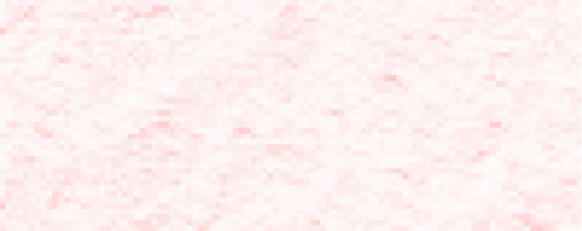}
     \\
     \parbox[b]{1.5cm}{\centering LETKF \\\cite{hunt2007efficient}\\$N=10$\vspace{0.15cm}}
     &\includegraphics[width=0.3\linewidth]{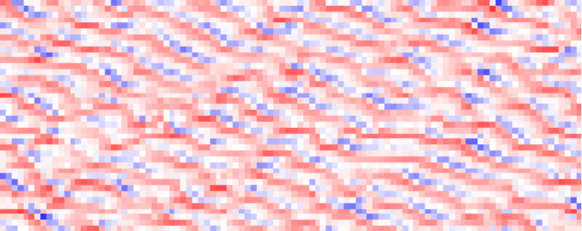}  &\includegraphics[width=0.3\linewidth]{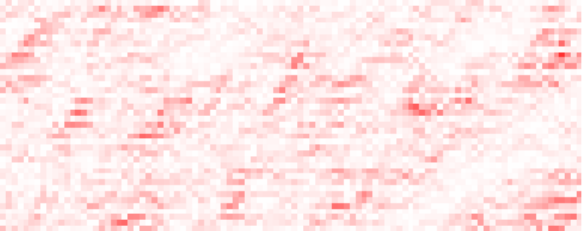}
     &\includegraphics[width=0.3\linewidth]{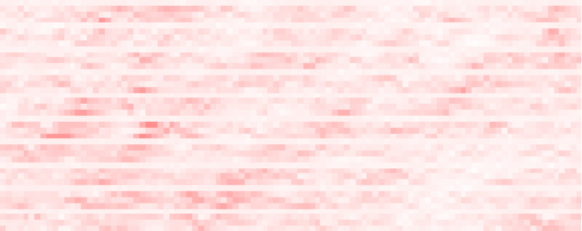}
     \\ 
     &(c) States (mean)
     &(d) Abs Error
     &(e) Spread (std)
     \\
     &\multicolumn{3}{c}{\includegraphics[trim={0 1.1cm 0 0},clip, width=0.6\linewidth]{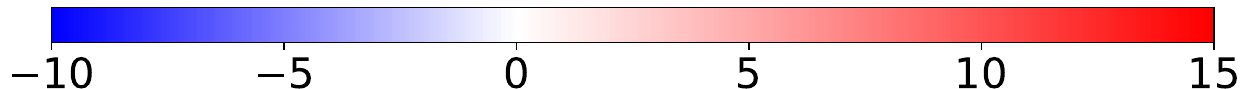}  }
\end{tabular}
}
\vspace{-0.3cm}
\caption{Visualization of one test trajectory (time steps 1401-1500, $\Delta t=0.15$) for Lorenz '96 states (vertical axis) over time (horizontal axis)with the observation noise $\sigma_y=1.0$.  Panel (a): Ground-truth states (unknown). Panel (b): Observations (known, every 4th dimension observed). Rows 2-3 show our method MNMEF pretrained on $N=10$, and the benchmark LETKF with the ensemble size $N=10$. Panel (c): state estimation (ensemble mean), Panel (d): absolute error of mean with respect to the ground truth, and Panel (e): ensemble spread (standard deviation).}
\label{fig:lorenz96_visualization}
\end{figure}

\begin{figure}[h!]
\centering
\resizebox{\textwidth}{!}{
\begin{tabular}{ccc}
     \parbox[b]{1.5cm}{\centering MNMEF \\(ours)\\$N=10$\\\;}
     &\includegraphics[width=0.45\linewidth]{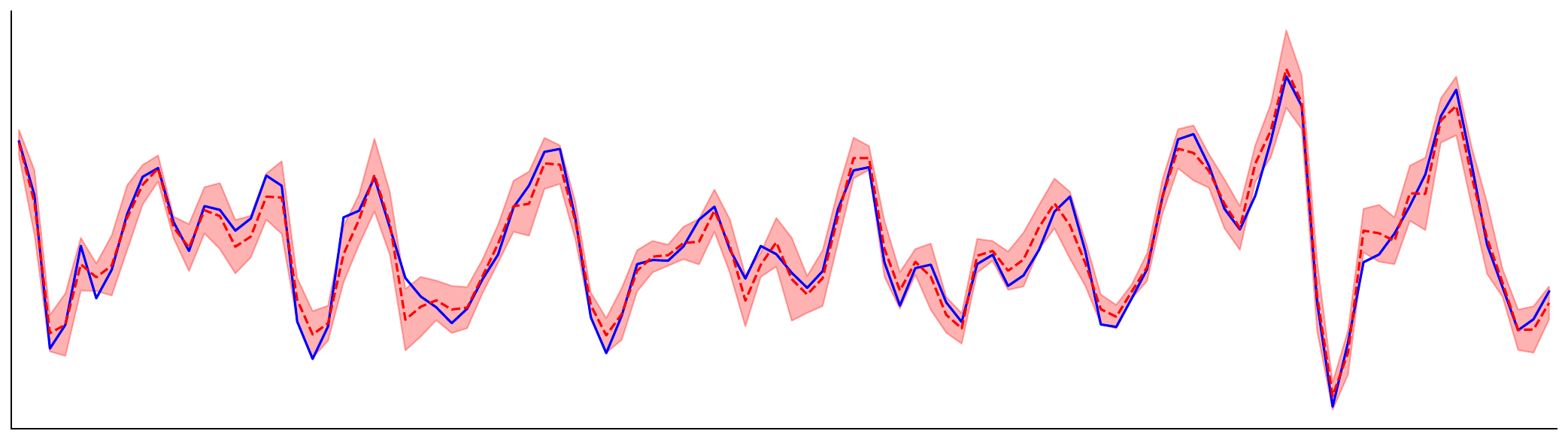}  
     &\includegraphics[width=0.45\linewidth]{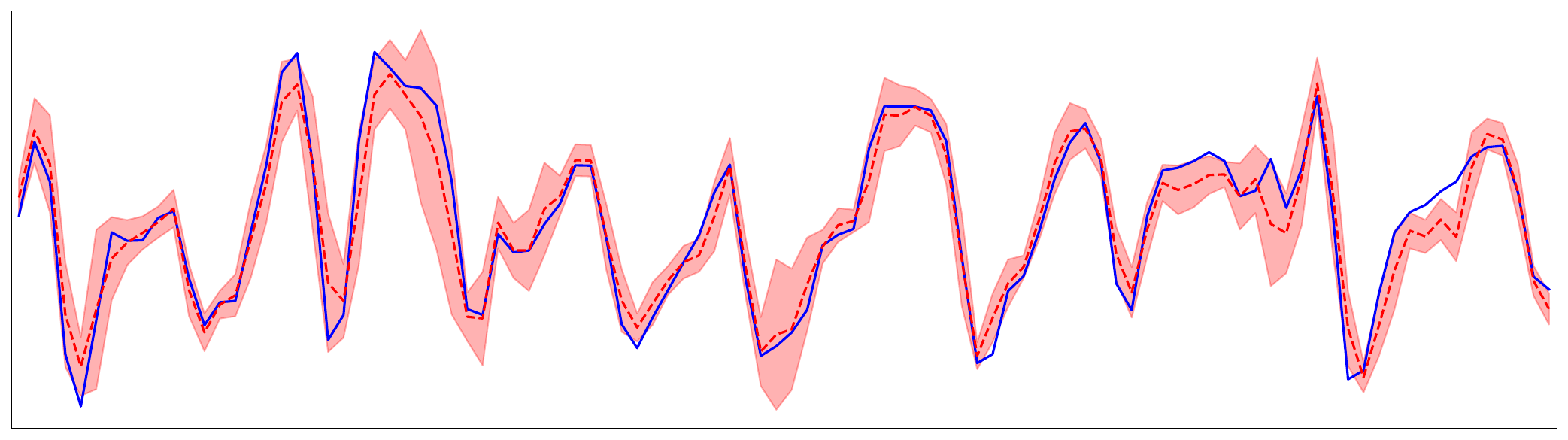}
     \\
     \parbox[b]{1.5cm}{\centering LETKF \\\cite{hunt2007efficient}\\$N=10$\\\;}
     &\includegraphics[width=0.45\linewidth]{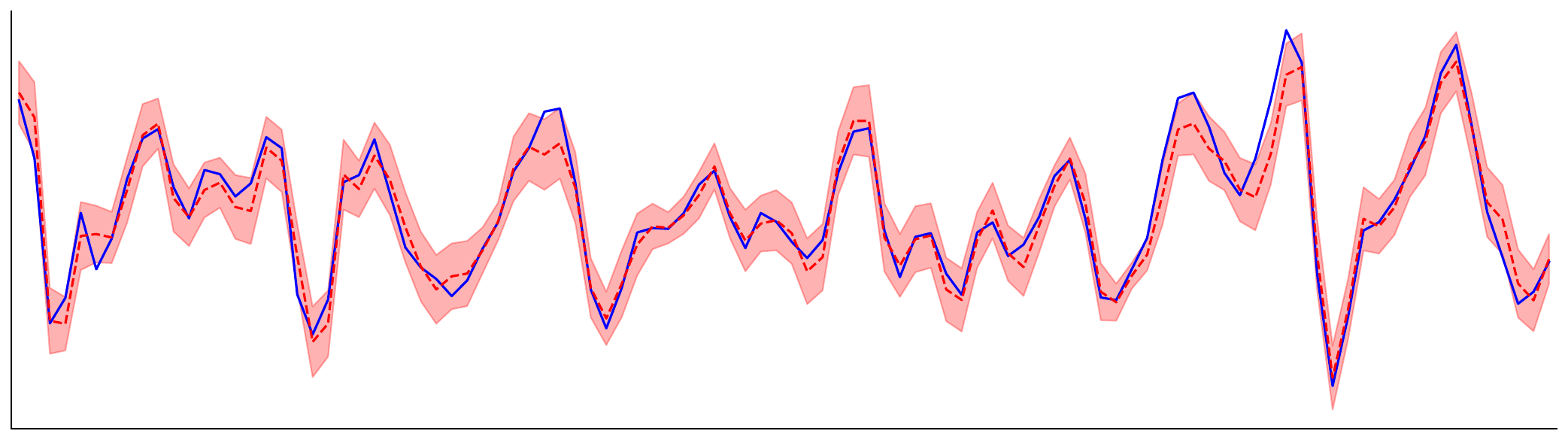}  
     &\includegraphics[width=0.45\linewidth]{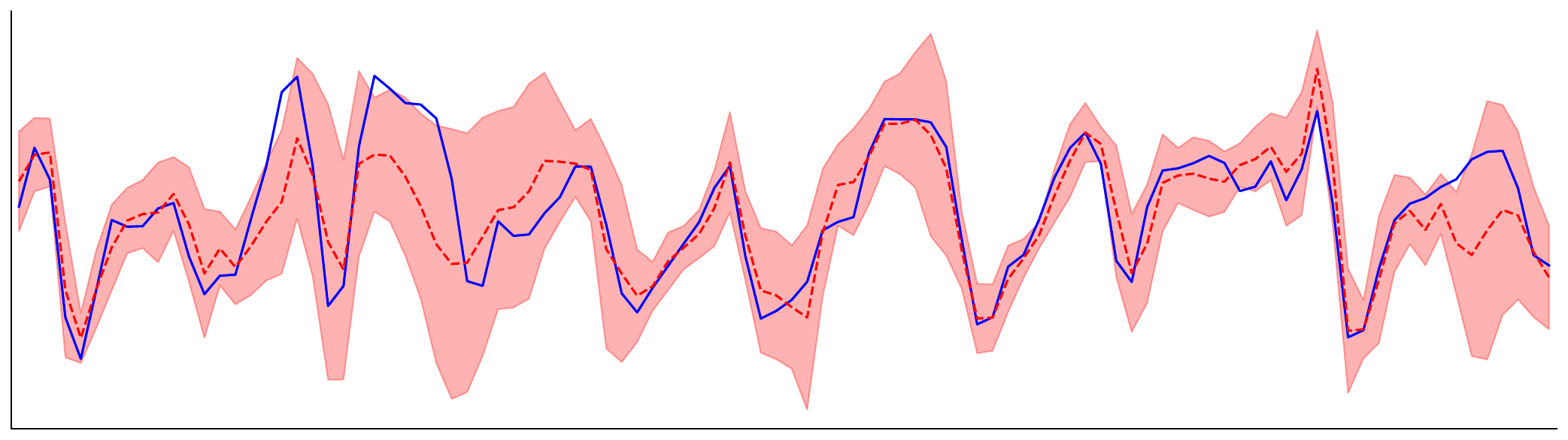}
     \\ 
     &(a) Dimension 1 (Observed)
     &(b) Dimension 2 (Not Observed)
     \\
     &\multicolumn{2}{c}{\includegraphics[trim={0 0 0 0},clip, width=0.7\linewidth]{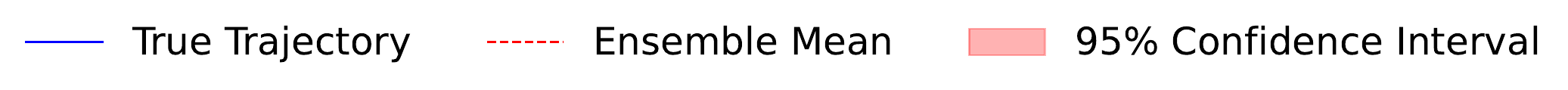}  }
     \vspace{-0.5cm}
\end{tabular}
}
\caption{Visualization of two dimensions (index 1 and 2) in one test trajectory (time steps 1401-1500, $\Delta t=0.15$) for Lorenz '96 state values (vertical axis) over time (horizontal axis) with the ensemble size $N=10$ and the observation noise $\sigma_y=1.0$. Panel (a): Dimension 1, observed. Panel (b): Dimension 2, not observed. The first row is our method MNMEF, pretrained on $N=10$, and the second row is the benchmark LETKF. The 95\% confidence intervals shown in figures are calculated as the ensemble mean $\pm$ 1.96 $\times$ the ensemble standard deviation. In summary, MNMEF performs significantly better on the unobserved dimension and comparably to LETKF on the observed dimension; meanwhile, MNMEF maintains an appropriate spread without suffering from filter degeneracy. }
\label{fig:lorenz96_dim01}
\end{figure}

In Figure~\ref{fig:L96}, we compare the performance of our pretrained MNMEF and fine-tuned MNMEF against four benchmark methods: EnKF, ESRF, LETKF, and IEnKF, on the Lorenz '96 system. The upper row of plots in Figure~\ref{fig:L96} displays the R-RMSE \eqref{eq:avg_r_rmse} for observation noise levels $\sigma_y = 0.7$ (left plot) and $\sigma_y=1.0$ (right plot). The lower row of plots in Figure~\ref{fig:L96} shows the relative improvement $\mathcal{E}_\mathrm{RI}$ \eqref{eq:relative_imp} of our fine-tuned method as compared to EnKF, IEnKF, and LETKF for these respective noise levels.

We observe that our fine-tuned model consistently outperforms all benchmarks.
Indeed, our fine-tuned method achieves substantial improvement for small ensemble sizes and maintains a notable advantage at larger ensemble sizes, outperforming LETKF by approximately 15--20\%.
Our pretrained model, trained with $N=10$ ensembles also behaves competitively, but performs slightly worse than LETKF for large ensemble size $40,60,100$.

To further illustrate the performance differences between our MNMEF and the best-performance benchmark LETKF, we provide visualizations in Figure~\ref{fig:lorenz96_visualization} and Figure~\ref{fig:lorenz96_dim01} for $N=10$ and $\sigma_y=1.0$. The visualization is on the last $100$ time steps (from $1401$ to $1500$) from a test trajectory of length  $1500$ and $\Delta t = 0.15$. In both figures, our MNMEF approach is the one pretrained with the ensemble size $N=10$.

Figure~\ref{fig:lorenz96_visualization} presents a comparison of the estimated state trajectory. Panel (a) shows the ground truth, and panel (b) shows the sparse observations. The subsequent rows compare MNMEF and LETKF. Visually,  MNMEF displays smaller magnitudes for the absolute error (panel (d)). In addition, the ensemble spread (panel (e)) for MNMEF is more consistent than LETKF. Figure~\ref{fig:lorenz96_dim01} focuses on the estimation of two specific dimensions: one observed (Dimension 1, panel (a)) and one unobserved (Dimension 2, panel (b)). For the observed dimension, both MNMEF and LETKF perform comparably well, with their ensemble means closely following the ground truth. However, for the unobserved dimension, MNMEF demonstrates a clear advantage in state estimation. Furthermore, MNMEF maintains an appropriate ensemble spread while avoiding  filter degeneracy. 

\subsection{Kuramoto--Sivashinsky}
\label{ssec:KS}

The Kuramoto–-Sivashinsky (KS) equation \cite{kuramoto1978diffusion, michelson1977nonlinear} is a widely studied nonlinear partial differential equation that describes the evolution of instabilities in spatially extended systems, capturing complex spatiotemporal chaotic dynamics arising in flame fronts, thin fluid films, and reaction-diffusion systems. Due to its chaotic behavior and sensitivity to initial conditions, the KS equation serves as a valuable testbed for developing and evaluating data assimilation algorithms. With the periodicity in space imposed to identity $x=L$ and $x=0$, the one-dimensional KS equation for $u: [0,L] \times \mathbb{R}^+ \to \mathbb{R}$ takes the form
\begin{subequations}\label{eq:KS_equation}
\begin{align}
    \frac{\partial u}{\partial t} + \frac{\partial^4 u}{\partial x^4} + \frac{\partial^2 u}{\partial x^2} + u\frac{\partial u}{\partial x} &= 0, \quad
    (x,t) \in (0,L) \times \bbR^+, \\[1ex]
    u(x,0) & = u_0, \quad \forall x \in [0,L].
    \end{align}
\end{subequations}

\begin{table}[h]
\centering
\caption{Kuramoto-–Sivashinsky (KS) System Settings}
\label{tab:ks_settings}
\begin{tabular}{ll}
    \toprule
    \textbf{Category} & \textbf{Values} \\
    \midrule
    Parameters & $L=32\pi$ \\
    \midrule
    \multirow{2}{*}{States} & $x_j=jL/128,\quad j=0,1,\ldots,127$ \\
    & $v=\bigr(u(x_0),u(x_1),\ldots,u(x_{127}\bigl))\in\bbR^d, \quad d=128$ \\
    \midrule
    Observations & $h(v)=\bigl(u(x_0), u(x_8), \cdots, u(x_{120})\bigr) \in \bbR^{d_\mathrm{obs}},\quad d_\mathrm{obs}=16$ \\ \midrule
    Time Step & Observation time step: $\Delta t=1$;  $4$ ETDRK4 integration steps: $\Delta t/4 = 0.25$ \\
    \bottomrule
\end{tabular}
\end{table}

\begin{figure}[htbp]
\centering
\resizebox{\textwidth}{!}{
\begin{tabular}{ccc}
\parbox[b]{1.6cm}{\centering Relative\\RMSE\\\;\\\;\\\;\\\;}
&\includegraphics[width=0.45\linewidth]{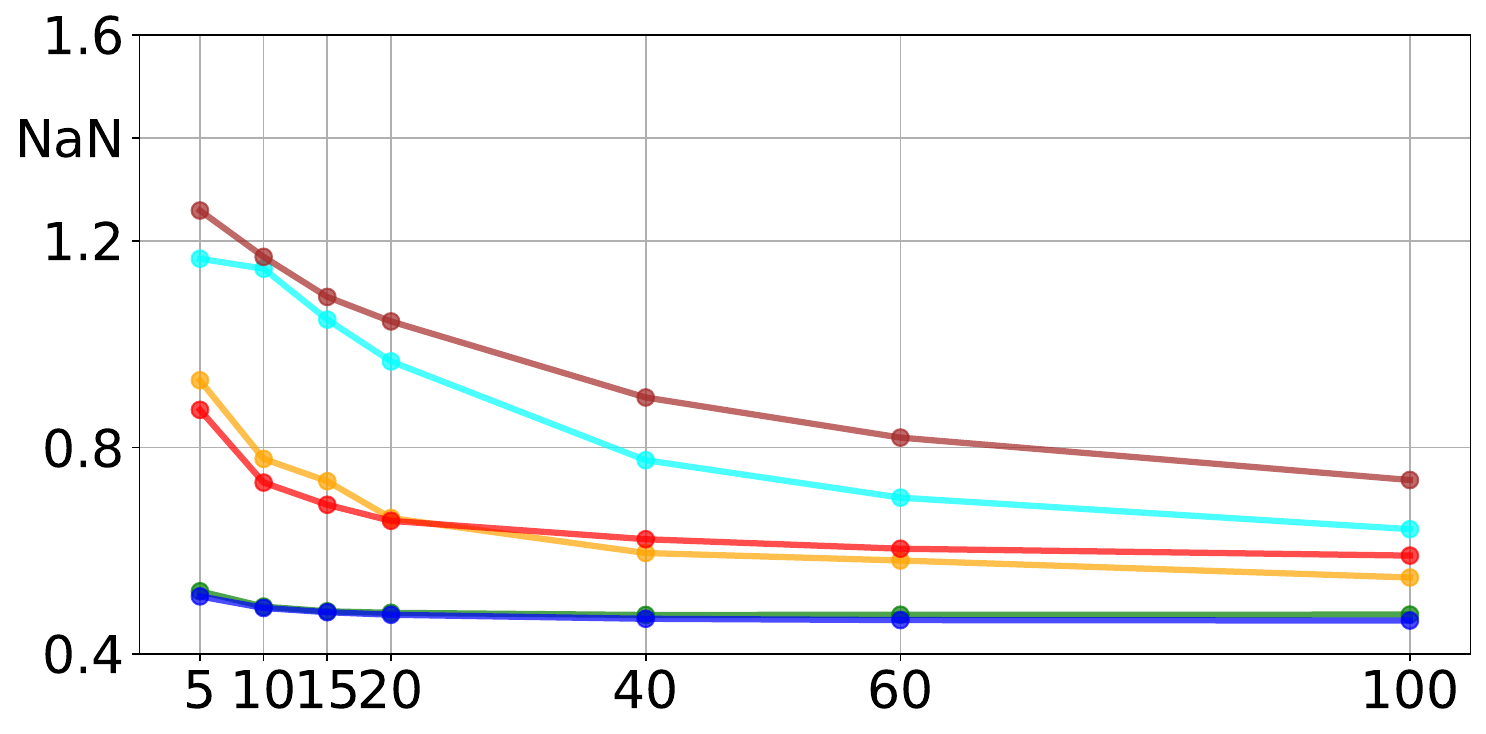}  
&\includegraphics[width=0.45\linewidth]{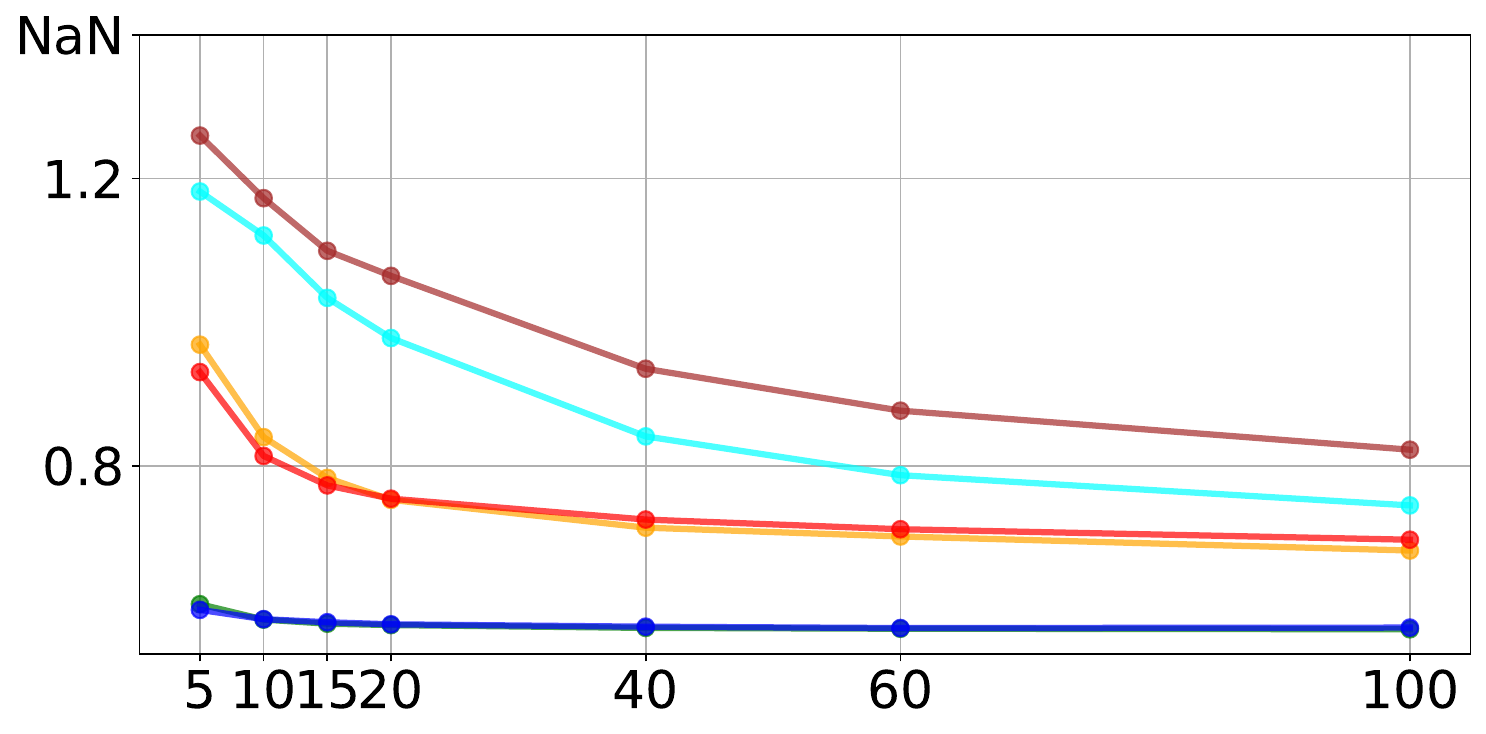}  
\\
\parbox[b]{1.6cm}{\centering Relative\\Improvement\\\;\\\;\\\;\\\;}
&\includegraphics[width=0.45\linewidth]{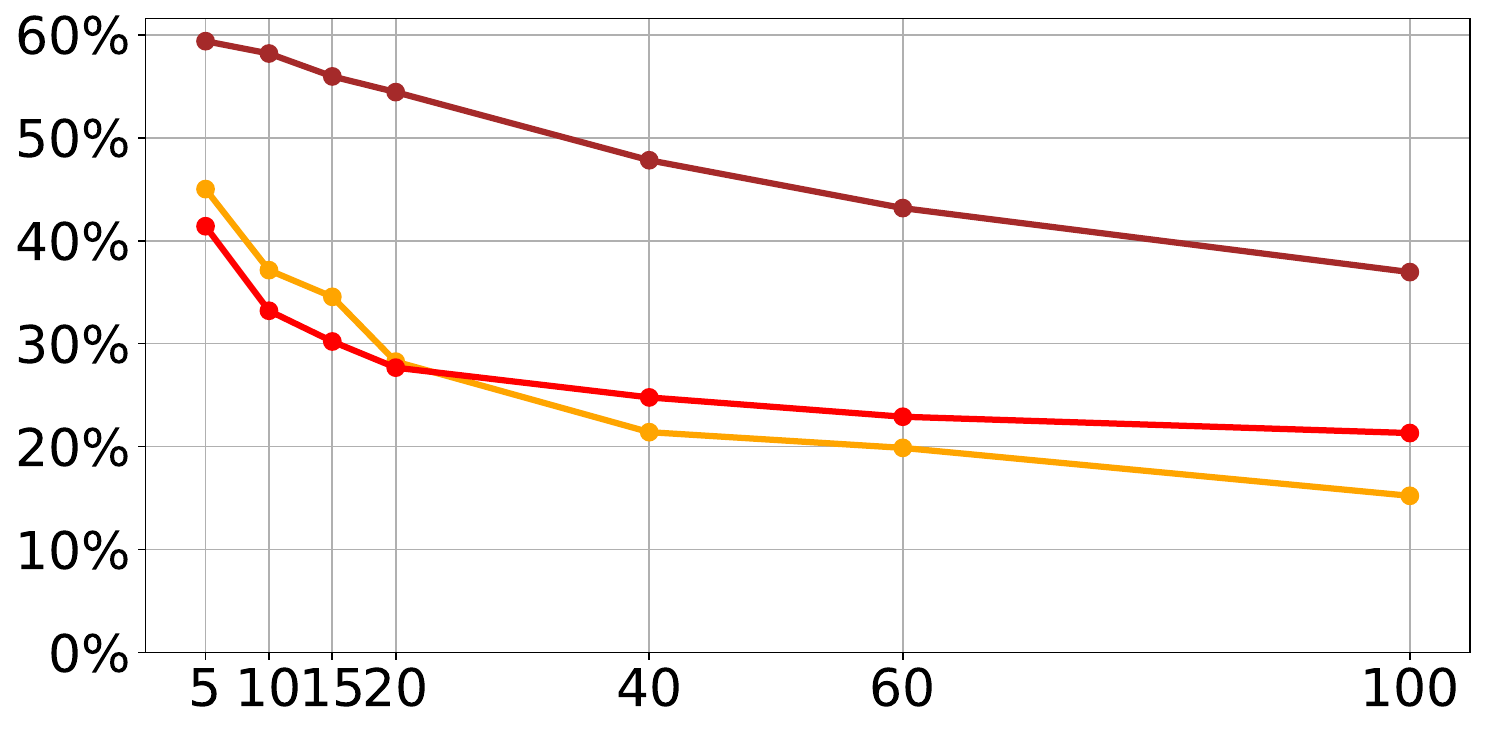}  
&\includegraphics[width=0.45\linewidth]{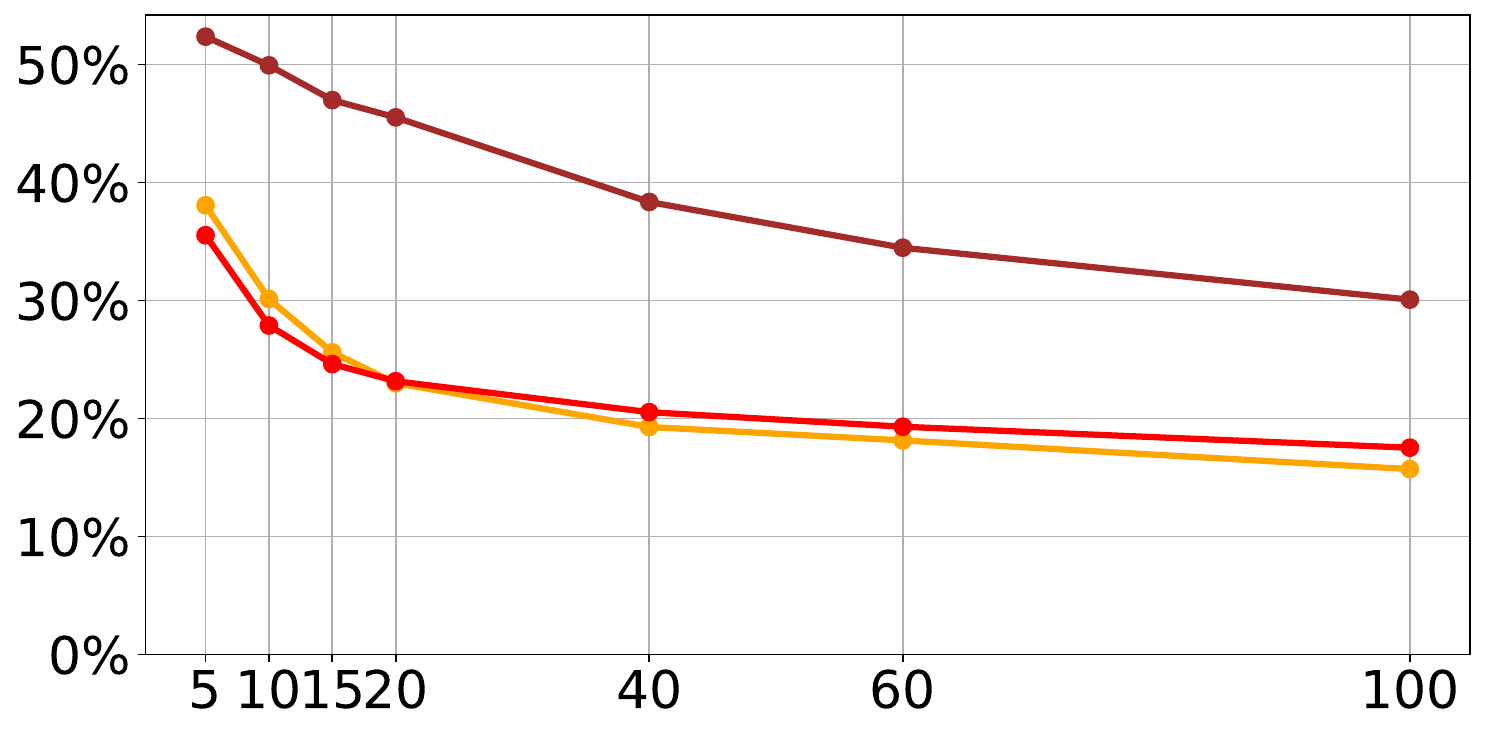}  
\\
\textbf{Kuramoto--}
& Ensemble Size
& Ensemble Size
\\
\textbf{Sivashinsky}
& $\sigma_y = 0.7$
& $\sigma_y = 1.0$
\\
& \multicolumn{2}{c}{\includegraphics[trim={0 1.5cm 0 1.5cm},clip, width=0.8\linewidth]{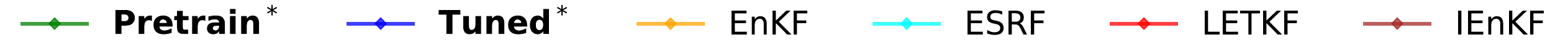}  }
\end{tabular}
}
\vspace{-0.5cm}
\caption{Comparison results on the Kuramoto–-Sivashinsky (KS) system. The upper row of plots shows direct R-RMSE comparisons between different methods, while the lower row illustrates the relative improvement of our fine-tuning method compared to benchmarks. These comparisons are presented for observation noise levels $\sigma_y=0.7$ (left column) and $\sigma_y=1.0$ (right column). Our proposed MNMEF method is highlighted in the legends with bold font and an asterisk (e.g. \textbf{Pretrain}$^*$). In summary, our fine-tuned MNMEF model consistently outperforms benchmarks, showing substantial improvements for small ensembles and around 20\% advantage over LETKF at larger sizes (eg. 60, 100).}
\label{fig:ks}
\end{figure}

\begin{figure}[h!]
\centering
\resizebox{\textwidth}{!}{
\begin{tabular}{cccc}
     \parbox[b]{1.5cm}{\centering Ground\\ Truth\\\;}
     &\includegraphics[width=0.3\linewidth]{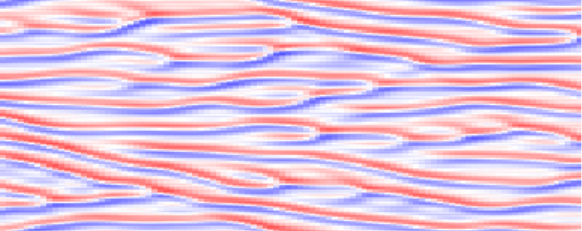}  
     & \parbox[b]{3.4cm}{\centering Observe every 8th \\dimension; $\sigma_y=1.0$\\$\xrightarrow{\hspace{3cm}}$}
     & \includegraphics[width=0.3\linewidth]{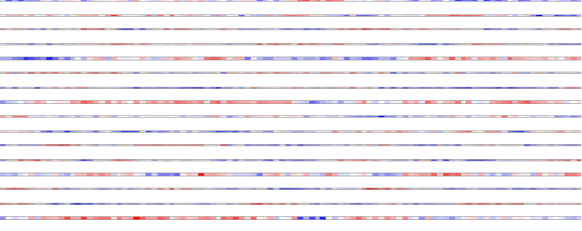}
     \\ 
     &(a) States 
     &
     &(b) Observations
     \\
     \parbox[b]{1.5cm}{\centering MNMEF \\(ours)\\$N=10$\vspace{0.15cm}}
     &\includegraphics[width=0.3\linewidth]{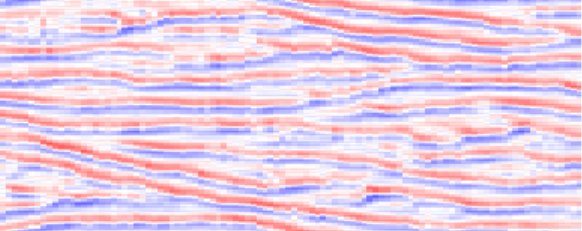}  &\includegraphics[width=0.3\linewidth]{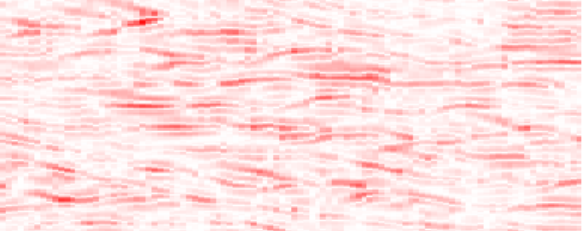}
     &\includegraphics[width=0.3\linewidth]{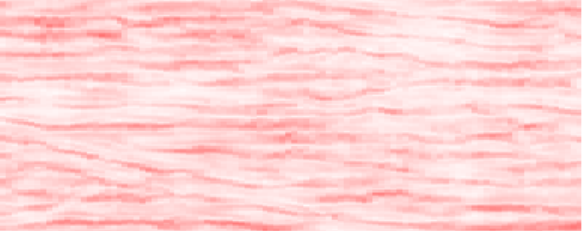}
     \\
     \parbox[b]{1.5cm}{\centering LETKF \\\cite{hunt2007efficient}\\$N=10$\vspace{0.15cm}}
     &\includegraphics[width=0.3\linewidth]{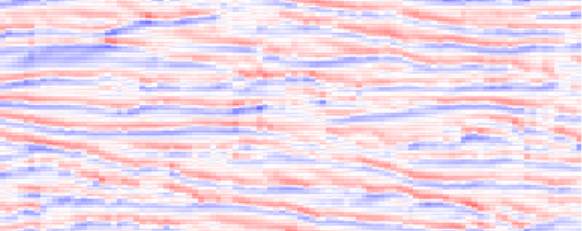}  &\includegraphics[width=0.3\linewidth]{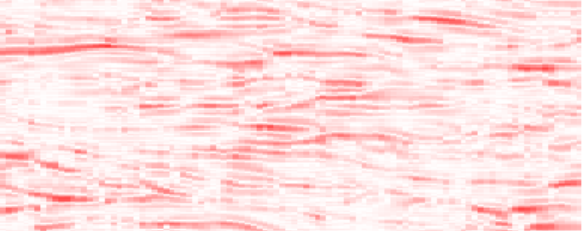}
     &\includegraphics[width=0.3\linewidth]{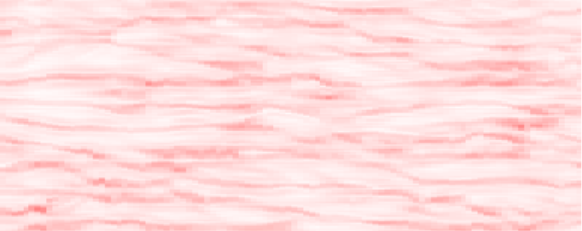}
     \\ 
     &(c) States (mean)
     &(d) Abs Error
     &(e) Spread (std)
     \\
     &\multicolumn{3}{c}{\includegraphics[trim={0 1.1cm 0 0},clip, width=0.6\linewidth]{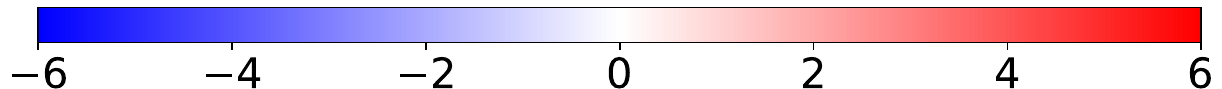}  }
\end{tabular}
}
\vspace{-0.3cm}
\caption{Visualization of one test trajectory (time steps 1901-2000, $\Delta t=1$) for Kuramoto--Sivashinshy (KS) states (vertical axis) over time (horizontal axis) with the observation noise $\sigma_y=1.0$.  Panel (a): Ground-truth states (unknown). Panel (b): Observations (known, every 8th dimension observed). Rows 2-3 show  our method MNMEF pretrained on $N=10$, and the benchmark LETKF with the ensemble size $N=10$. Panel (c): state estimation (ensemble mean), Panel (d): absolute error with ground truth, and Panel (e): ensemble spread (standard deviation). 
\label{fig:ks_visualization}}
\end{figure}

\begin{figure}[h!]
\centering
\resizebox{\textwidth}{!}{
\begin{tabular}{ccc}
     \parbox[b]{1.5cm}{\centering MNMEF \\(ours)\\$N=10$\\\;}
     &\includegraphics[width=0.45\linewidth]{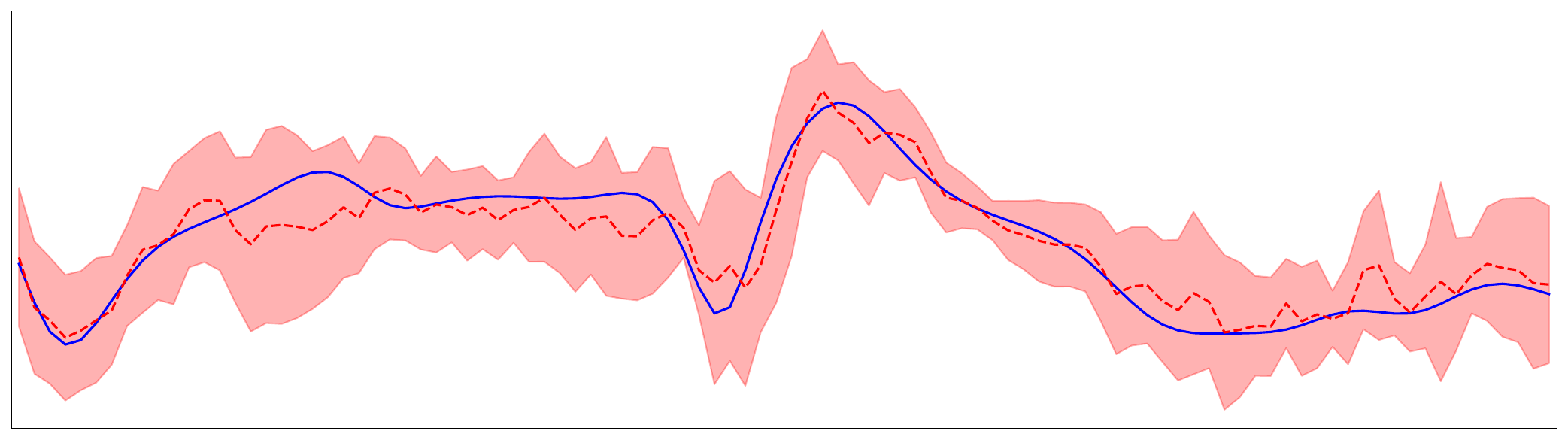}  
     &\includegraphics[width=0.45\linewidth]{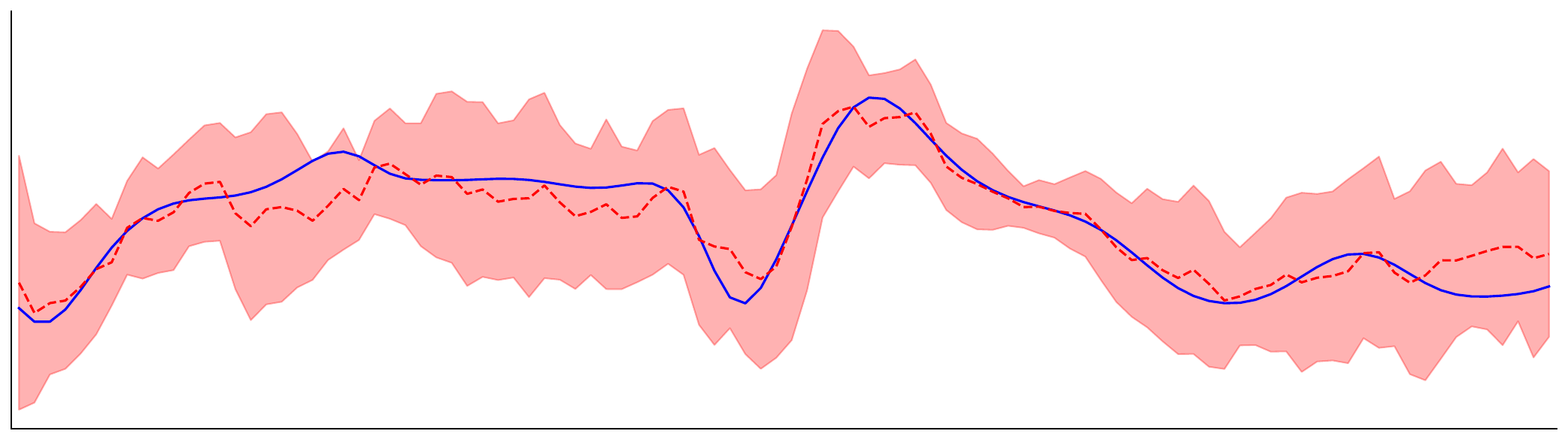}
     \\
     \parbox[b]{1.5cm}{\centering LETKF \\\cite{hunt2007efficient}\\$N=10$\\\;}
     &\includegraphics[width=0.45\linewidth]{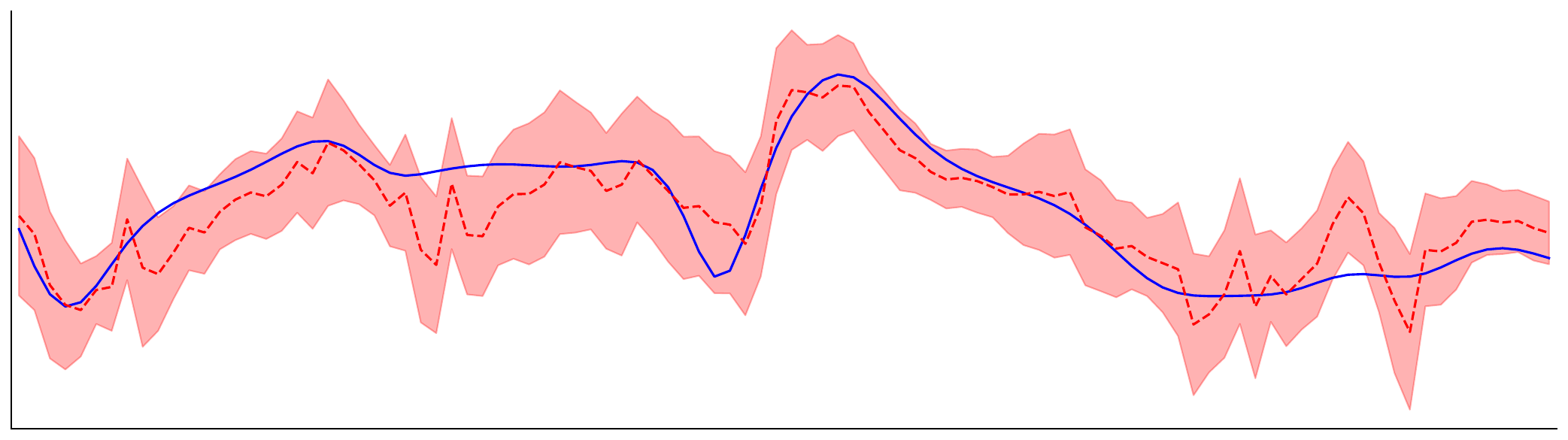}  
     &\includegraphics[width=0.45\linewidth]{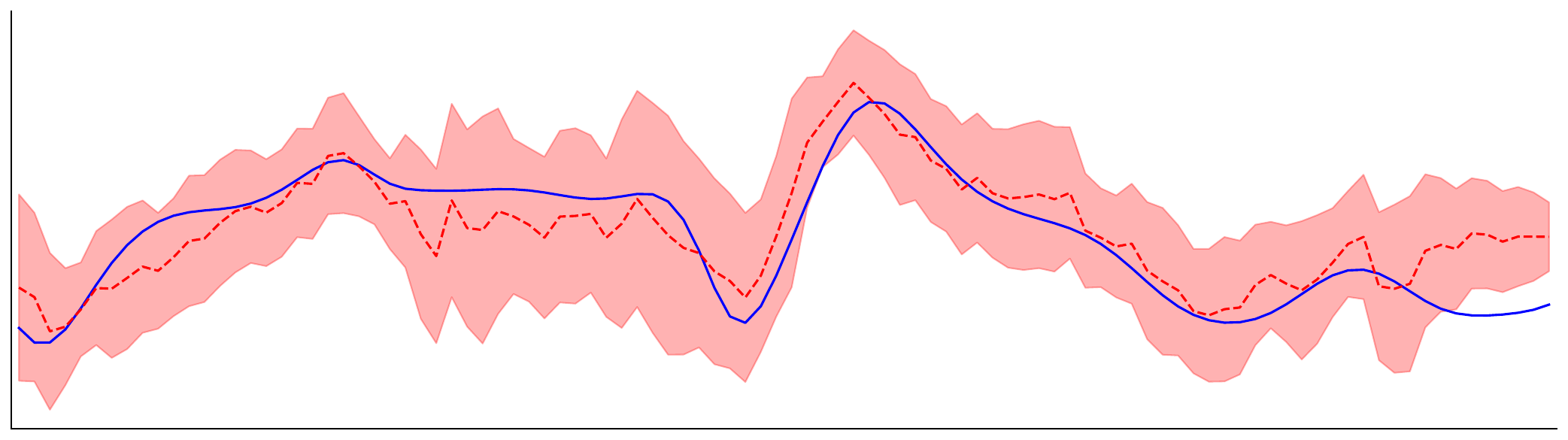}
     \\ 
     &(a) Dimension 1 (Observed)
     &(b) Dimension 2 (Not Observed)
     \\
     &\multicolumn{2}{c}{\includegraphics[trim={0 0 0 0},clip, width=0.7\linewidth]{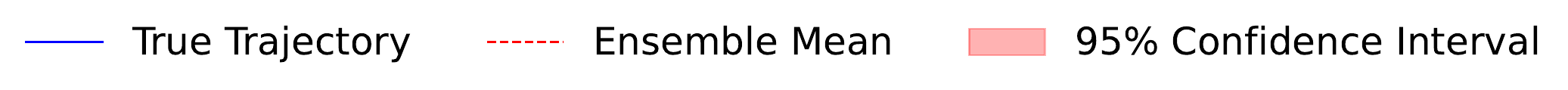}  }
     \vspace{-0.5cm}
\end{tabular}
}
\caption{Visualization of two dimensions (index 1 and 2) in one test trajectory (time steps 1901-2000, $\Delta t=1$) for Kuramoto--Sivashinshy (KS) state values (vertical axis) over time (horizontal axis) with the ensemble size $N=10$ and the observation noise $\sigma_y=1.0$. Panel (a): Dimension 1, observed. Panel (b): Dimension 2, not observed. The first row is our method MNMEF, pretrained on $N=10$, and the second row is the benchmark LETKF. The 95\% confidence intervals shown in figures are calculated as the ensemble mean $\pm$ 1.96 $\times$ the ensemble standard deviation. In summary, MNMEF performs slightly better on both the observed and the unobserved dimensions; meanwhile, MNMEF maintains an appropriate spread without suffering from filter degeneracy. }
\label{fig:ks_dim01}
\end{figure}

The detailed experimental settings and parameters for the KS system are summarized in Table~\ref{tab:ks_settings}. 
Numerically, the KS equation~\eqref{eq:KS_equation} is discretized spatially using a Fourier pseudo-spectral method to handle periodic boundary conditions effectively. Time integration is performed with the exponential time-differencing fourth-order Runge–Kutta (ETDRK4) scheme \cite{kassam2005fourth}, which advances the stiff linear operator analytically and is therefore not subject to the explicit linear Courant–Friedrichs–Lewy (CFL) restriction ($\Delta t\le C\Delta x^{4}$) that would apply to fully explicit RK methods; the time step is chosen based on accuracy and nonlinear resolution rather than linear stability.

In Figure~\ref{fig:ks}, we compare the performance of our pretrained model and fine-tuned model against four benchmark methods: EnKF, ESRF, LETKF, and IEnKF, for the KS system with $\sigma_y = 0.7$ and $1.0$, and provide the relative improvement of our fine-tuned method as compared to EnKF, IEnKF, and LETKF.

The comparison between our pretrained MNMEF and fine-tuned MNMEF methods and the benchmark methods, for the KS dynamical system, slightly differs from what we observed under the Lorenz '96 dynamics. In the KS setting, fine-tuning provides minimal additional benefit, with both our pretrained and fine-tuned MNMEF models performing nearly identically. Both variants significantly outperform benchmark methods across all ensemble sizes. The advantage is particularly pronounced at smaller ensemble sizes. Even at the largest ensemble size ($N=100$), both our pretrained and fine-tuned methods still maintain approximately 20\% improvement over LETKF, the benchmark method with the best performance.

Similarly to the Lorenz '96 discussion in Subsection~\ref{ssec:L96}, we provide visualizations in Figure~\ref{fig:lorenz96_visualization} and Figure~\ref{fig:lorenz96_dim01} for $N=10$ and $\sigma_y=1.0$ to further illustrate the performance differences between our MNMEF and the best-performance benchmark LETKF on the KS system. The visualization is on the last $100$ time steps (from $1901$ to $2000$) of a test trajectory with length  $2000$ and $\Delta t = 1$.  In both figures, our MNMEF is the one pretrained with the ensemble size $N=10$.

Figure~\ref{fig:ks_visualization} presents a comparison of the estimated state trajectory. Figure~\ref{fig:ks_dim01} visualizes the estimation of two specific dimensions: one observed (Dimension 1, panel (a)) and one unobserved (Dimension 2, panel (b)). In summary, our MNMEF performs slightly better than LETKF on both the observed and the unobserved dimensions, which is verified by the R-RMSE quantitative comparison in Figure~\ref{fig:ks}. MNMEF maintains an appropriate ensemble spread, similarly to the LETKF. 

\begin{remark}[Implementation of Localization]
It is worth noting that methods other than LETKF could potentially accommodate localization techniques. For example, LETKF is essentially the ESRF combined with localization. However, since our implementation is based on the DAPPER package, which only implements localization for LETKF, we do not consider localization for the other four methods. Our results on Lorenz '96 and KS problems sufficiently demonstrate the advantages of our approach over other methods, even without localization. For larger ensemble sizes (e.g., $N=60,100$), where the benefits of localization diminish, our method still significantly outperforms the alternatives, like IEnKF.
\end{remark}

\subsection{Lorenz '63}
\label{ssec:L63}

The Lorenz '63 system is a simplified mathematical model for atmospheric convection \cite{lorenz63}, proposed prior to the Lorenz '96 model discussed in Subsection \ref{ssec:L96}. It represents a highly simplified version of atmospheric dynamics with only three dimensions, described by the following set of ordinary differential equations:
\begin{subequations}\label{eq:L63_dynamic}
\begin{align} 
\frac{dx}{dt} &= \sigma (y - x), \\ 
\frac{dy}{dt} &= x (\rho - z) - y, \\ 
\frac{dz}{dt} &= xy - \beta z. 
\end{align} 
\end{subequations}

\begin{table}[h]
\centering
\caption{Lorenz '63 System Settings}
\label{tab:lorenz63_settings}
\begin{tabular}{ll}
    \toprule
    \textbf{Category} & \textbf{Values} \\
    \midrule
    Parameters & $\sigma = 10, \quad \rho = 28, \quad \beta = \frac{8}{3}$ \\
    \midrule
    States & $v = (x, y, z) \in \mathbb{R}^d, \quad d = 3$ \\
    \midrule
    Observations & $h(v) = x \in \mathbb{R}^{d_{\mathrm{obs}}}, \quad d_{\mathrm{obs}} = 1$ \\ 
    \midrule
    Time Step & Observation time step: $\Delta t=0.15$; $5$ RK4 integration steps: $\Delta t/5 = 0.03$ \\
    \bottomrule
\end{tabular}
\end{table}

\begin{figure}[h]
\centering
\resizebox{\textwidth}{!}{
\begin{tabular}{ccc}
\parbox[b]{1.6cm}{\centering Relative\\RMSE\\\;\\\;\\\;\\\;}
&\includegraphics[width=0.45\linewidth]{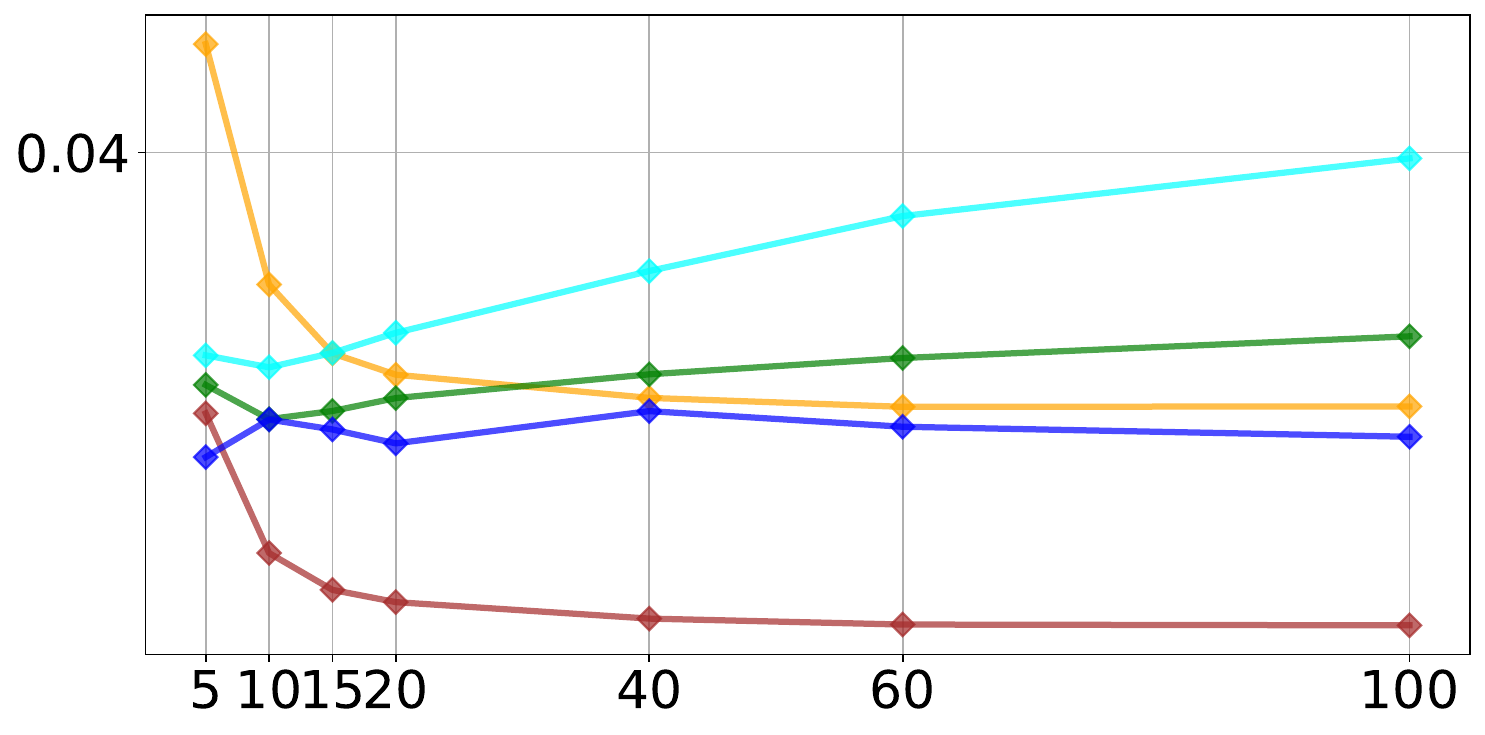}  
&\includegraphics[width=0.45\linewidth]{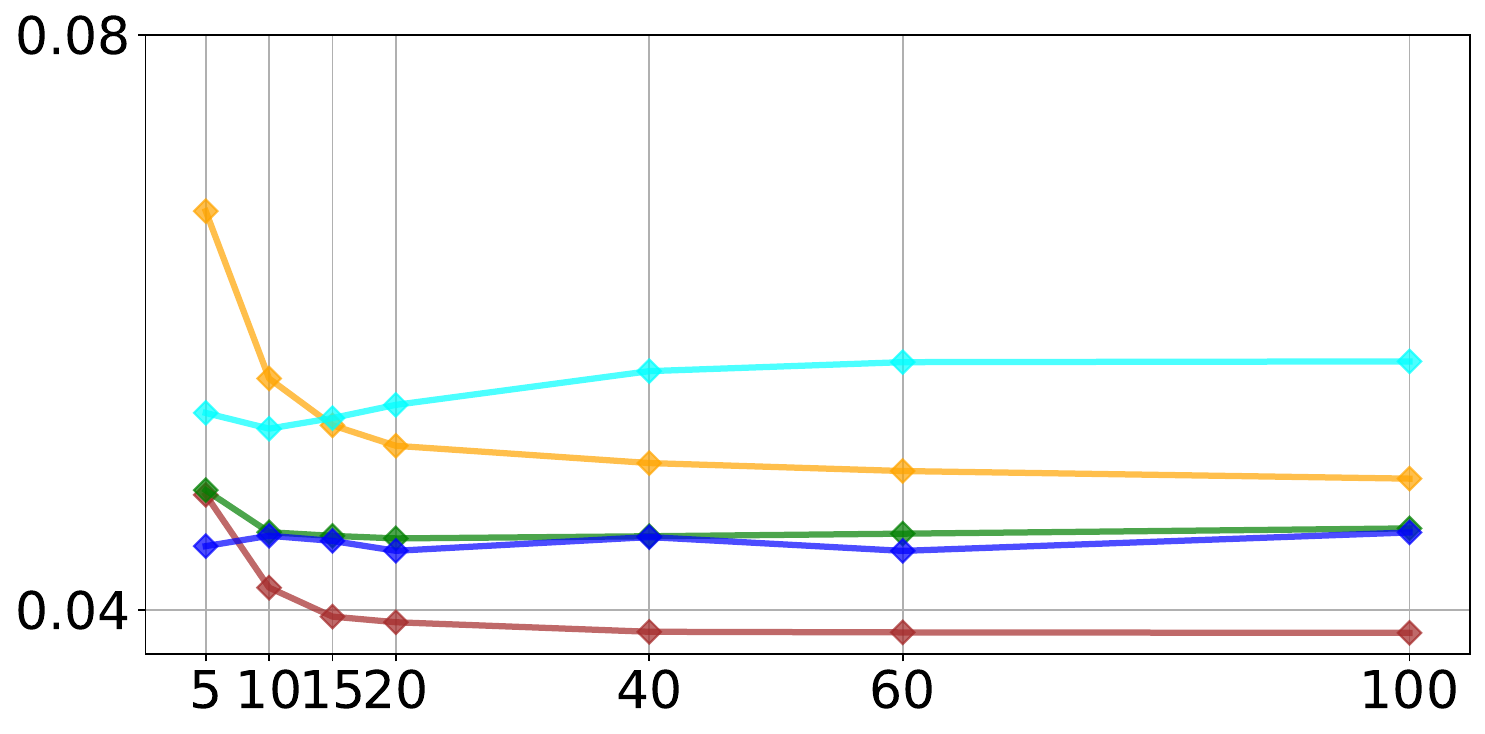}  
\\
\parbox[b]{1.6cm}{\centering Relative\\Improvement\\\;\\\;\\\;\\\;}
&\includegraphics[width=0.45\linewidth]{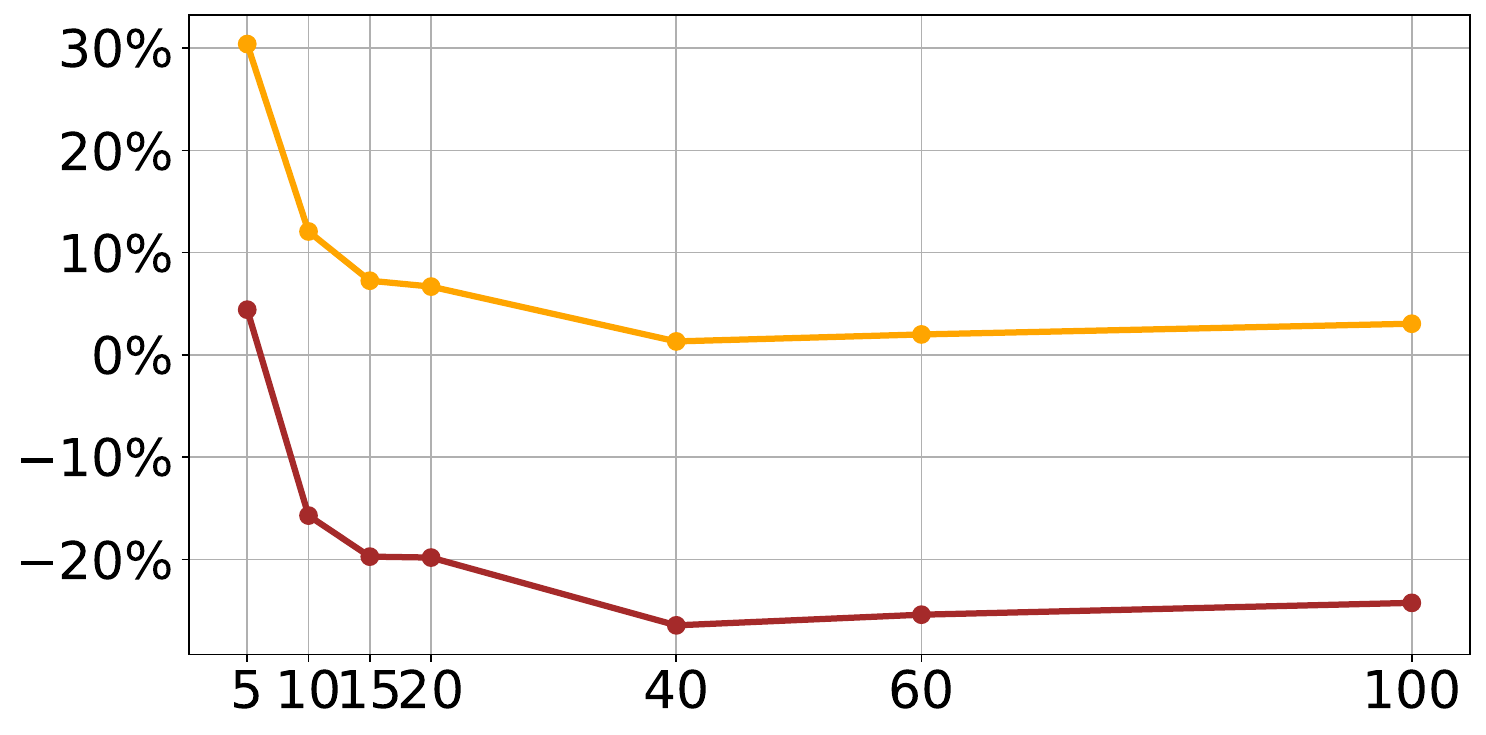}  
&\includegraphics[width=0.45\linewidth]{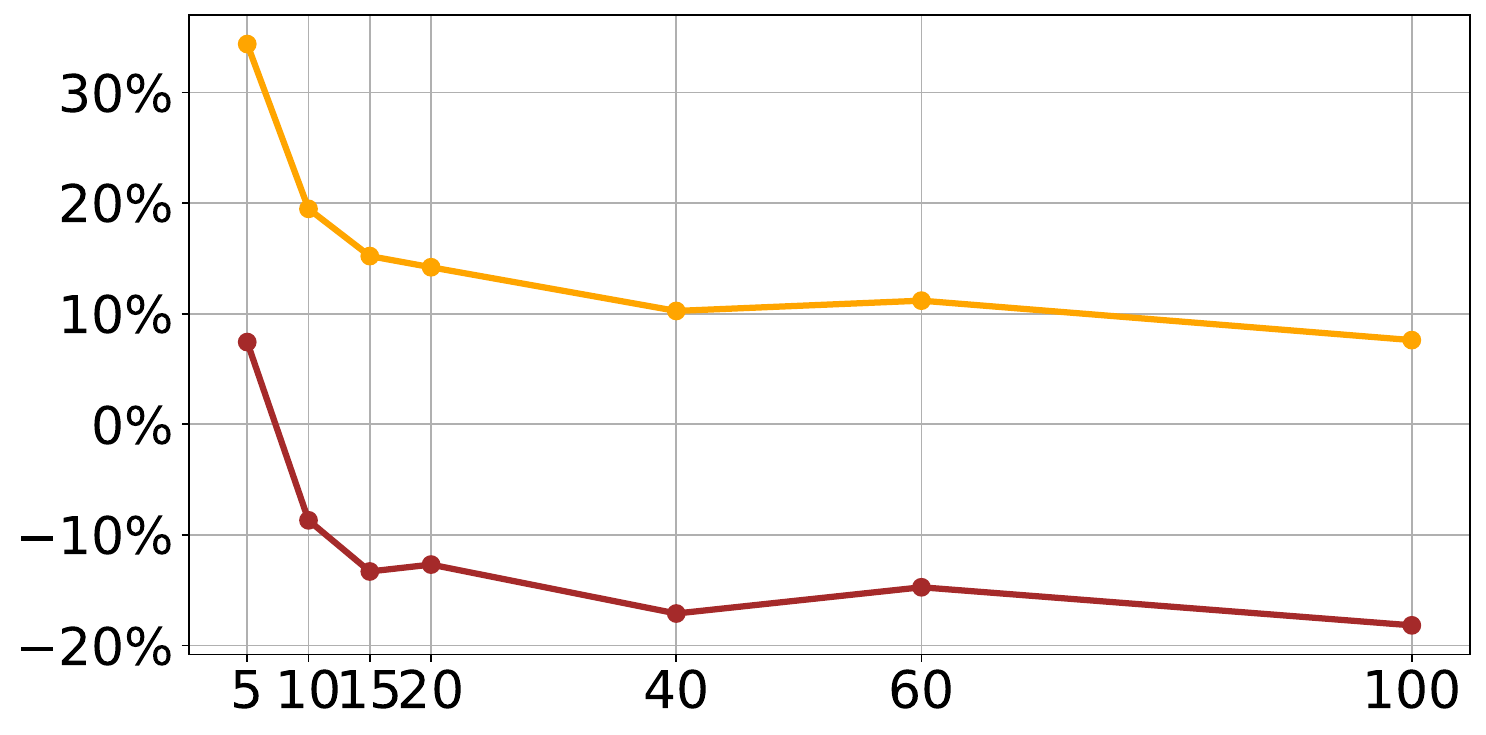}  
\\
\multirow{2}{*}{\textbf{Lorenz '63}}
& Ensemble Size
& Ensemble Size
\\
& $\sigma_y = 0.7$
& $\sigma_y = 1.0$
\\
& \multicolumn{2}{c}{\includegraphics[trim={0 1.5cm 0 1.5cm},clip, width=0.8\linewidth]{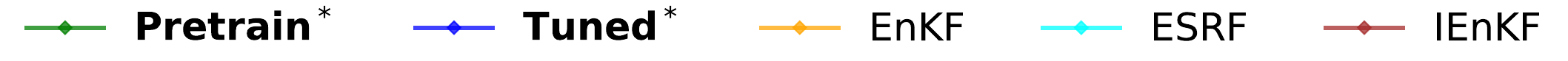}  }
\end{tabular}
}
\vspace{-0.5cm}
\caption{Comparison results on the Lorenz '63 system. The upper row of plots shows direct R-RMSE comparisons between different methods, while the lower row illustrates the relative improvement of our fine-tuning method compared to benchmarks. These comparisons are presented for observation noise levels $\sigma_y=0.7$ (left column) and $\sigma_y=1.0$ (right column). Our proposed MNMEF method is highlighted in the legends with bold font and an asterisk (e.g. \textbf{Pretrain}$^*$).}
\label{fig:L63}
\end{figure}

The parameter settings for the Lorenz '63 system are shown in Table~\ref{tab:lorenz63_settings}. The three parameters $\sigma$, $\rho$, and $\beta$ represent the Prandtl number, Rayleigh number, and geometric factor, respectively, chosen by Lorenz based on simplified thermal convection experiments to produce chaotic dynamics characterized by sensitivity to initial conditions.

In Figure~\ref{fig:L63}, we compare the R-RMSE \eqref{eq:avg_r_rmse} performance of our pretrained MNMEF and fine-tuned MNMEF models against benchmark methods EnKF, ESRF, and IEnKF for the Lorenz '63 system. The upper row of plots in Figure~\ref{fig:L63} displays these R-RMSE comparisons for observation noise levels $\sigma_y = 0.7$ (left plot) and $\sigma_y = 1.0$ (right plot). The LETKF method is not included in this comparison since the system has only three dimensions, making localization unnecessary. For a fair comparison, we turn off the localization part in our architecture by setting all entries in the localization weight matrices to be $1$. We only fine-tune the parameters $\theta_\mathrm{gain}$ for $F^\mathrm{gain}$ \eqref{eq:nn_F} and $\theta_\mathrm{infl}$ for $F^\mathrm{infl}$ \eqref{eq:f_infl} for ensemble sizes $N\ne 10$. The lower row of plots in Figure~\ref{fig:L63} shows the relative improvement $\mathcal{E}_\mathrm{RI}$ \eqref{eq:relative_imp} of our fine-tuned MNMEF method as compared to EnKF and IEnKF, for the respective noise levels shown.

We observe that the performance of different methods does not significantly change with increasing ensemble size. Our method consistently outperforms the EnKF and ESRF, but in the low-dimensional Lorenz '63 problem, it is outperformed by the IEnKF. As noted in Remark~\ref{rmk:IEnKF_dim}, IEnKF demonstrates strong performance in low-dimensional, highly nonlinear problems while struggling in high-dimensional systems without localization. This is consistent with our findings in the higher-dimensional experiments in Subsections~\ref{ssec:L96} and~\ref{ssec:KS}, where our method substantially outperforms IEnKF. Furthermore, the stability analysis in Subsection~\ref{ssec:stab_analysis} demonstrates that our method exhibits higher stability (lower standard deviation across test trajectories) than IEnKF.

\begin{remark}[IEnKF Dimensionality Challenges]\label{rmk:IEnKF_dim}
The IEnKF \cite{sakov2012iterative} uses the same observational information as the standard EnKF, but reformulates the filtering problem as a lag-1 smoothing problem, and solves it iteratively to better handle nonlinearities. While the IEnKF demonstrates excellent performance in low-dimensional systems (e.g., the Lorenz '63 model, $d=3$), its effectiveness diminishes as dimensionality increases (e.g., the Lorenz '96 model, $d=40$). This deterioration occurs due to lack of localization in the implementation of the IEnKF we use; however, even with large ensembles with less need for localization, our method outperforms the IEnKF in the higher-dimensional systems. Moreover, IEnKF generally provides a substantial improvement over non-iterative methods only when the observations are sufficiently dense, as shown in \cite{sakov2012iterative} using experiments with the Lorenz '96 model.

\end{remark}

\subsection{Computational Cost}
\label{ssec:cost}

In this subsection, we compare the run time per assimilation step (s/step) of MNMEF against benchmark methods (EnKF, ESRF, LETKF, and IEnKF) under matched CPU-only settings to ensure a fair comparison.

For an ensemble of size $N$, the per-step cost of MNMEF scales as $\mathcal{O}(N^2)$. The dominant terms are (i) the attention blocks in the set transformer, which entail pairwise interactions among $N$ members, and (ii) the computation of the learned Kalman gain $K_\theta$ and its gradients, implemented via batched linear solves with $N\times N$ systems. We do not form matrix inverses explicitly. While classical baselines admit textbook complexities, their practical scaling depends on solver details, localization, and parallelization, so a full theoretical comparison is beyond the scope of this paper.

To reflect practical efficiency, we conduct a CPU-only head-to-head timing in a single Python environment. All CPU timings are run on Intel(R) Core(TM) i9-14900KF.  We follow implementations in DAPPER~\cite{raanes_dapper_2024} and provide our own CPU-based implementations of EnKF, ESRF, LETKF, and IEnKF. We deliberately avoid GPU ports for these baselines because methods such as LETKF and IEnKF do not map cleanly to efficient, large-batch vectorization on current accelerators. 

\begin{figure}[t]
\centering
\begin{subfigure}[t]{0.48\linewidth}
\centering
\includegraphics[width=\linewidth]{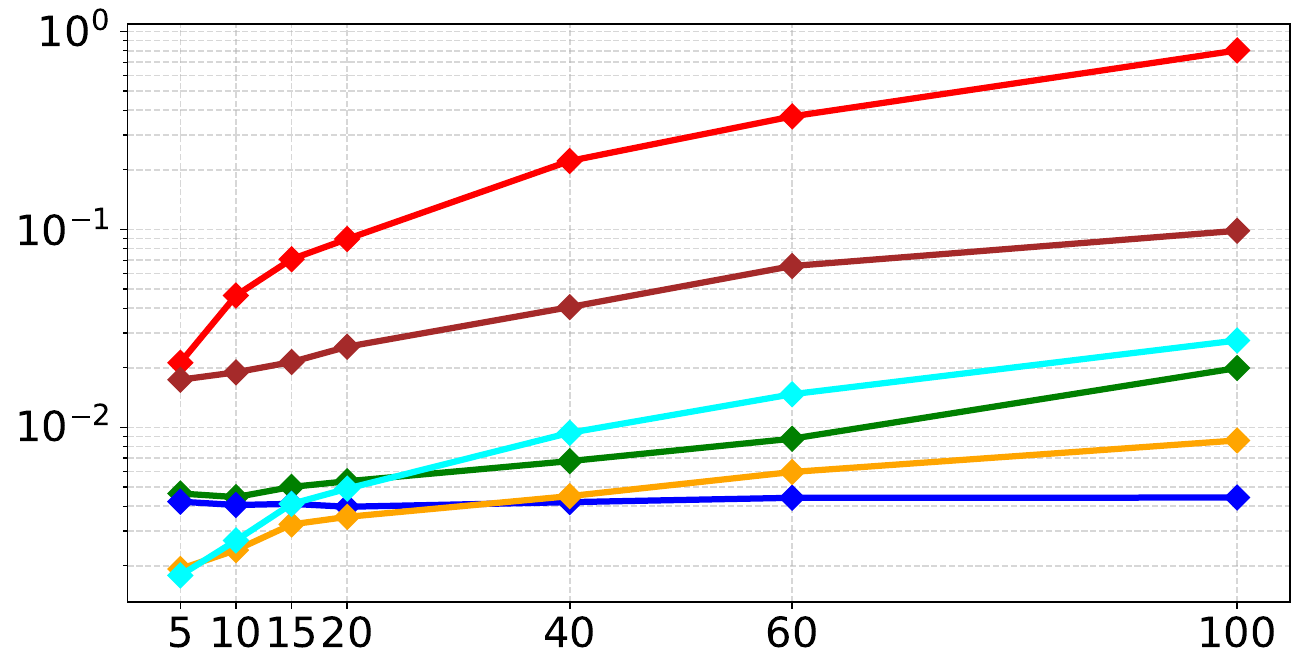}
\caption{Lorenz '96, $\sigma_y = 0.7$}
\label{fig:comp_time_l96_07}
\end{subfigure}\hfill
\begin{subfigure}[t]{0.48\linewidth}
\centering
\includegraphics[width=\linewidth]{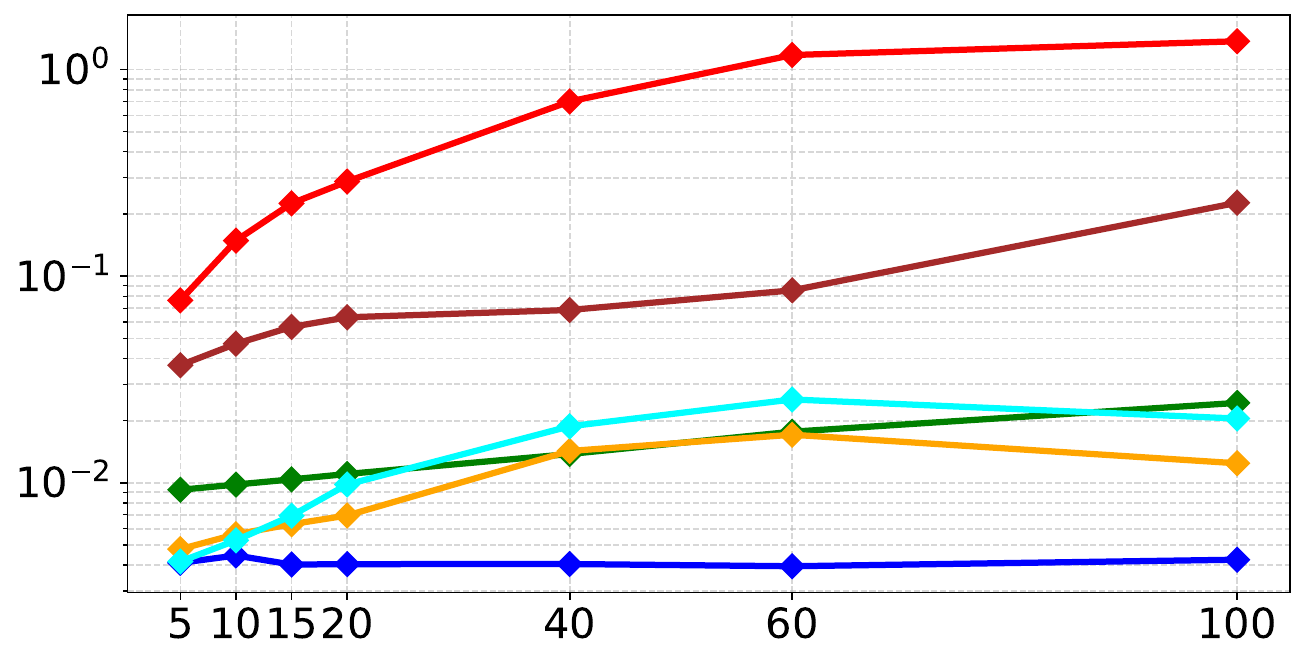}
\caption{KS, $\sigma_y = 0.7$}
\label{fig:comp_time_ks_07}
\end{subfigure}

\vspace{0.6em}
\includegraphics[trim={0 0.5cm 0 0.5cm},clip,width=0.75\linewidth]{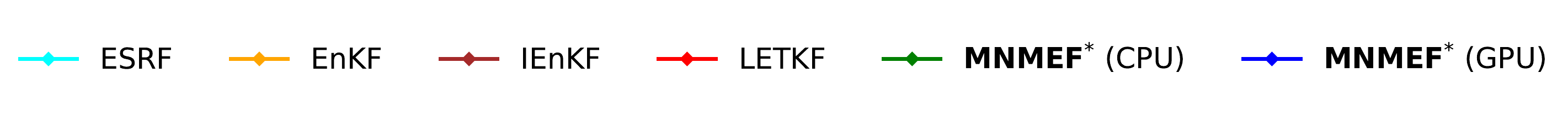}
\vspace{-0.3em}

\caption{Run time (s/analysis step), $\sigma_y{=}0.7$ only. All methods are timed on CPU (Intel(R) Core(TM) i9-14900KF). MNMEF is comparable in speed to EnKF and ESRF, and is slightly slower than EnKF because MNMEF additionally computes the correction terms. MNMEF remains much faster than LETKF and IEnKF. For MNMEF, we additionally report GPU timing on a single GPU (RTX 4080 Super), shown as the blue curve. The GPU version shows very stable runtime across these ensemble sizes due to efficient parallelization.}
\label{fig:comp_time}
\end{figure}

We evaluate on Lorenz~’96 and KS with the same configurations as in Subsections~\ref{ssec:L96} and~\ref{ssec:KS}. For each ensemble size $N$ and observation noise level $\sigma_y=0.7$, we report the mean of the per-step run time (s/analysis step), averaged over trajectories and time steps. We do not distinguish the pretrained and fine-tuned variants of our MNMEF since they share the same architecture. The set-transformer parameter counts follow Table~\ref{tab:params_ft}. Results are visualized in Figure~\ref{fig:comp_time}. MNMEF is comparable in speed to EnKF and ESRF, and is slightly slower than EnKF because MNMEF additionally computes the correction terms. MNMEF remains much faster than LETKF and IEnKF. 

For MNMEF, we provide an additional GPU-based run time in Figure~\ref{fig:comp_time}, since its original implementation is designed for GPU. MNMEF’s additional GPU timings are obtained on a single NVIDIA RTX 4080 Super. The GPU version of MNMEF shows very stable runtime across these ensemble sizes due to efficient parallelization.

\subsection{Robustness To Inherent Randomness}\label{ssec:stab_analysis}
Due to the inherent randomness in both the filtering problem and the algorithm, the same method may exhibit varying performance across different test trajectories. To mitigate this randomness and obtain a more reliable assessment of each method's effectiveness, we evaluate their performance on $M_{\mathrm{test}}$ trajectories and report the mean R-RMSE $\bar{\mathcal{E}}$ \eqref{eq:avg_r_rmse}. While the mean R-RMSE provides a measure of overall accuracy, we are interested in the stability of each method across different trajectories. To quantify this stability, we compute the standard deviation (std) of the R-RMSE values across all test trajectories $\{V_m^\dagger\}_{m=1}^{M_\mathrm{test}}$ and their corresponding estimated trajectories $\{V_m\}_{m=1}^{M_\mathrm{test}}$, defined as:
\begin{equation}\label{eq:std_rrmse}
\sigma_{\mathcal{E}} = \sqrt{\frac{1}{M_{\mathrm{test}}}\sum_{m=1}^{M_{\mathrm{test}}}\left(\mathcal{E}(V_m,V_m^\dagger) - \bar{\mathcal{E}}\right)^2}
\end{equation}
A smaller $\sigma_{\mathcal{E}}$ indicates better stability of the method. Therefore, in this section, we consider methods with lower std values to be more stable and thus preferable.

\begin{remark}
In ensemble-based data assimilation, variance- or standard-deviation–type quantities are commonly used to characterize ensemble spread~\cite{bach2024inverse}. In our work, the spread is visualized in Figs.~\ref{fig:lorenz96_dim01} and \ref{fig:ks_dim01} via the shaded bands: the 95\% confidence interval is provided as the ensemble mean $\pm 1.96\times$ ensemble standard deviation. Our focus here is state estimation rather than full uncertainty calibration, so we do not present a quantitative comparison of spread metrics. Note that the stability measure $\sigma_{\mathcal{E}}$ in~\eqref{eq:std_rrmse} serves a different purpose: it assesses the consistency of estimation errors across test trajectories, not the dispersion of ensemble members.
\end{remark}

In Table~\ref{tab:std_comparison}, we show the std of different methods on the dataset Lorenz '96, KS and Lorenz '63 with the observation noise $\sigma_y=1.0,\,0.7$. The std values in the table is the averaged std over the ensemble sizes $N=5,10,15,20,40,60,100$.
Based on Table~\ref{tab:std_comparison}, we observe that our pretrained MNMEF and fine-tuned MNMEF demonstrate comparable standard deviation values, both of which are significantly lower than all benchmark methods. This indicates that our approaches maintain more consistent performance across different test trajectories. The enhanced stability suggests that our methods are more reliable in practical applications.

\begin{table}[htbp]
\centering
\caption{Standard deviation (std) of the relative root mean square error (R-RMSE) as defined in~\eqref{eq:std_rrmse}. We consider both \textbf{Pretrain*} and \textbf{Tuned*} variants of our MNMEF method. The std shown in the table is an averaged std over ensemble sizes $N=5,10,15,20,40,60,100$.
A smaller std indicates better stability of the method. The lowest std in each column is highlighted in bold. The stds for both the pretrained and fine-tuned variants of our MNMEF method are similar and consistently lower than those of the benchmark methods, indicating greater robustness of our approach to the inherent randomness across different test trajectories.}
\label{tab:std_comparison}
\renewcommand{\arraystretch}{1.2}
\begin{tabular}{l|cc|cc|cc}
\hline
\multirow{2}{*}{Method} & \multicolumn{2}{c|}{Lorenz '96 \eqref{eq:L96_equations}} & \multicolumn{2}{c|}{KS \eqref{eq:KS_equation}} & \multicolumn{2}{c}{Lorenz '63 \eqref{eq:L63_dynamic}} \\
\cline{2-7}
 & $\sigma_y=1.0$ & $\sigma_y=0.7$ & $\sigma_y=1.0$ & $\sigma_y=0.7$ & $\sigma_y=1.0$ & $\sigma_y=0.7$ \\
\hline
\textbf{Pretrain*} & \textbf{1.03e-2} & 1.26e-2 & \textbf{8.60e-3} & 1.07e-2 & \textbf{1.78e-3} & 1.40e-3 \\
\textbf{Tuned*} & 1.08e-2 & \textbf{1.12e-2} & 8.91e-3 & \textbf{1.07e-2} & 1.79e-3 & \textbf{1.29e-3} \\
EnKF \cite{evensen2003ensemble} & 2.16e-2 & 3.10e-2 & 2.66e-2 & 3.15e-2 & 3.41e-3 & 2.32e-3 \\
ESRF \cite{tippett_ensemble_2003} & 3.09e-2 & 1.95e-2 & 2.59e-2 & 3.27e-2 & 2.78e-3 & 2.01e-3 \\
LETKF \cite{hunt2007efficient} & 2.63e-2 & 6.13e-2 & 2.66e-2 & 3.58e-2 & - & - \\
IEnKF \cite{sakov2012iterative} & 1.83e-2 & 2.61e-2 & 2.23e-2 & 3.03e-2 & 2.15e-3 & 1.44e-3 \\
\hline
\end{tabular}
\end{table}

\subsection{Inflation and Localization}
\label{ssec:Exp_inf_loc}
In this section, we will conduct further experiments on learning the inflation (Subsection~\ref{sssec:inflation}) and localization (Subsection~\ref{sssec:localization}) components in the architecture of our MNMEF method. These experiments demonstrate that what our method learns has many similarities with what is used in the application of inflation and localization techniques to the benchmark methods.

\subsubsection{Inflation Experiments}\label{sssec:inflation_exp}
In Subsection~\ref{sssec:inflation}, we introduce our learning-based inflation scheme \eqref{eq:our_infl} which depends on the learned output $\hun_\theta$ from the neural network $F^\mathrm{infl}$ \eqref{eq:f_infl}. 

\begin{figure}[htbp]
\centering
\resizebox{\textwidth}{!}{
\begin{tabular}{ccc}
\parbox[b]{1.6cm}{\centering Relative\\RMSE\\\;\\\;\\\;\\\;}
&\includegraphics[width=0.45\linewidth]{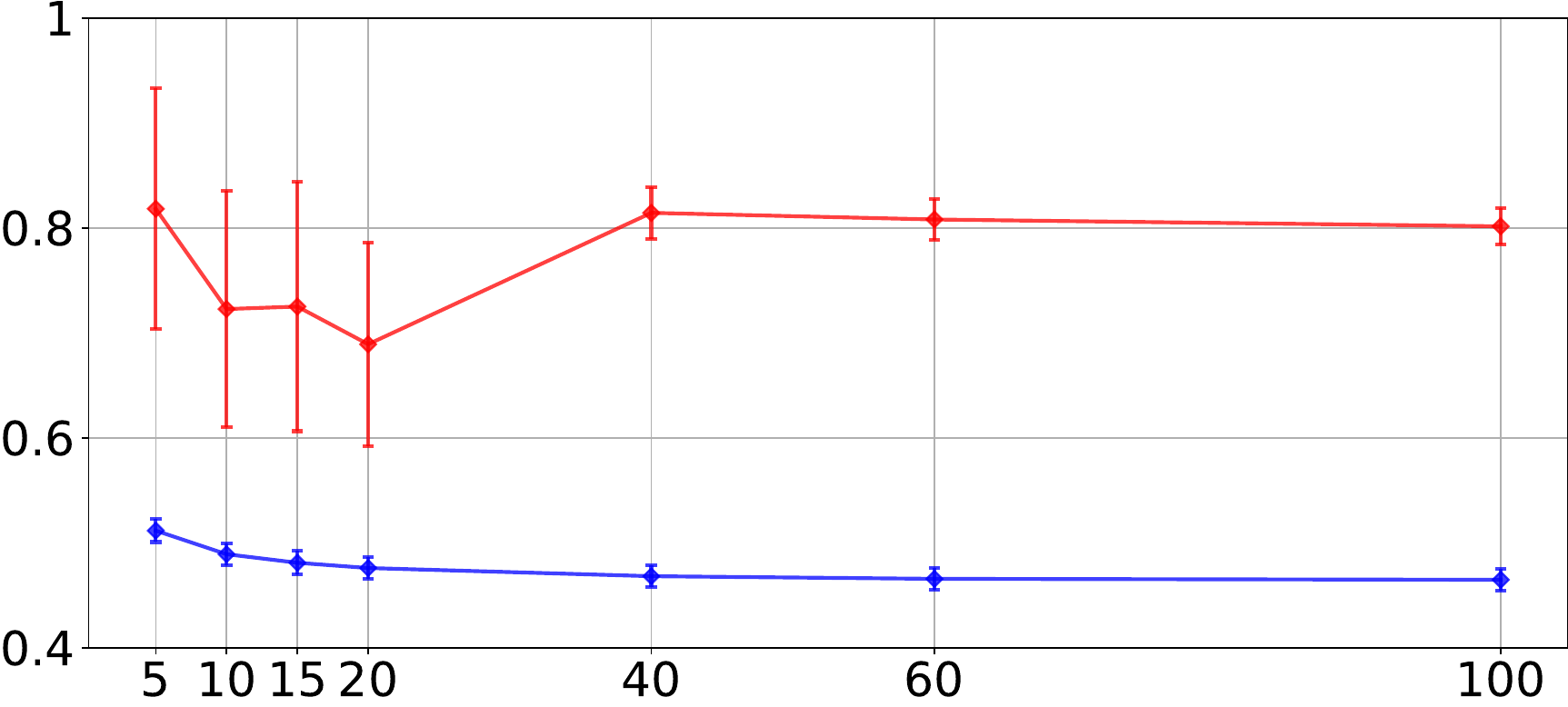}  
&\includegraphics[width=0.45\linewidth]{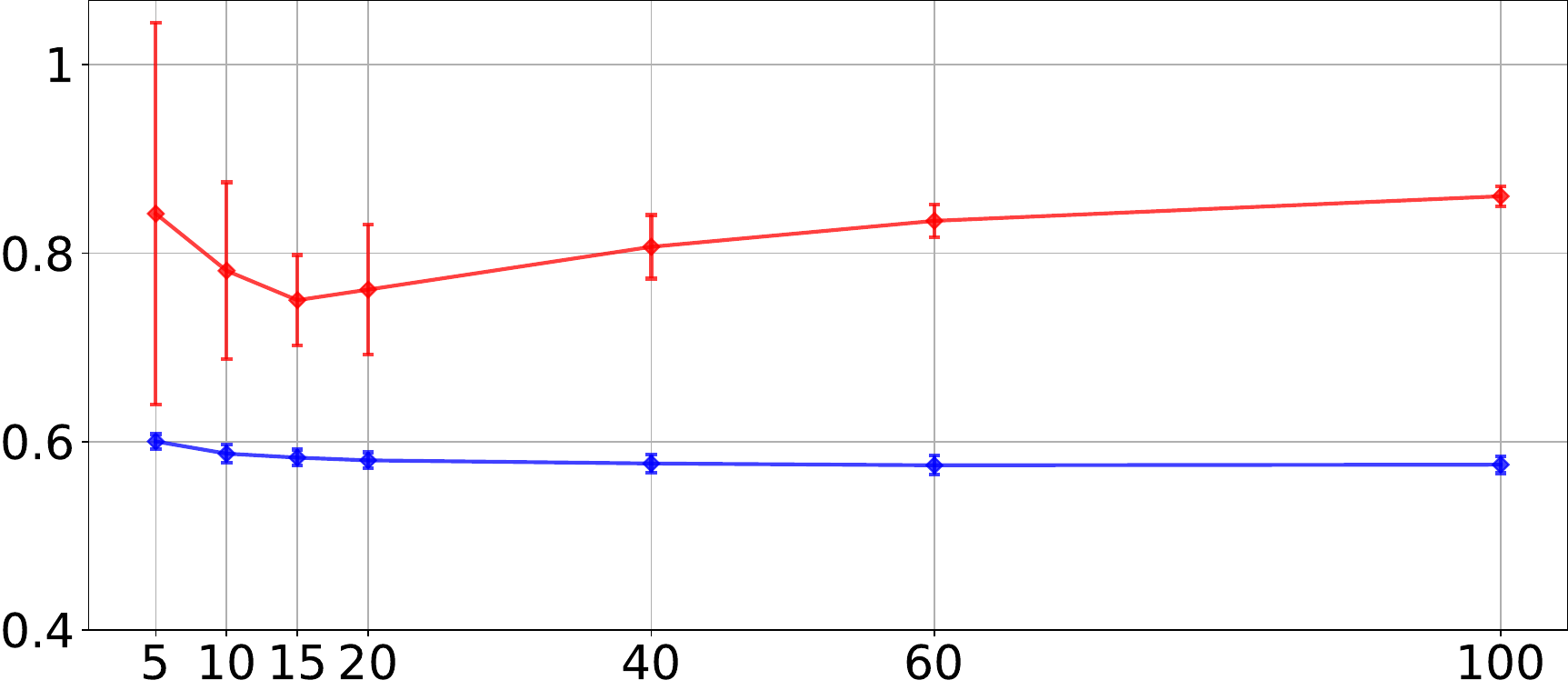}  
\\
\textbf{Kuramoto--}
& Ensemble Size
& Ensemble Size
\\
\textbf{Sivashinsky}
& $\sigma_y = 0.7$
& $\sigma_y = 1.0$
\\
& \multicolumn{2}{c}{\includegraphics[trim={0 1.5cm 0 1.5cm},clip, width=0.7\linewidth]{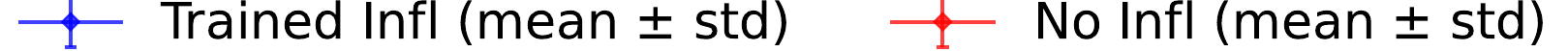}  }
\end{tabular}
}
\vspace{-0.4cm}
\caption{Comparison of the trained inflation versus the no inflation scenario when applied to the KS system \eqref{eq:KS_equation}, using our MNMEF method, fine-tuned for different ensemble sizes. The plots compare R-RMSE with and without inflation at observation noise levels $\sigma_y = 0.7$ (left plot) and $\sigma_y = 1.0$ (right plot). In summary, our trained inflation effectively improves the performance, but even without inflation, our MNMEF does not completely fail.}
\label{fig:ks_infl}
\end{figure}

We are particularly interested in quantifying the effect of our proposed inflation scheme on the overall performance of our method. Additionally, despite not providing the inflation term $\hun_\theta$ with information about the true observation $y^\dagger$ or observation noise $\Gamma$ in $F^\mathrm{infl}$, $\hun_\theta$ has significantly more degrees of freedom than classical inflation approaches. This raises concerns that $\hun_\theta$ might exceed the scope of inflation: that it might not act merely  as a correction term but might completely overwhelm the updates, thus rendering other parts of our architecture ineffective. We show that this is not the case, but at the same time we showcase the benefits of including this learned inflation-like correction.

For our experimental approach, we first pretrain the models and then fine-tune them separately for various ensemble sizes. During inference, we deliberately disable the learned inflation by setting $\hun_\theta=0$ and analyze the resulting change in the average R-RMSE $\bar{\mathcal{E}}$~\eqref{eq:avg_r_rmse}. We stress that this manipulation does not represent an optimal or even well-trained MNMEF filter without inflation; instead, it intentionally removes the correction term to create a more challenging robustness test—probing whether the filter remains stable and avoids divergence in the absence of inflation. These experiments are conducted on the KS dynamical system~\eqref{eq:KS_equation} only.

Our results are presented in Figure~\ref{fig:ks_infl}. We compare the mean R-RMSE $\bar{\mathcal{E}}$ between our standard architecture with the trained inflation (blue curve) and the no inflation version by setting $\hun_\theta=0$ (red curve) for observation noise $\sigma_y=0.7$ and $1.0$. The results in Figure~\ref{fig:ks_infl} clearly demonstrate the beneficial impact of our trained inflation approach. In addition, our method remains functional, with increased R-RMSE values, when we remove inflation by setting $\hun_\theta=0$ (equivalent to $\alpha=1$ in the EnKF). This confirms that our experiment, while intentionally suboptimal, demonstrates that the proposed MNMEF remains numerically stable and does not diverge even in this deliberately more difficult configuration. This addresses our concern, because, if inflation were dominating the architecture by masking the contribution of the state $v^{(n)}$, removing it would cause model collapse.

\subsubsection{Localization Experiments}\label{sssec:localization_exp}
According to Subsection~\ref{sssec:localization}, our learning-based localization approach essentially learns a parameterized function $g_\theta$, which maps distance to weight values. During training, we only enforce the constraint that the output of this function must lie within the interval [0,2], i.e., $g_\theta: \mathbb{R} \rightarrow [0,2]$. Unlike traditional localization functions used in EnKF such as the Gaspari–Cohn function, our approach does not impose any explicit constraints on the shape or monotonicity of $g_\theta$. 

\begin{figure}[htb]
    \centering
    \resizebox{\textwidth}{!}{
    \begin{tabular}{cccccccc}
         \parbox[b]{1.5cm}{\centering $\sigma_y=0.7$\\\;}
         &\includegraphics[width=0.11\linewidth]{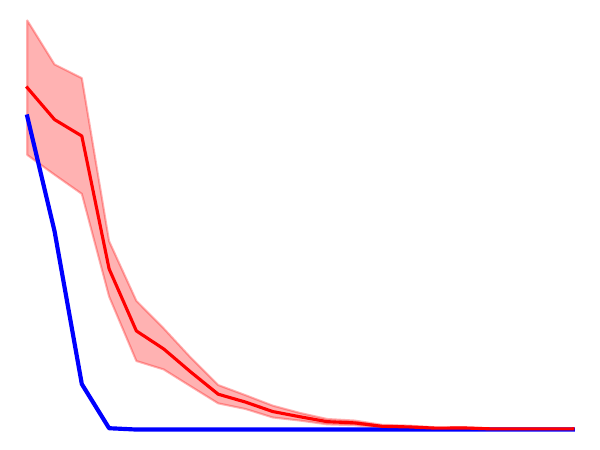} 
         &\includegraphics[width=0.11\linewidth]{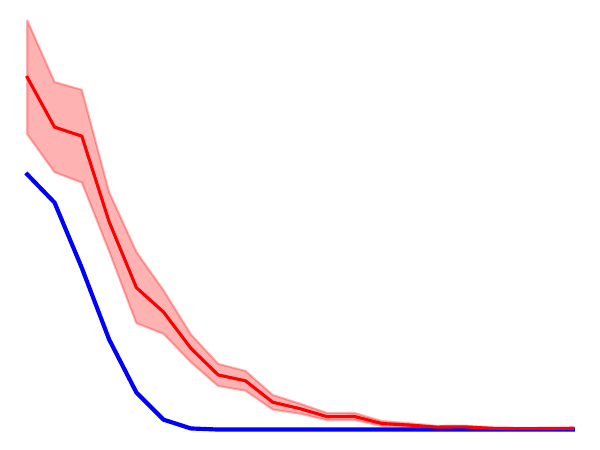} 
         &\includegraphics[width=0.11\linewidth]{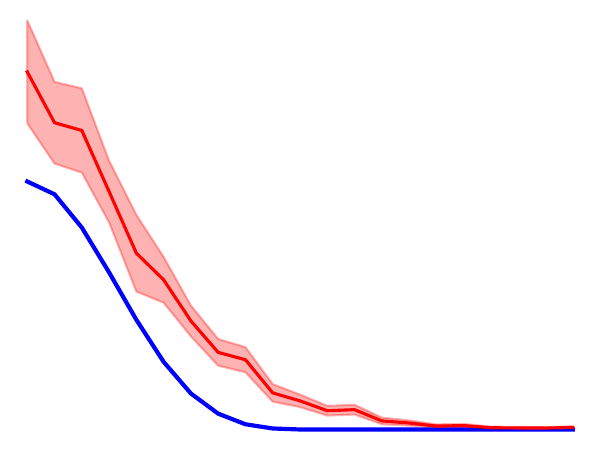}
         &\includegraphics[width=0.11\linewidth]{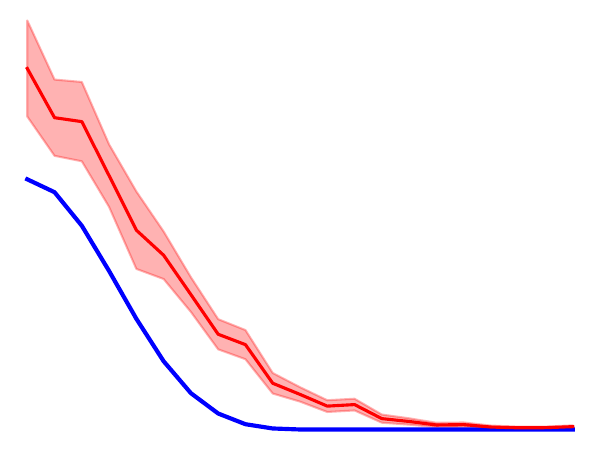} 
         &\includegraphics[width=0.11\linewidth]{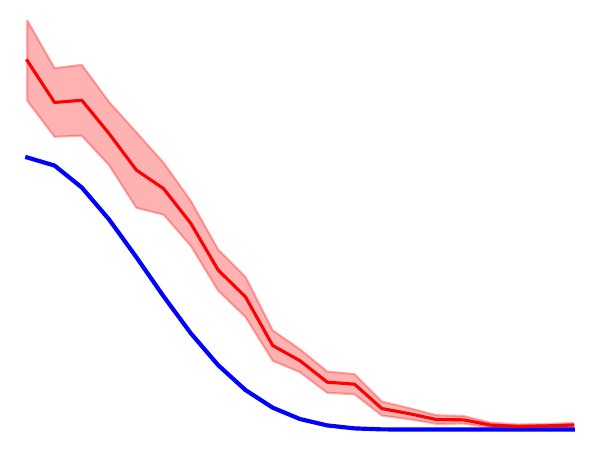}
         &\includegraphics[width=0.11\linewidth]{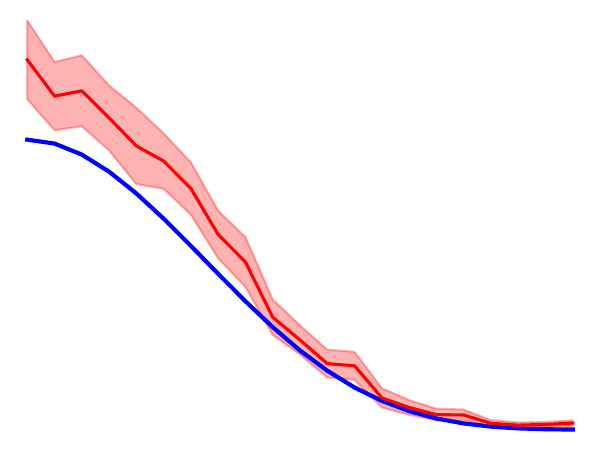}
         &\includegraphics[width=0.11\linewidth]{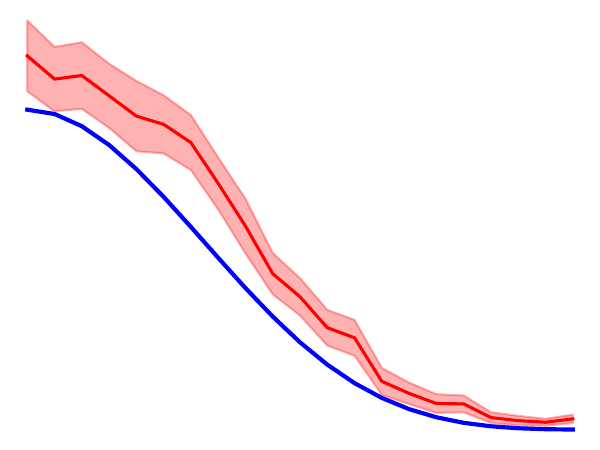} 
         \\
         \parbox[b]{1.5cm}{\centering $\sigma_y=1.0$\\\;}
         &\includegraphics[width=0.11\linewidth]{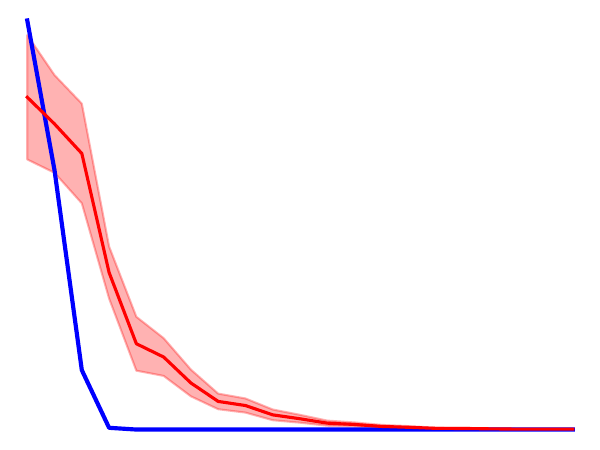} 
         &\includegraphics[width=0.11\linewidth]{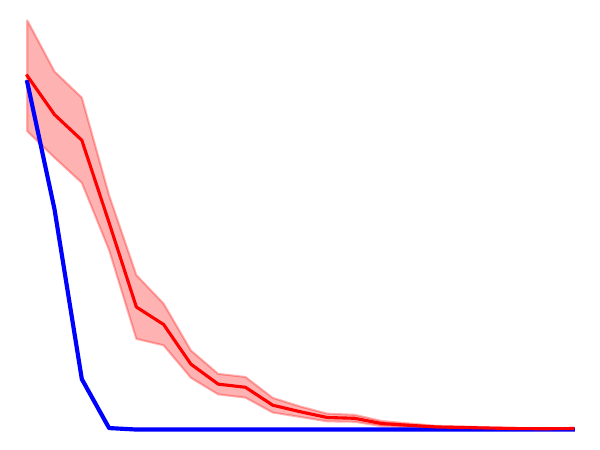} 
         &\includegraphics[width=0.11\linewidth]{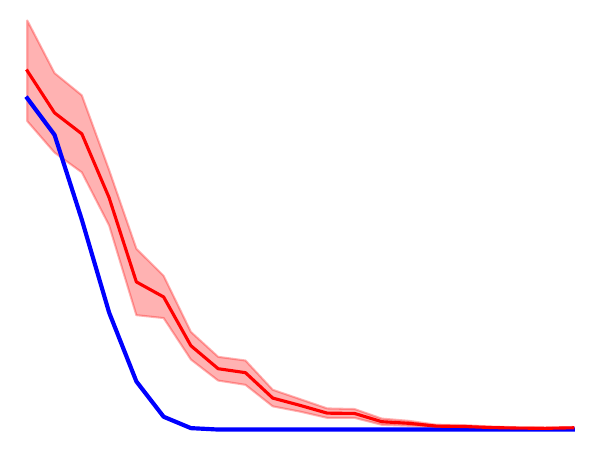}
         &\includegraphics[width=0.11\linewidth]{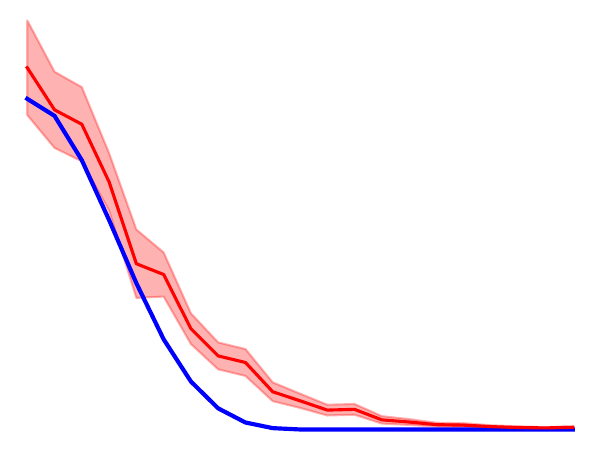} 
         &\includegraphics[width=0.11\linewidth]{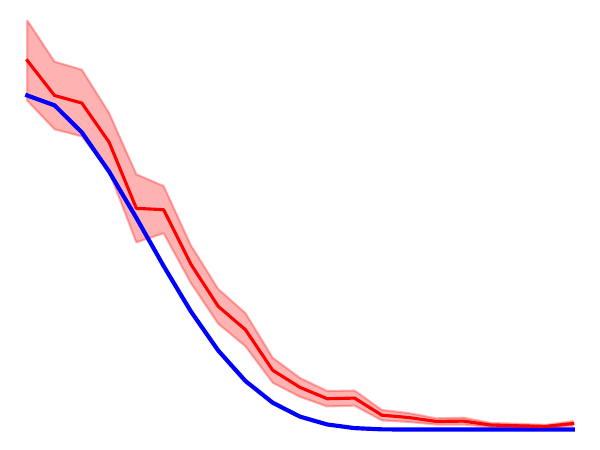}
         &\includegraphics[width=0.11\linewidth]{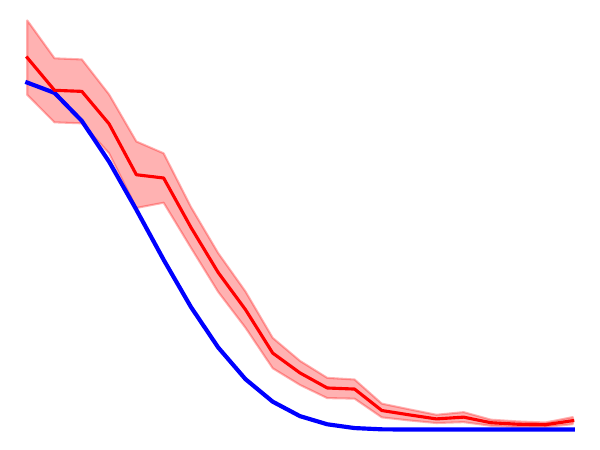}
         &\includegraphics[width=0.11\linewidth]{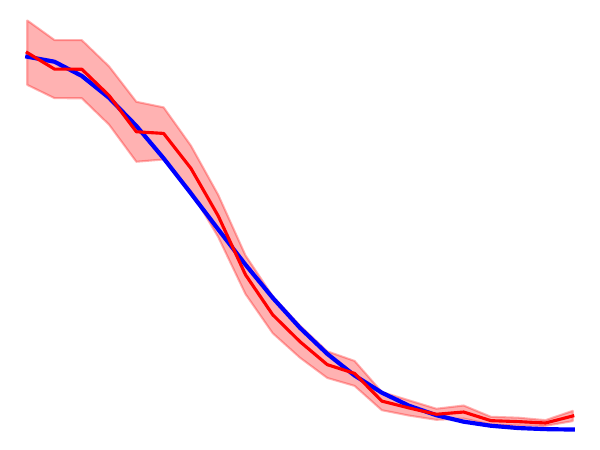} 
         \\ \midrule
         $N$ & 5 & 10 & 15 & 20 & 40 & 60 & 100
         \\
         &\multicolumn{7}{c}{\includegraphics[trim={0 0 0 0},clip, width=0.7\linewidth]{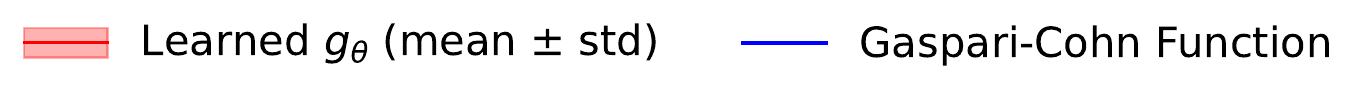}  }
         \vspace{-0.5cm}
    \end{tabular}
    }
    \caption{Comparison of our learned distance-to-weight function $g_\theta$ (red, mean $\pm$ 1 std across different time steps due to its adaptivity) with the LETKF Gaspari--Cohn (GC) function (blue) on the Lorenz '96 model for $\sigma_y=1.0,\,0.7$ and ensemble sizes $N=5,10,15,20,40,60,100$. The learned distance-to-weight function is from our MNMEF pretrained on $N=10$ and fine-tuned for different ensemble sizes. The LETKF GC radius parameters are optimally selected via grid search. Remarkably, even without any explicit constraints on the shape of $g_\theta$, the learned weights naturally decrease as the distance increases, similar to the empirical GC function in LETKF.}
    \label{fig:gc_vs_learned_lorenz96}
\end{figure}

In this subsection, we present experimental comparisons between our learned function $g_\theta$ and the Gaspari–Cohn (GC) function with the optimal localization radius, found by grid-search, for LETKF; 
see Figure~\ref{fig:gc_vs_learned_lorenz96} .
For ensemble sizes $N=5,10,15,20,40,60,100$, the optimal GC radius values chosen are $1,1,2,3,4,4,6$ (multiplied by $\sqrt{10/3}$ when implemented \cite{raanes_dapper_2024}), respectively. The learned function $g_\theta$ (shown as red curves with the mean and standard deviation of the learned $g_\theta$ across assimilation time steps) demonstrates a similar shape to the GC function with optimally chosen radius (blue curves). Notably, our approach dynamically adjusts the localization weights at each assimilation step based on ensemble size and distribution, while the GC function remains static throughout the assimilation process.

The results in Figure~\ref{fig:gc_vs_learned_lorenz96} demostrates the effectiveness of our learned localization. Despite imposing no explicit constraints on shape or monotonicity, the learned function $g_\theta$ naturally exhibits similar behavior to the GC function: it decreases monotonically with distance and shares a similar shape. A key difference is that our function occasionally exceeds values of 1 at short distances, reflecting a combined localization and inflation effect where covariance between closely spaced dimensions is amplified, since we are learning the inflation and localization simultaneously. This adaptivity allows $g_\theta$ to dynamically adjust to the ensemble distribution, leading to improved performance over the fixed GC function, as supported by RMSE results in Subsections~\ref{ssec:L96} and \ref{ssec:KS}.

\begin{remark}[Using GC localization directly]
Experimental results suggest that one can use the GC function to avoid training the localization. However, this approach still requires selecting the radius hyperparameter and typically yields inferior performance compared to our learned localization.
\end{remark}

\subsection{Fine-Tuning for Different Ensemble Size}
\label{ssec:Exp_FT}

According to Subsection~\ref{ssec:fine-tuning}, our proposed fine-tuning strategy freezes parameters $\theta_\mathrm{ST}$ of the set transformer $F^\mathrm{ST}$ \eqref{eq:FST}, while updating parameters $\theta_\mathrm{gain}$, $\theta_\mathrm{infl}$ and $\theta_\mathrm{loc}$ of the MLPs that control the gain \eqref{eq:nn_F}, inflation \eqref{eq:f_infl}, and localization \eqref{eq:f_loc}. Here we conduct experiments to illustrate the efficiency and effectiveness of our proposed fine-tuning approach.

Firstly, we consider the time complexity between the pretraining stage and fine-tuning stage. Let $P_\mathrm{pretrain}$ and $P_\mathrm{ft}$ denote the total number of pretraining and fine-tuning parameters respectively, i.e.
\begin{subequations}
\begin{align}
    P_\mathrm{pretrain} &= |\theta_\mathrm{ST}| + |\theta_\mathrm{gain}| + |\theta_\mathrm{infl}| + |\theta_\mathrm{loc}|,\\
    P_\mathrm{ft} &= |\theta_\mathrm{gain}| + |\theta_\mathrm{infl}| + |\theta_\mathrm{loc}|.
\end{align}
\end{subequations}
For fixed dynamic and training trajectories, the time complexity for pretraining with the ensemble size $N$ is $\mathcal{O}\bigl(ENP_t\bigr)$ while the fine-tuning with the ensemble size $N'$ has the time complexity $\mathcal{O}\bigl(E'N'P_{ft}\bigr)$, where $E$ and $E'$ are the epochs for pretraining and fine-tuning,  respectively. 

We provide a detailed comparison in Table~\ref{tab:params_ft}, presenting the total number of pretraining parameters $P_\mathrm{pretrain}$, fine-tuning parameters $P_\mathrm{ft}$, and the number of epochs $E$ used during the pretraining and fine-tuning phases. When the ensemble size during pretraining ($N$) is close to the ensemble size used in fine-tuning ($N'$), Table~\ref{tab:params_ft} shows that the fine-tuning stage requires less than 1\% of the pretraining computational time. To quantitatively support this observation, we present detailed computational time comparisons in Table~\ref{tab:time_comp}. All training times reported here are obtained using a single RTX 4080 Super GPU. The training batch sizes are adaptively chosen to fully use the 16 GB memory.

\begin{table}[h!]
\centering
\caption{Comparison of total parameters, fine-tuning parameters, and epochs for various datasets.}
\label{tab:params_ft}
\begin{tabular}{lcccc}
\toprule
Dataset & Total Params & FT Params & Pretrain Epochs & FT Epochs \\
\midrule
Lorenz '96 & 341551 & 63343 & 1000 & 20 \\
\midrule
KS         & 374289 & 90065 & 1000 & 20 \\
\midrule
Lorenz '63 & 309383 & 34119 & 1000 & 20 \\
\bottomrule
\end{tabular}
\end{table}

\begin{table}[h!]
\centering
\caption{Comparison of pretraining and fine-tuning computational time for observation noise $\sigma_y=1.0$ with different ensemble sizes on a single RTX 4080 Super GPU. The ratio is between the computation time of the fine-tuning stage and the time of the pretraining stage. The fine-tuning time increases with the ensemble size, but it remains significantly faster compared to the pretraining stage.}
\label{tab:time_comp}
\begin{tabular}{l|c|cc|cc|cc}
\toprule
Ens Size & $N=10$ & \multicolumn{2}{c|}{$N'=20$} & \multicolumn{2}{c|}{$N'=40$} & \multicolumn{2}{c}{$N'=100$} \\
\midrule
Dataset & Pretrain (hrs) & FT (hrs) & Ratio & FT (hrs) & Ratio & FT (hrs) & Ratio \\
\midrule
Lorenz '96 & 4.03 & 0.051 & 1.27\% & 0.065 & 1.61\% & 0.133 & 3.30\%\\
KS         & 8.22 & 0.106 & 1.29\% & 0.132 & 1.61\% & 0.260 & 3.16\%\\
Lorenz '63 & 1.82 & 0.022 & 1.21\% & 0.038 & 2.09\% & 0.086 & 4.73\% \\
\bottomrule
\end{tabular}
\end{table}

\begin{figure}[h!]
    \centering
    \begin{subfigure}{0.45\textwidth}
        \centering
        \includegraphics[width=\linewidth]{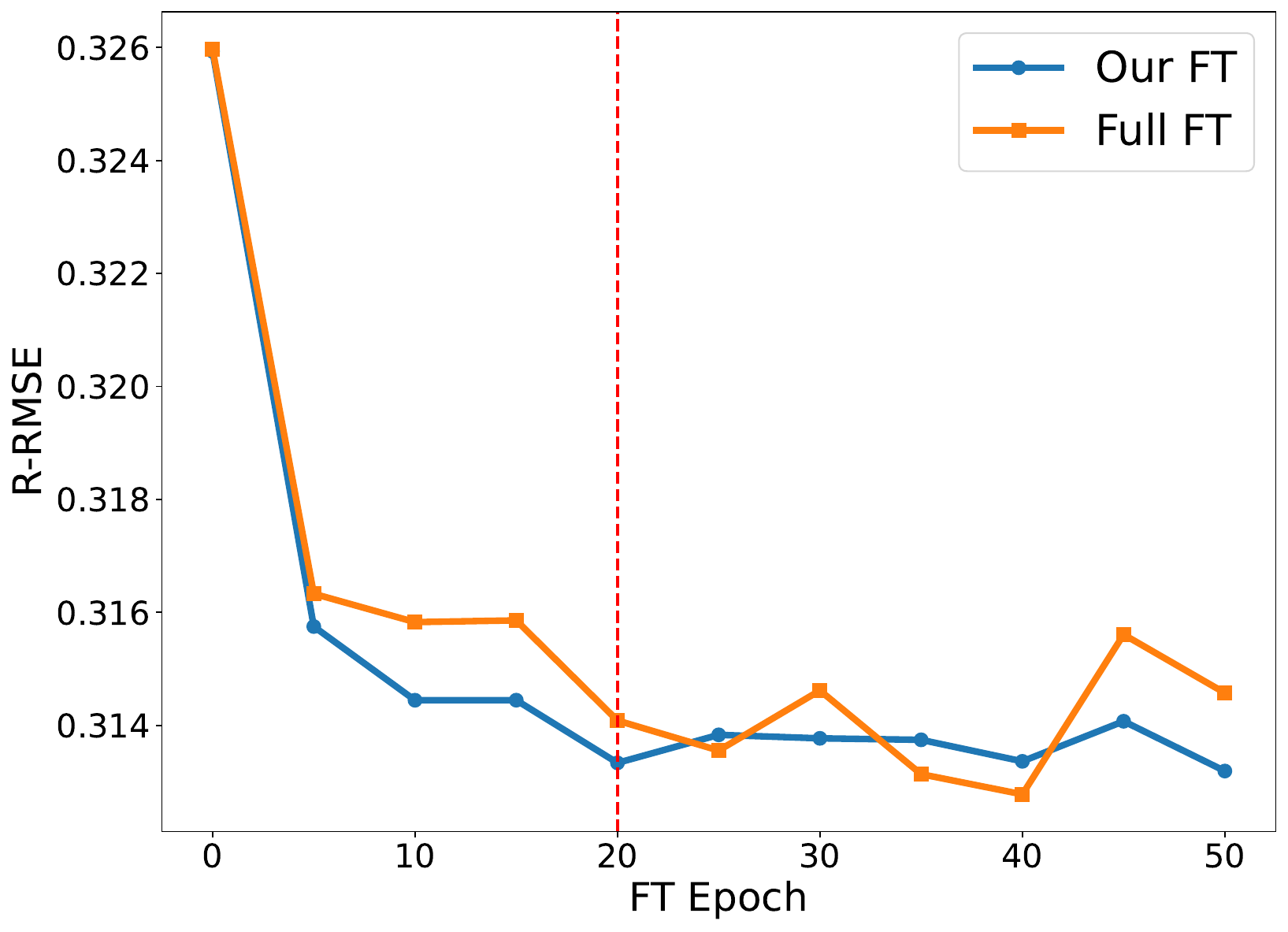}
        \caption{Ensemble Size $20$}
        \label{fig:ft_20}
    \end{subfigure}
    \begin{subfigure}{0.45\textwidth}
        \centering
        \includegraphics[width=\linewidth]{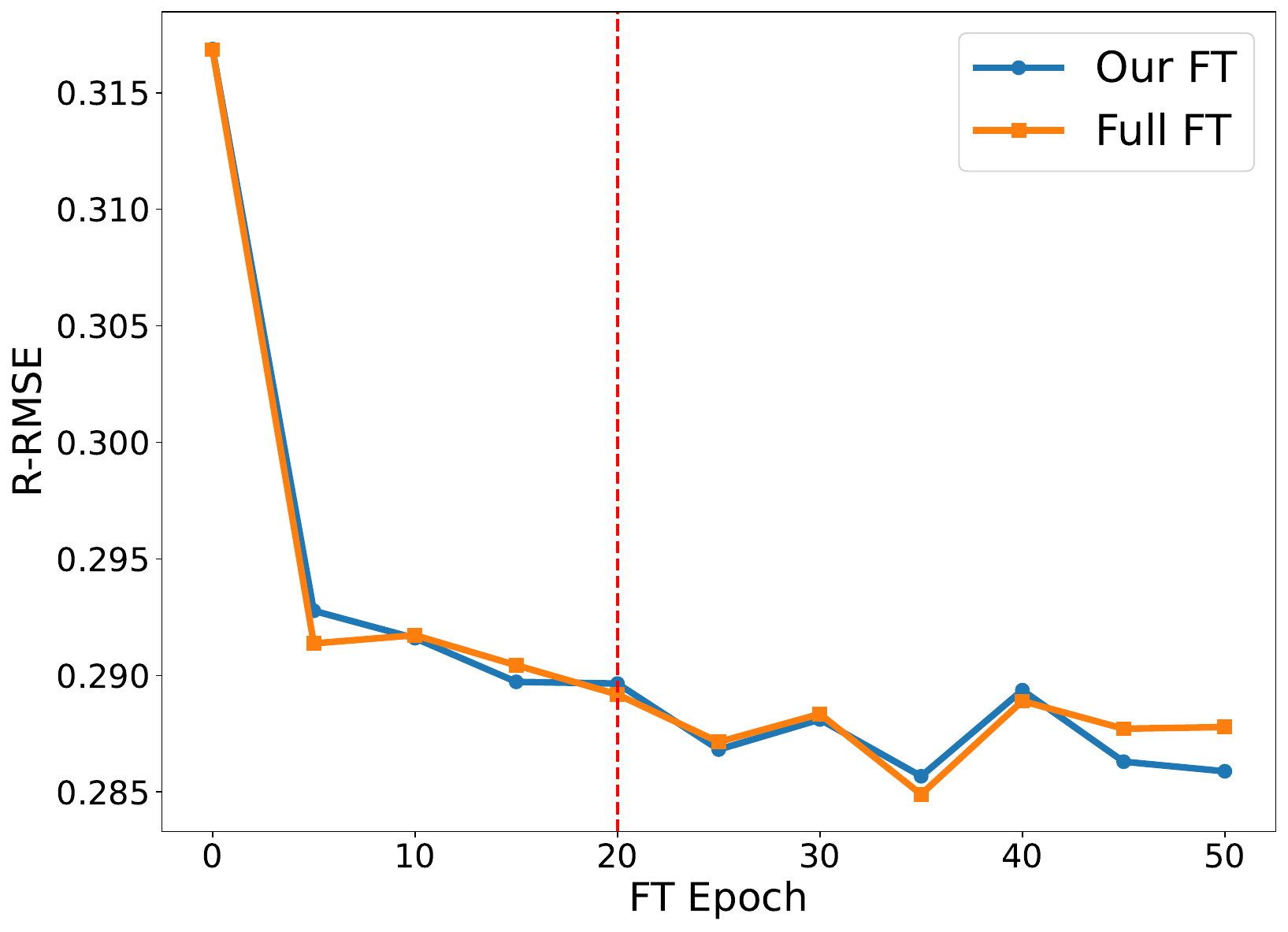}
        \caption{Ensemble Size $40$}
        \label{fig:ft_40}
    \end{subfigure}
    \caption{Comparison of R-RMSE across training epochs on the Lorenz '96 model for two fine-tuning strategies under two ensemble sizes, (a) 20 and (b) 40. The vertical dashed red line indicates  the stopping epoch of our proposed fine-tuning approach. Results demonstrate that our fine-tuning on a small subset of parameters achieves comparable RMSE performance to fine-tuning all parameters.}
    \label{fig:ft_comparison}
\end{figure}

Combining the information from Tables~\ref{tab:params_ft} and \ref{tab:time_comp}, we observe that our fine-tuning strategy demonstrates remarkable efficiency compared to the pretraining stage. As detailed in Subsections~\ref{ssec:L63}--\ref{ssec:KS}, this highly efficient fine-tuning significantly improves the model's performance, particularly when the ensemble sizes for inference and pretraining differ substantially.

To further validate the effectiveness of our proposed fine-tuning approach, we perform additional experiments comparing our method (Subsection~\ref{ssec:fine-tuning}) against alternative fine-tuning strategies. Specifically, we investigate and quantitatively compare the RMSE performance of the following fine-tuning scenarios: (1) fine-tuning all model parameters and (2) fine-tuning for a longer period (i.e., more epochs). 

Figure~\ref{fig:ft_comparison} shows RMSE results from fine-tuning experiments on the Lorenz '96 model with ensemble sizes $N'=20,40$, using a model pretrained with ensemble size $N=10$. We compare our proposed method from Subsection~\ref{ssec:fine-tuning} against fine-tuning all parameters, both running up to 50 epochs. The results demonstrate that both approaches achieve nearly identical RMSE performance, with minimal improvement observed after 20 epochs. Combined with our efficiency analysis in Tables~\ref{tab:params_ft} and \ref{tab:time_comp}, these findings confirm that fine-tuning all parameters is unnecessary, and that 20 epochs is sufficient for the near optimal performance.

\section{Conclusions}
\label{sec:CON}

In this paper, we propose a novel machine learning-based approach, MNMEF, for state estimation in data assimilation. Our method features a scheme which subsumes the ensemble Kalman filter (EnKF), as a special case, ensuring optimality for linear, Gaussian problems, while extending beyond Gaussian applications through our learnable correction terms. A key advantage of our method is its ability to be trained at one ensemble size and then directly applied, possibly with a cheap fine-tuning of a small subset of parameters, for different ensemble sizes. This property is a consequence of the mean-field interpretation of the set transformer, the machine learning architecture underlying the methodology. Indeed, in this paper we introduce neural operators acting on the metric space of probability measures, which we call \emph{measure neural mappings} (MNM). We generalize transformers to this setting and show how the set transformer itself may be viewed as a particle approximation of such a mean-field model.

Our method adopts a form similar to the Kalman filter in the mean-field perspective, with the core innovation being a parameterized gain matrix analogous to the Kalman gain (Section~\ref{sec:MFF}). Since this parameterized gain matrix depends on the current filtering distribution, we employ the set transformer neural network architecture to process such measure inputs (Section~\ref{sec:AFM}). In practical applications, our method operates on an ensemble of particles (viewed as an empirical measure) and incorporates learnable mechanisms analogous to inflation and localization techniques in EnKF-based methods, enhancing performance with smaller ensemble sizes (Section~\ref{sec:LIE}). Our experiments across multiple challenging systems (Lorenz '63, Lorenz '96, and Kuramoto-Sivashinsky) demonstrate that our proposed method consistently outperforms classical approaches including the local ensemble transform Kalman filter (LETKF), with relative improvements of 15--30\% for most scenarios (Section~\ref{sec:NUM_EXP}).

Despite the promising results, our method has limitations and several directions for future improvement. Our current approach is limited by the loss function design, which primarily addresses state estimation rather than general filtering problems. Future work will consider probabilistic loss functions that can be used to estimate the filtering distribution.

Additionally, we plan to extend our approach beyond EnKF to include other filtering schemes, such as the ensemble square root filter (ESRF), which is a deterministic version of the EnKF, or variational methods such as iterative EnKF (iEnKF). The flexibility of our approach stems from the proposed MNM-based learning architecture, which can optimize variables related to the empirical distributions represented by ensembles.

We currently formulate the transformer as a neural operator accepting probability measures as inputs. In future work, we aim to enhance both the architecture and training methodology to simultaneously handle measure and function valued inputs, enabling amortized neural operators that generalize across different dynamics and observation models. Establishing universal approximation results and statistical guarantees for this formulation will also be of interest to further strengthen the theoretical foundations of our approach.

In conclusion, our work advances the field of data assimilation by introducing a novel machine learning approach, which is efficient for deployment, requiring only one pretraining stage and lightweight fine-tuning. Beyond immediate applications in state estimation, our theoretical contribution of extending attention mechanisms to operate as measure-to-measure transformations opens new research directions across multiple disciplines.

\section{Acknowledgements}

The authors are grateful to Arnaud Doucet for pointing them to work on the set transformer. The authors acknowledge support from a Department of Defense (DoD) Vannevar Bush Faculty Fellowship (award N00014-22-1-2790),  NSF award AGS1835860 and from the Resnick Sustainability Institute; all support is held by AMS.

\bibliographystyle{elsarticle-num}
\bibliography{references}

\newpage
\appendix

\section{Overview of the Architecture and our Complete Learning Framework}\label{app:framework_overview}

In this appendix, we provide an overview of our complete learning framework. Our architecture (Subsection~\ref{ssec:architecture}) is designed to handle varying ensemble sizes efficiently while maintaining high performance through a combination of neural network-based corrections and fine-tuning strategies. The training process follows the methodology described in Subsection~\ref{ssec:training_parms}, where the model parameters are initially trained on a fixed ensemble size $N$. When our approach is applied to a different ensemble size $N' \neq N$, we perform fine-tuning using the procedure detailed in Subsection~\ref{ssec:fine-tuning}. 

Recall that in both the initial training and the fine-tuning process, we always work under Data Assumption~\ref{da:fix_params}: we fix dynamic operator $\Psi$ (time step $\Delta t$ implicitly embedded), observation operator $h$, and noise covariance matrices $\Sigma$ and $\Gamma$. We generate multiple trajectories of the state-observation system through independent realizations of the driving noise sequences.

The core of our framework (Algorithm~\ref{alg:update_ensemble}) is to update the ensemble from $\{\vn_{j}\}_{n=1}^N$ to $\{\vn_{j+1}\}_{n=1}^N$. This is a basic component for both the training and inference. Based on this, we can calculate the loss and do the initial training with the ensemble size $N$ according to Algorithm~\ref{alg:pretraining}. When we need to proceed with a different ensemble size $N'\ne N$, we follow Algorithm~\ref{alg:fine-tuning} to efficiently fine-tune part of the parameters.

After completing the pretraining (Algorithm~\ref{alg:pretraining}) on ensemble size $N$, the inference procedure depends on the desired inference ensemble size $N'$. If $N' = N$, then no additional fine-tuning is required. However, if $N' \neq N$, for optimal performance, we recommend employing Algorithm~\ref{alg:fine-tuning} to efficiently fine-tune the model before inference. Below, we describe the inference steps for the general case where $N' \neq N$ after fine-tuning has been completed.

We first load the pre-trained parameter $\theta_\mathrm{ST}^{(N)}$ that remain unchanged regardless of ensemble size, along with the fine-tuned parameters $\theta_\mathrm{gain}^{(N')}$, $\theta_\mathrm{infl}^{(N')}$, and $\theta_\mathrm{loc}^{(N')}$ that were specifically optimized for ensemble size $N'$. We then initialize the ensemble $\{v_0^{(n)}\}_{n=1}^{N'}$ according to equation~\eqref{eq:initial_ensemble}. For each time step $j$, we evolve the ensemble from $\{v_j^{(n)}\}_{n=1}^{N'}$ to $\{v_{j+1}^{(n)}\}_{n=1}^{N'}$ using Algorithm~\ref{alg:update_ensemble}. This process can be extended to observation trajectories of arbitrary length. At each time step, we can approximate the true state $v_j^\dagger$ using the ensemble mean $\bar{v}_j = \frac{1}{N'}\sum_{n=1}^{N'} v_j^{(n)}$.

\begin{algorithm}[h]
\caption{One Step Evolution of the Ensemble}
\label{alg:update_ensemble}
\begin{algorithmic}[1]
\REQUIRE Parameters $\theta = \{\theta_\mathrm{ST}, \theta_\mathrm{gain}, \theta_\mathrm{infl}, \theta_\mathrm{loc}\}$; current ensemble $\{v^{(n)}_j\}_{n=1}^N$.
\ENSURE Ensemble for the next time step $\{v^{(n)}_{j+1}\}_{n=1}^N$.
\STATE Get predicted states $\{\hvn_{j+1}\}_{n=1}^N$ from $\{v_j^{(n)}\}_{n=1}^N$ \eqref{eq:ours_ens_pred}.
\STATE Get predicted observations $\{\hyn_{j+1}\}_{n=1}^N$ from $\{\hvn_{j+1}\}_{n=1}^N$ \eqref{eq:ours_ens_extend}.
\STATE Process the ensemble into a feature vector according to \eqref{eq:FST}:
\vspace{-0.3cm}
$$f_v = F^\mathrm{ST}\bigl(\{(\hvn_{j+1},h(\hvn_{j+1}))\}_{n=1}^N;\theta_\mathrm{ST}\bigr).$$
\vspace{-0.7cm}
\STATE Calculate the correction terms according to \eqref{eq:nn_F}:
\vspace{-0.3cm}
$$\bigl(\hwn_\theta, \hzn_\theta\bigr) = F^\mathrm{gain}\bigl(\hvn_{j+1}, h(\hvn_{j+1}), y_{j+1}^\dagger, f_v; \theta_\mathrm{gain}\bigr).$$
\vspace{-0.7cm}
\STATE Calculate $(\Kt^{(1)})_{j+1}$ and $(\Kt^{(2)})_{j+1}$ \eqref{eq:full_K}.
\STATE Process with the neural network for localization:
\vspace{-0.3cm}
\begin{align*}
    \hg_{j+1} = F^\mathrm{loc}\left(f_v; \theta_\mathrm{loc}\right) \text{ according to \eqref{eq:f_loc}}.
\end{align*}
\vspace{-0.7cm}
\STATE Calculate localization matrices $(\Lt^{(1)})_{j+1}$ and $(\Lt^{(2)})_{j+1}$ (Subsection~\ref{sssec:localization}).
\STATE Calculate the Gain matrix $(\Kt)_{j+1}$ with localization \eqref{eq:loc_k-theta}.
\STATE Get $\{v_{j+1}^{(n)}\}_{n=1}^N$ according to the analysis step:
\vspace{-0.3cm}
$$\vn_{j+1} = \hvn_{j+1} + (\Kt)_{j+1} \left( y^\dagger_{j+1} - \hyn_{j+1} \right)$$
\vspace{-0.7cm}
\STATE Process with the neural network for inflation:
\vspace{-0.3cm}
\begin{align*}
    \hun_\theta = F^\mathrm{infl}(\vn_{j+1}, f_v;\theta_\mathrm{infl}) \text{ according to \eqref{eq:f_infl}}.
\end{align*}
\vspace{-0.7cm}
\STATE Update the estimation with the learned inflation term $\hun_\theta$ by
\vspace{-0.3cm}
\begin{align*}
    \vn_{j+1}\gets \vn_{j+1} + \hun_\theta.
\end{align*}
\vspace{-0.7cm}
\RETURN Ensemble $\{v^{(n)}_{j+1}\}_{n=1}^N$.
\end{algorithmic}
\end{algorithm}

\begin{algorithm}[h]
\caption{Pretraining}
\label{alg:pretraining}
\begin{algorithmic}[1]
\REQUIRE $J, M\in \bbN$; ensemble size $N$.
\ENSURE Trained parameters $\theta^{(N)} = \{\theta_\mathrm{ST}^{(N)}, \theta_\mathrm{gain}^{(N)}, \theta_\mathrm{infl}^{(N)}, \theta_\mathrm{loc}^{(N)}\}$.
\STATE \textbf{Training Data Generation:} Generate $M$ training trajectories with length $J+1$ and corresponding observations according to Subsection~\ref{ssec:training_parms}.
\FOR{each epoch, each minibatch indexed by $M_B\subset [M]$}
    \FOR{$m\in M_B$ and the corresponding trajectory $\{v^\dagger_j\}_{j=1}^J$}
        \STATE Initialize the ensemble $\{v_0^{(n)}\}_{n=1}^N$ \eqref{eq:initial_ensemble}.
        \FOR{$j=0,1,\ldots,J-1$}
            \STATE Update from $\{v_j^{(n)}\}_{n=1}^{N}$ to $\{v_{j+1}^{(n)}\}_{n=1}^{N}$ according to Algorithm~\ref{alg:update_ensemble}.
        \ENDFOR
        \STATE Calculate the loss $\mathcal{L}_m$ \eqref{eq:loss}.
    \ENDFOR
    \STATE Calculate the batch loss $\mathcal{L}_{M_B}$ \eqref{eq:batch_loss}.
    \STATE Update $\theta$ via gradient descent on $\mathcal{L}_{M_B}$.
\ENDFOR
\RETURN Optimized parameters $\theta^{(N)} = \{\theta_\mathrm{ST}^{(N)}, \theta_\mathrm{gain}^{(N)}, \theta_\mathrm{infl}^{(N)}, \theta_\mathrm{loc}^{(N)}\}$
\end{algorithmic}
\end{algorithm}

\begin{algorithm}[h!]
\caption{Fine-tuning}
\label{alg:fine-tuning}
\begin{algorithmic}[1]
\REQUIRE Ensemble size $N'\ne N$; pretrained parameter $\theta_\mathrm{ST}^{(N)}$
\ENSURE Updated parameters $\theta_\mathrm{gain}^{(N')}, \theta_\mathrm{infl}^{(N')}, \theta_\mathrm{loc}^{(N')}$ for new ensemble size $N'$
\STATE Use the same synthetic training data as Algorithm~\ref{alg:pretraining}.
\FOR{each epoch, each minibatch indexed by $M_B\subset [M]$}
    \FOR{$m\in M_B$ and the corresponding trajectory $\{v^\dagger_j\}_{j=1}^J$}
        \STATE Initialize the ensemble $\{v_0^{(n)}\}_{n=1}^{N'}$ with size $N'$\eqref{eq:initial_ensemble}.
        \FOR{$j=0,1,\ldots,J-1$}
            \STATE Update from $\{v_j^{(n)}\}_{n=1}^{N'}$ to $\{v_{j+1}^{(n)}\}_{n=1}^{N'}$ according to Algorithm~\ref{alg:update_ensemble}.
        \ENDFOR
        \STATE Calculate the loss $\mathcal{L}_m^{(N')}$ with ensemble size $N'$ \eqref{eq:loss}.
    \ENDFOR
    \STATE Calculate the batch loss $\mathcal{L}_{M_B}^{(N')}$ \eqref{eq:batch_loss} with ensemble size $N'$
    \STATE Update $\theta_\mathrm{gain}^{(N')}, \theta_\mathrm{infl}^{(N')}, \theta_\mathrm{loc}^{(N')}$ according to \eqref{eq:ft_problem}.
\ENDFOR
\RETURN Optimized parameters $\theta_\mathrm{gain}^{(N')}, \theta_\mathrm{infl}^{(N')}, \theta_\mathrm{loc}^{(N')}$
\end{algorithmic}
\end{algorithm}

\section{Description of Multihead Attention}\label{app:multiheadatt}
In Subsection~\ref{ssec:discrete_ST}, we introduce the definition of attention. However, in practical applications, multihead attention is commonly employed instead of the single-head variant. This appendix provides supplementary information on multihead attention, which is used in the set transformer architecture (Subsection~\ref{ssec:discrete_ST}) rather than standard single-head attention.

Recall the definition of the set of sequence of a finite length, $\mathcal{U}_F(\bbR^d) = \cup_{N=1}^\infty\mathcal{U}([N];\bbR^{d})$. In Subsection~\ref{ssec:discrete_ST}, we defined attention as a sequence-to-sequence operator $\sA: \mathcal{U}_F(\bbR^{d_u})\times\mathcal{U}_F(\bbR^{d_w})\rightarrow \mathcal{U}_F(\bbR^{d_v})$ \eqref{eq:seq_fA} with learnable parameters $\theta(\sA) = \{Q,K,V\}$. For two sequences $u\in \mathcal{U}([N];\bbR^{d_u})$ and $w\in \mathcal{U}([M];\bbR^{d_w})$, we have 
\begin{equation}\label{eq:app_att}
    \sA(u, w)(j) = \frac{\sum_{j=1}^M\exp\bigl(\langle Qu(j),Kw(k)\rangle\bigr)Vw(k)}{\sum_{\ell=1}^M\exp\bigl(\langle Qu(j),Kw(\ell)\rangle\bigr)},\quad \forall j\in [N].
\end{equation}

In multihead attention, we employ several parallel attention heads, each with its own set of learnable parameters. This approach allows the model to jointly attend to information from different representation subspaces at different positions. Let us denote the multihead attention operator as $\sA^R(\placeholder,\placeholder)$, where the superscript $R$ indicates the number of attention heads. For each head $r \in [R]$, we have a separate set of learnable parameters $\theta(\sA_r) = \{Q_r, K_r, V_r\}$, where $Q_r \in \mathbb{R}^{d_k \times d_u}$, $K_r \in \mathbb{R}^{d_k \times d_w}$, and $V_r \in \mathbb{R}^{\hat{d}_V \times d_w}$. 

For each head $r$, we compute the attention output $\sA_r(u, w)$ according to \eqref{eq:app_att}. The multihead attention mechanism $\sA^R(u, w)$ combines the outputs from $R$ heads through concatenation followed by linear projection. For $j\in [N]$, we compute the output as
\begin{equation}
\sA^R(u, w)(j) = W^O \bigl[\sA_1(u, w)(j);\sA_2(u, w)(j);\ldots;\sA_R(u, w)(j)\bigr],
\end{equation}
where $\bigl[\sA_1(u, w)(j);\sA_2(u, w)(j);\ldots;\sA_R(u, w)(j)\bigr]\in\mathbb{R}^{R \hat{d}_V}$ is a vector vertically concatenated from the outputs from all attention heads, and $W^O \in \mathbb{R}^{d_V \times (R \hat{d}_V)}$ is an additional learnable parameter matrix that projects the concatenated outputs to the desired dimension.

The complete set of learnable parameters for the multihead attention mechanism is therefore:
\begin{equation}
    \theta(\sA^R) = \{Q_r, K_r, V_r\}_{r=1}^R \cup \{W^O\}.
\end{equation}

In Section~\ref{sec:AFM}, we view attention as an operator on probability measures. This formulation naturally extends to multihead attention. The multihead attention operator acts on probability measures similarly to standard attention, but utilizes $R$ parallel attention heads. In addition, we introduce two major properties of the set transformer architecture in Subsection~\ref{ssec:discrete_ST}, i.e. the permutation invariance and the adaptation to variable input lengths. These two properties are preserved for the version using multihead attention, since multihead attention with $R$ heads effectively performs $R$ standard attention operations in parallel and concatenates their results.

\section{Implementation Details}\label{app:imp_details}
In this appendix, we provide additional implementation details to complement the methods presented in Section~\ref{sec:LIE} and their application in the numerical experiments of Section~\ref{sec:NUM_EXP}. 

The complete source code for reproducing our method is publicly available\footnote{\url{https://github.com/wispcarey/DALearning}}. Additionally, for the benchmark methods, we performed extensive hyperparameter optimization using grid search \footnote{\url{https://github.com/wispcarey/DapperGridSearch}}, implemented with the DAPPER package \cite{raanes_dapper_2024}. 

\subsection{Feature Dimensions}\label{app:nn_details}
We provide detailed information about the latent dimensions used in our learning-based approach. While this information can be directly obtained from our code, we include this analysis to highlight two important properties of the set transformer architecture employed in our method: (1) the output is invariant to permutations of the input sequence, and (2) the output dimension remains consistent regardless of the length of the input sequence. 

We introduce the general set transformer architecture in Subsection~\ref{ssec:discrete_ST}, where the encoder and decoder can include several self-attention blocks (SAB) $\sSST(\placeholder)$ \eqref{eq:st_architecture}. In our experiments (Section~\ref{sec:NUM_EXP}), our actual architecture contains two SABs in both the encoder and decoder. This simplified structure provides sufficient representational capacity while maintaining computational efficiency. Our actual set tranformer architecture can be written as
\begin{equation}\label{eq:st_actual}
    \sFST(u) = \sFNN_2(\placeholder)\circ\sFCat(\placeholder)\circ\sSST_4(\placeholder)\circ\sSST_3(\placeholder)\circ\sCST(s,\placeholder)\circ\sSST_2(\placeholder)\circ\sSST_1(\placeholder)\circ\sFNN_1(u),
\end{equation}
where $\sFNN_1$ and $\sFNN_2$ are multilayer perceptrons consisting of multiple feedforward layers and activation layers \eqref{eq:NN}; $\sSST_{1,2}$ and $\sSST_{3,4}$ are SABs serving as the encoder and decoder respectively; $\sFCat$ is a layer that concatenate all elements in a sequence into a long feature vector. Note that layers $\sSST_1$, $\sSST_2$, $\sCST$, $\sSST_3$ and $\sSST_4$ contain attention mechanisms, which are implemented as multihead attention with 8 heads (\ref{app:multiheadatt}).

\begin{table}[htbp]
\centering
\caption{Feature dimensions at each layer of the set transformer architecture to process the ensemble states for different dynamical systems, Lorenz '63 (L'63) \eqref{eq:L63_dynamic}, Lorenz '96 (L'96) \eqref{eq:L96_equations}, and Kuramoto-Sivashinsky (KS) \eqref{eq:KS_equation}. We use $(N,d)$ to denote a sequence in $\bbR^d$ with length $N$ (i.e. in $\mathcal{U}([N],\bbR^d)$). The trainable seed is a sequence in $\bbR^{64}$ with length $16$. The input sequence dimension is the sum of state dimension and the observation dimension.}
\label{tab:feature_dimensions}
\begin{tabular}{lccccccccc}
    \toprule
    \multirow{2}{*}{Dataset} &  \multicolumn{7}{c}{Feature Dimensions after Each Layer} \\
    \cmidrule{2-10}
    & Input & $\sFNN_1$ & $\sSST_1$ & $\sSST_2$ & $\sCST(s,\placeholder)$ & $\sSST_3$ & $\sSST_4$ & $\sFCat$ & $\sFNN_2$ \\
    \midrule
    L'63 & (N,4) & \multirow{3}{*}{(N,64)} & \multirow{3}{*}{(N,64)} & \multirow{3}{*}{(N,64)}  & \multirow{3}{*}{(16,64)} & \multirow{3}{*}{(16,64)} & \multirow{3}{*}{(16,64)} & \multirow{3}{*}{1024}& \multirow{3}{*}{64} \\
    L'96 & (N,50) & & & & & & & & \\
    KS & (N,144) & & & & & & & & \\
    \bottomrule
\end{tabular}
\end{table}

Table~\ref{tab:feature_dimensions} provides the specific layer-wise dimension design in our architecture. We adopt similar network structures for all three experimental datasets: Lorenz '63 (L'63)~\eqref{eq:L63_dynamic}, Lorenz '96 (L'96)~\eqref{eq:L96_equations}, and Kuramoto-Sivashinsky (KS)~\eqref{eq:KS_equation}. For all experiments, we use a trainable seed $s \in \mathcal{U}([16],\mathbb{R}^{64})$, which is a sequence of length 16 in $\bbR^{64}$. The first linear layer $\sFNN_1$ maps input sequences of varying dimensions to $\mathbb{R}^{64}$, after which all subsequent layers maintain consistent dimensions across different dataset dynamics. 

Table~\ref{tab:feature_dimensions} clearly illustrates the fact that the set transformer can process input sequences of arbitrary length and output fixed-dimension feature vectors. Specifically, regardless of the input sequence length $N$, after processing through the PMA block $\sCST(s,\placeholder)$ \eqref{eq:PMA}, the sequence length always aligns with the seed length. This critical dimensional reduction occurs because the cross-attention mechanism in $\sCST(s,\placeholder)$ uses the fixed-length seed $s$ as queries to attend to the input sequence. The output dimension is then determined by the concluding $\sFNN_2$ layer. For detailed computational mechanisms, please refer to Section~\ref{sec:AFM} and Subsection~\ref{ssec:discrete_ST}.

The table also helps illustrate where the permutation invariance is established in the architecture. Before the PMA $\sCST(s,\placeholder)$, the latent features are permutation equivariant with respect to the input, meaning that if the input sequence elements are reordered, the corresponding features will be reordered in the same way. However, after the PMA block, the feature ordering becomes entirely dependent on the seed sequence and completely independent of the input sequence ordering. This transformation from permutation equivariance to permutation invariance is a crucial property (Proposition~\ref{prop:permutation_invariance}) of the set transformer, enabling them to process sets rather than sequences.

\subsection{An Example of Learning the Localization Weight Matrix}
\label{app:learn_loc}

Here we provide a concrete example of how to compute the parameterized localization weight matrices $\Lt^{(1)}$ and $\Lt^{(2)}$ in the Lorenz ’96 system (Subsection~\ref{ssec:L96}) with state dimension $d_v=40$ and observation dimension $d_y=10$. Specifically, the 40 state variables are indexed by ${1,2,\ldots,40}$ in a cyclic manner, and the observations are available at indices ${4,8,12,\ldots,40}$ (i.e., every 4th state variable). We employ a periodic distance metric for all index pairs $k, \ell \in [d_v]$,
\begin{equation} \label{eq:periodic_dist} 
d_\mathrm{peri}(k, \ell) = \min\bigl(\lvert k - \ell\rvert, 40 - \lvert k - \ell\rvert\bigr). 
\end{equation}
Then we can consider all the unique distance values given by:
\begin{equation}
    \{D_{k\ell} = d_\mathrm{peri}(k,\ell)|k,\ell\in [40]\} = \{0,1,\ldots,20\}.
\end{equation}
Recall that $\circ$ denotes the Hadamard pointwise matrix product. In the classic localization setting \eqref{eq:EnKF_loc} for the EnKF approach, 
\begin{equation}
    K = \hat{C}^{vh}\circ L^{vh}\left(\hat{C}^{hh} \circ L^{hh} + \Gamma\right)^{-1},
\end{equation}
we have 
\begin{equation}
\begin{split}
    (L^{vh})_{k,\ell} &= g(d_\mathrm{peri}(k,4\ell)),\quad k=1,2,\ldots,40,\quad \ell=1,2,\ldots,10,\\
    (L^{hh})_{k,\ell} &= g(d_\mathrm{peri}(4k,4\ell)),\quad k=1,2,\ldots,10,\quad \ell=1,2,\ldots,10.
\end{split}
\end{equation}
In our learning framework introduced in Subsection~\ref{sssec:localization}, we learn a vector $\hg\in\bbR^{21}$ according to \eqref{eq:f_loc} for the function values $g_\theta(D)$ for $D=0,1,2,\ldots,20$. Then in our proposed scheme \eqref{eq:loc_k-theta},
\begin{equation}
    \Kt = \Kt^{(1)}\circ \Lt^{(1)} \left( \Kt^{(2)}\circ \Lt^{(2)} + \Gamma \right)^{-1},
\end{equation}
we can calculate the parameterized localization weight matrices by
\begin{equation}
\begin{split}
    (\Lt^{(1)})_{k,\ell} &= g_\theta(d_\mathrm{peri}(k,4\ell)),\quad k=1,2,\ldots,40,\quad \ell=1,2,\ldots,10,\\
    (\Lt^{(2)})_{k,\ell} &= g_\theta(d_\mathrm{peri}(4k,4\ell)),\quad k=1,2,\ldots,10,\quad \ell=1,2,\ldots,10.
\end{split}
\end{equation}

\subsection{Initial Conditions}
\label{app:initial_conditions}
Here we provide details of the initial conditions for the experiments in Section~\ref{sec:NUM_EXP}. To ensure our initial states lie on the attractor, we first select a base value and then evolve the system's dynamics forward for a substantial number of time steps. This serves as a ``burn-in" period. The initial conditions detailed below for each system are subsequently used to generate all trajectories for pretraining, fine-tuning, and testing.
\begin{enumerate}
    \item \textbf{Lorenz '96 (Subsection~\ref{ssec:L96})}: Sample $x_0\sim \normal(5, I_{40})$ and run the forward dynamic \eqref{eq:L96_equations} for random $10^3$ to $5\times 10^5$ time steps with step size $\Delta t = 0.15$ to get $\vd_0$. Within each $\Delta t$ time interval, there are $5$ RK4 numerical integration steps with the step size $\Delta t / 5 = 0.03$. The initial distribution is set to be $\normal (\vd_0, I_{40})$. 
    \item \textbf{Kuramoto–-Sivashinsky (Subsection~\ref{ssec:KS})}: Set the interval $[0,L]$ as $L=32\pi$. For $u_0(x) = \cos(2x/L)(1 + \sin(2x/L))$, run the forward dynamic \eqref{eq:KS_equation} for random $10^3$ to $5\times 10^5$ time steps with step size $\Delta t = 1.00$ to get $\hat{u}$. Within each $\Delta t$ time interval, there are $4$ RK4 numerical integration steps with the step size $\Delta t / 4 = 0.25$. The initial state $\vd_0 = (\hat{u}(x_0),\ldots,\hat{u}(x_{127}))$ with $x_\ell = \ell L / 128$, $\ell=0,1,\ldots,127$. The initial distribution is set to be $\normal (\vd_0, I_{128})$.
    \item \textbf{Lorenz '63 (Subsection~\ref{ssec:L63})}: Sample $x_0\sim \normal(0, I_{3})$ and run the forward dynamic \eqref{eq:L63_dynamic} for random $10^3$ to $5\times 10^5$ time steps with step size $\Delta t = 0.15$ to get $\vd_0$. Within each $\Delta t$ time interval, there are $5$ RK4 numerical integration steps with the step size $\Delta t / 5 = 0.03$. The initial distribution is set to be $\normal (\vd_0, I_{3})$. 
\end{enumerate}

\subsection{Hyperparameter Setting for Our Training}
\label{app:hyperparameter_for_training}

We provide details of the training hyperparameters in experiments. The parameters listed in Table \ref{tab:Training_Hyperparameters} are utilized for pretraining with an ensemble size of $N=10$ in the experiments detailed in Section~\ref{sec:NUM_EXP}. When fine-tuning with a different ensemble size $N' \ne N$ (Subsection~\ref{ssec:fine-tuning}), the number of training trajectories is halved compared to the quantity used for pretraining. Furthermore, the learning rate for fine-tuning is set to $1/10$ of the learning rate used during pretraining. We use the AdamW optimizer \cite{loshchilov2017decoupled} for both the pretraining and fine-tuning.

During the training process, we implement a clamping mechanism to ensure numerical stability and prevent issues such as exploding estimated trajectories. If the absolute value of any dimension of an estimated particle exceeds the specified clamp value (provided in the ``Clamp'' column of Table \ref{tab:Training_Hyperparameters}), its sign is preserved, but its absolute value is replaced by the clamp threshold.

\begin{table}[htbp]
\centering
\caption{Training hyperparameters for pretraining with an ensemble size N=10. For fine-tuning with N' $\ne$ N, the number of training trajectories is halved, and the learning rate is reduced to $1/10$ of the values listed. Obs $\Delta t$ refers to the observation time step.}
\label{tab:Training_Hyperparameters}
\begin{tabular}{lcccccccc}
    \hline
    \multirow{2}{*}{Dataset} &  \multicolumn{2}{c}{Train Traj} & \multicolumn{2}{c}{Test Traj} & Obs & Learning & Batch & \multirow{2}{*}{Clamp}\\
    \cline{2-5}
    & Num & Length & Num & Length & $\Delta t$ & Rate & Size &  \\
    \hline
    Lorenz '96 & 8192 & 60 & 64 & 1500 & 0.15 & 1e-3 & 512 & 20 \\
    KS & 8192 & 60 & 64 & 2000 & 1.00 & 5e-4 & 256 & 10 \\
    Lorenz '63 & 8192 & 60 & 64 & 1500 & 0.15 & 1e-3 & 1024 & 60\\
    \bottomrule
\end{tabular}
\end{table}

\subsection{Hyperparameter Optimization via Grid Search}
\label{app:grid_search}
\begin{figure}[htbp]
\centering

\begin{subfigure}{0.45\textwidth}
    \centering
    \includegraphics[width=\textwidth]{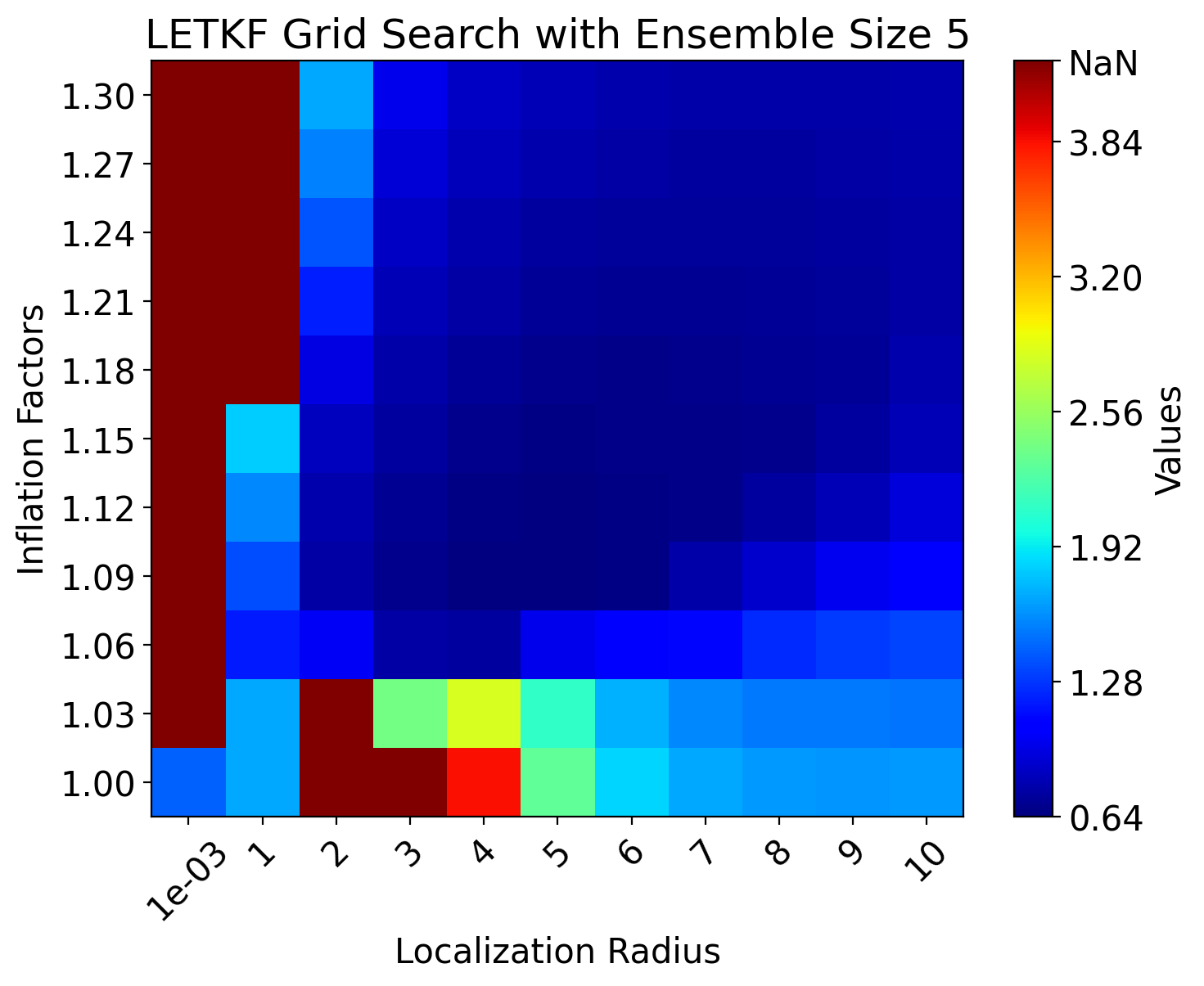}
    \caption{$N=5$}
    \label{fig:letkf_5}
\end{subfigure}
\hfill
\begin{subfigure}{0.45\textwidth}
    \centering
    \includegraphics[width=\textwidth]{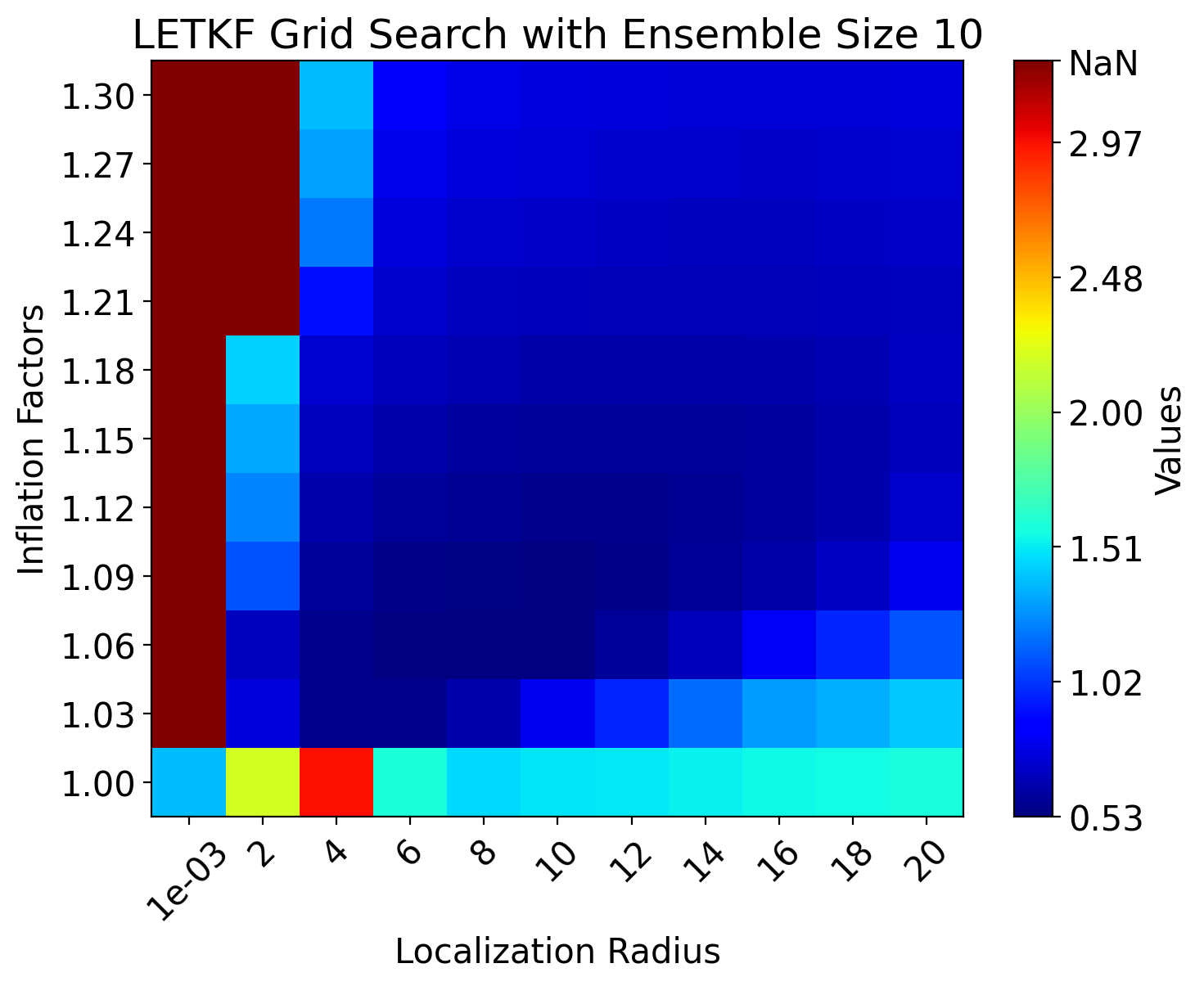}
    \caption{$N=10$}
    \label{fig:letkf_10}
\end{subfigure}

\begin{subfigure}{0.45\textwidth}
    \centering
    \includegraphics[width=\textwidth]{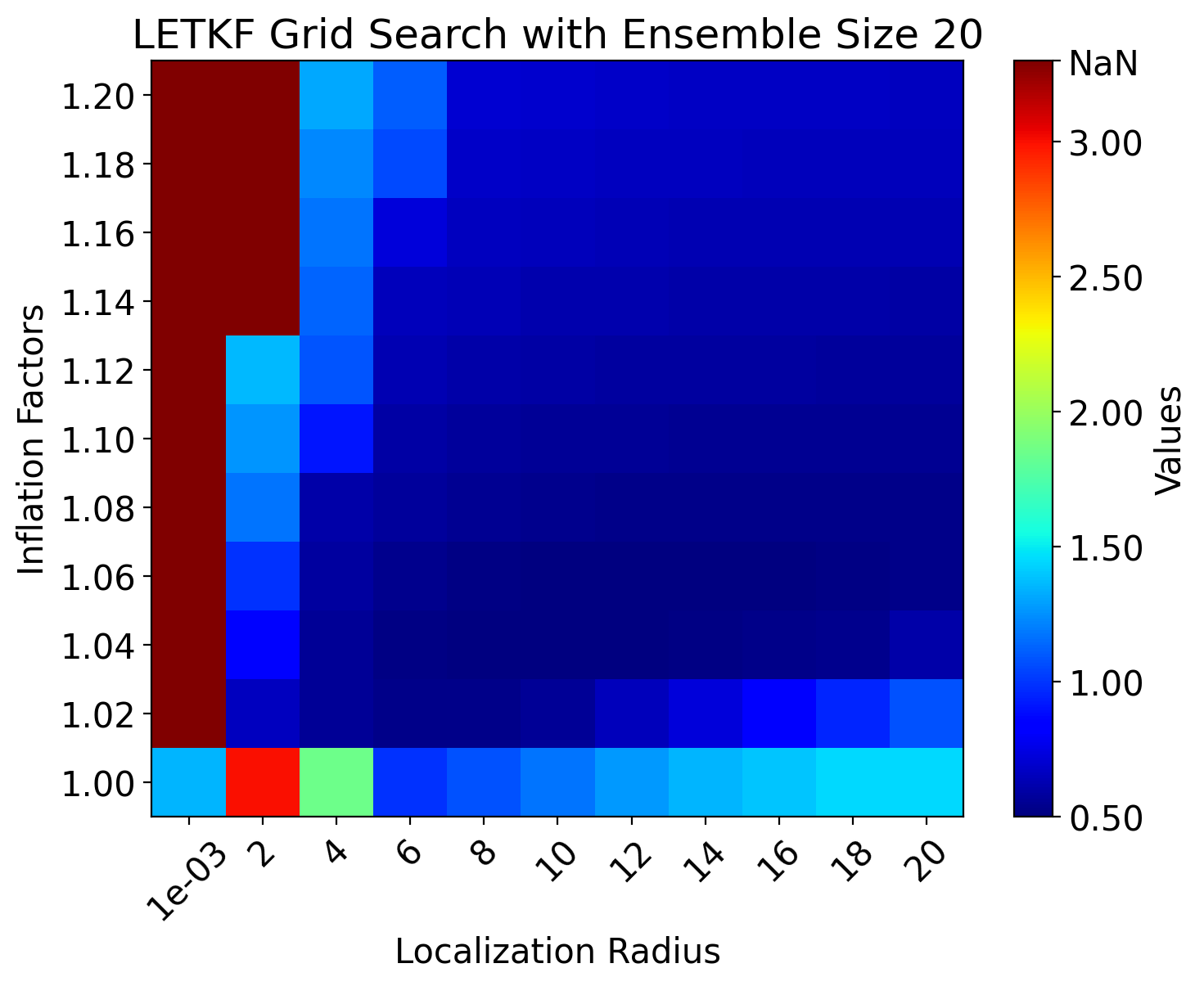}
    \caption{$N=20$}
    \label{fig:letkf_20}
\end{subfigure}
\hfill
\begin{subfigure}{0.45\textwidth}
    \centering
    \includegraphics[width=\textwidth]{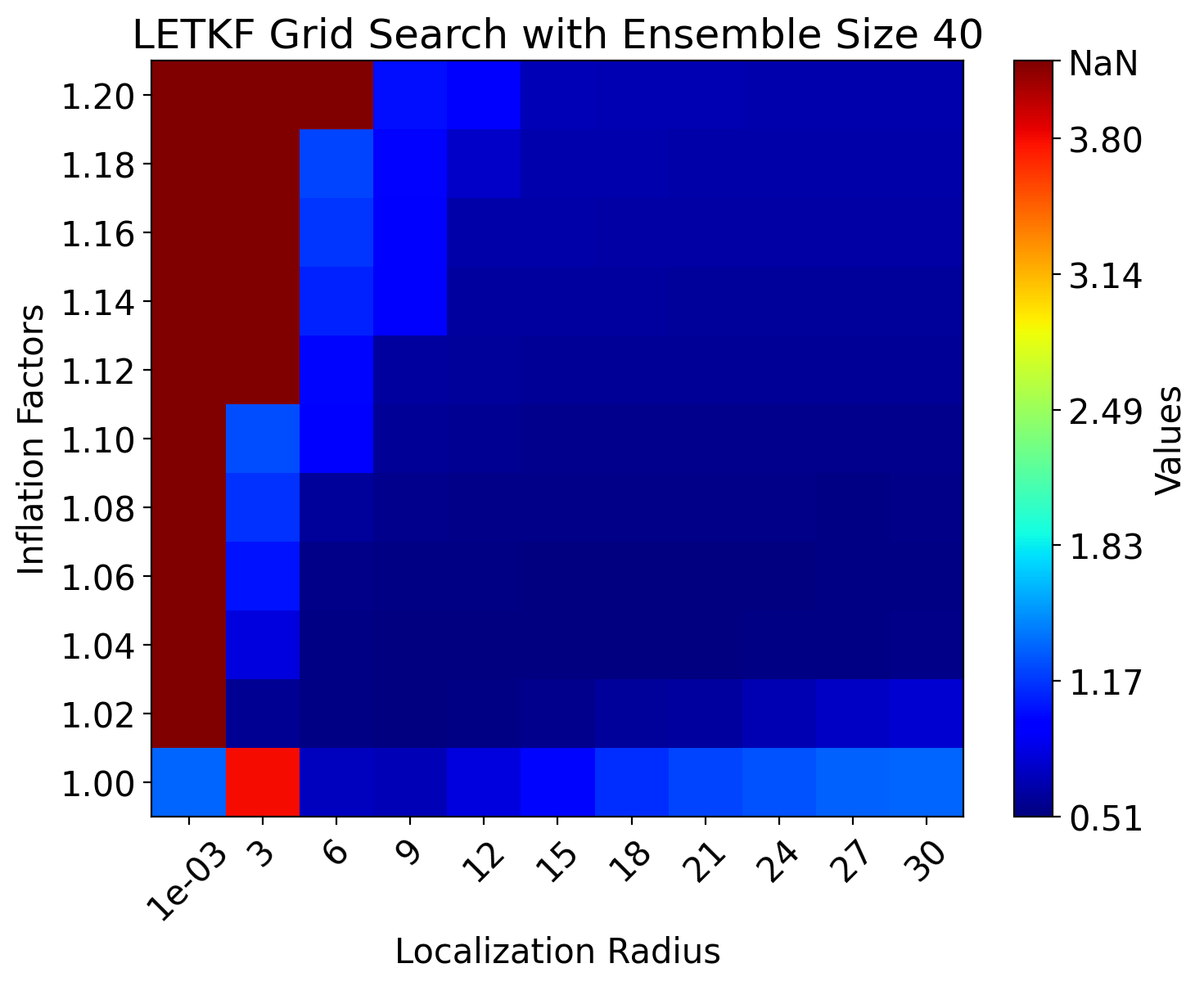}
    \caption{$N=40$}
    \label{fig:letkf_40}
\end{subfigure}

\begin{subfigure}{0.45\textwidth}
    \centering
    \includegraphics[width=\textwidth]{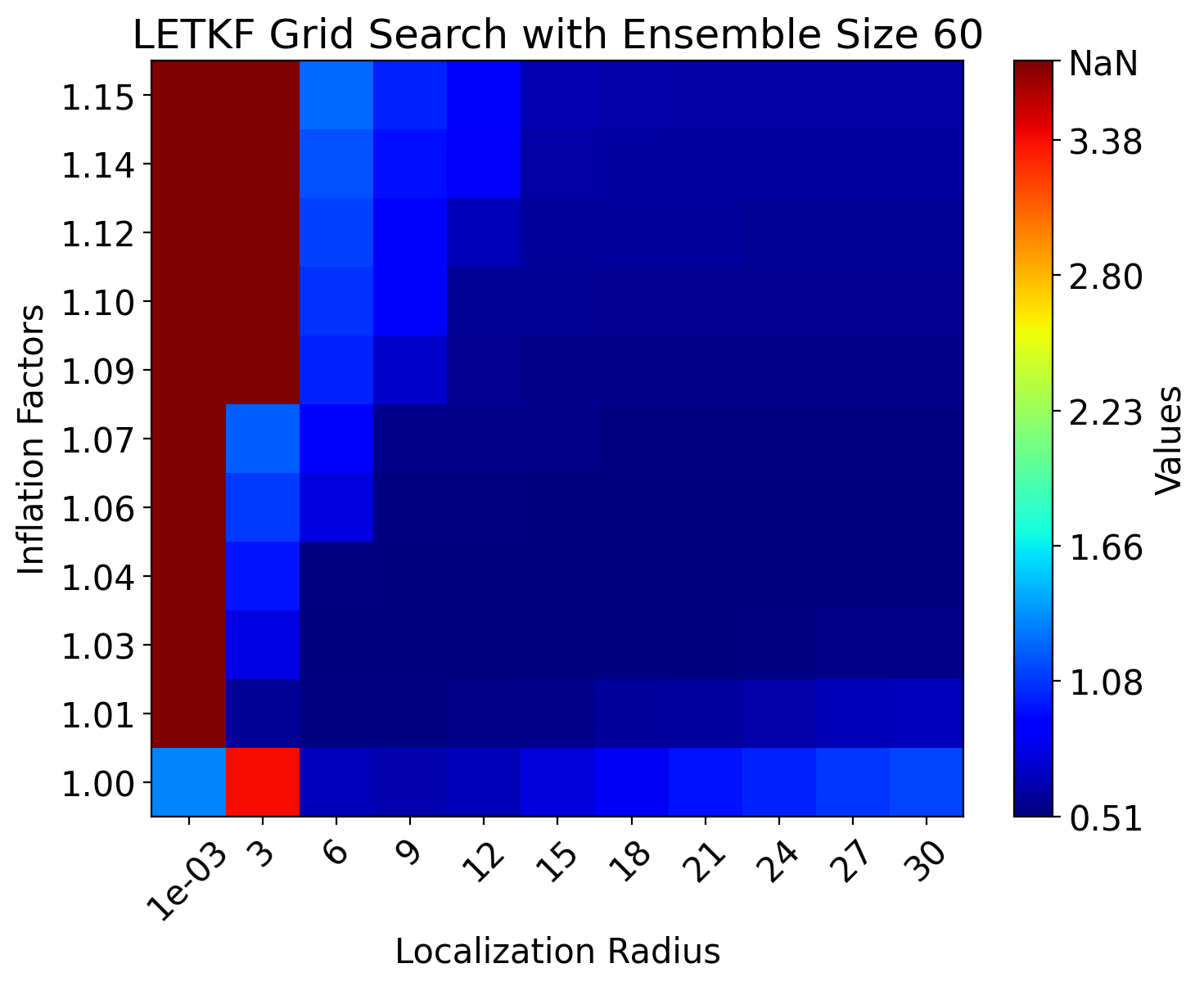}
    \caption{$N=60$}
    \label{fig:letkf_60}
\end{subfigure}
\hfill
\begin{subfigure}{0.45\textwidth}
    \centering
    \includegraphics[width=\textwidth]{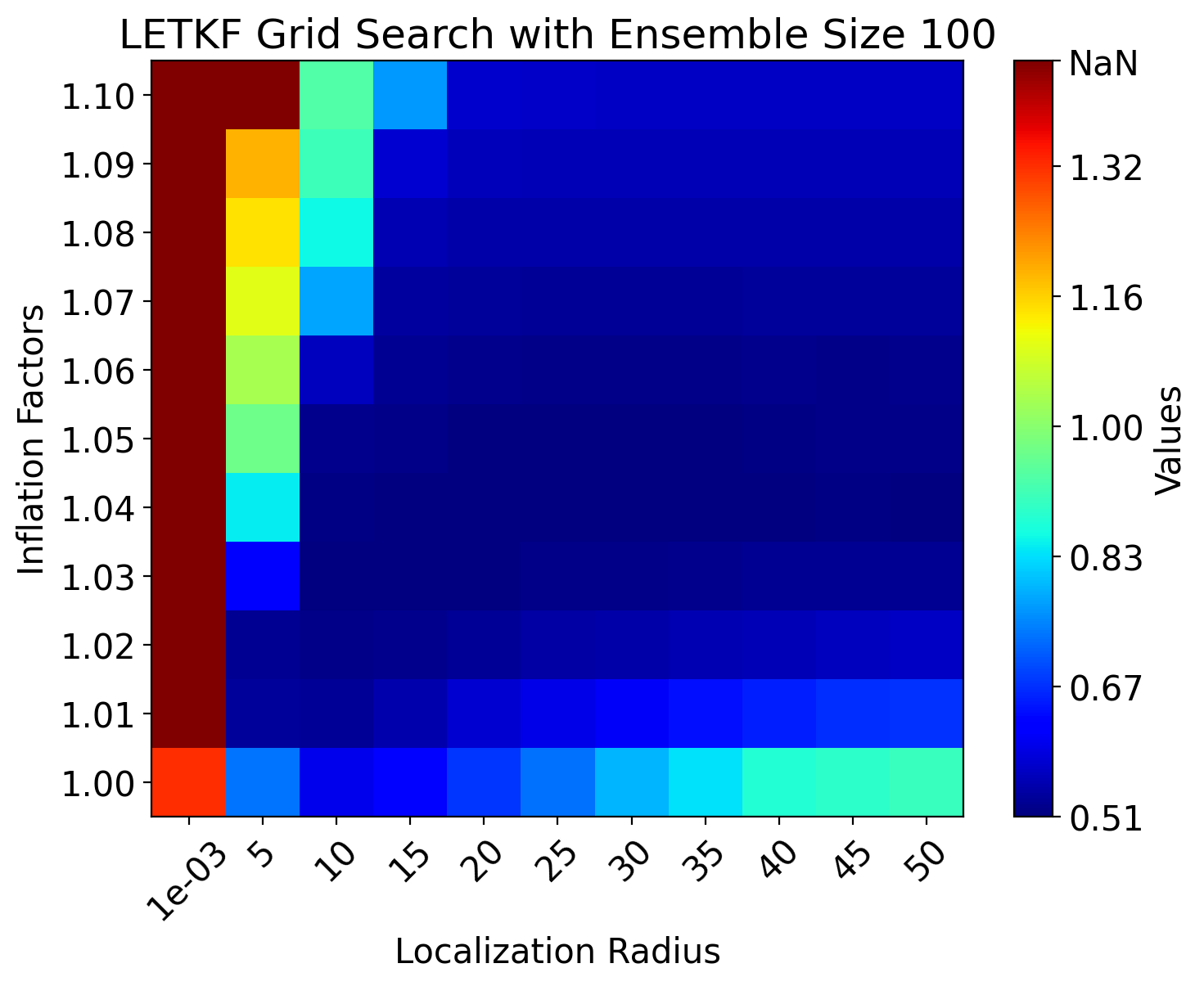}
    \caption{$N=100$}
    \label{fig:letkf_100}
\end{subfigure}
\caption{Grid search results for the Local Ensemble Transform Kalman Filter (LETKF) \cite{hunt2007efficient} on the Kuramoto--Sivashinsky (KS) dynamical system \eqref{eq:KS_equation} varying ensemble sizes. Each subplot shows the RMSE performance across different combinations of inflation parameter $\alpha$ and localization radius parameter $r$. Darker colors indicate lower relative RMSE (R-RMSE) values \eqref{eq:avg_r_rmse} (better performance).}
\label{fig:letkf_grid_search}
\end{figure}

In our experiments (Section~\ref{sec:NUM_EXP}), we compare our pretrained and fine-tuned methods introduced in Section~\ref{sec:LIE} with benchmarks, EnKF \cite{burgers1998analysis}, ESRF \cite{tippett_ensemble_2003}, LETKF \cite{hunt2007efficient}, and iEnKF\cite{sakov2012iterative}, detailed in Subsection~\ref{ssec:ESU}.
The benchmark methods we compared require hyperparameter tuning for different ensemble sizes so that they are optimized at each ensemble size and hence indeed present a challenging benchmark for our proposed methodology. We use the DAPPER package \cite{raanes_dapper_2024}, which implements these benchmark methods, allowing us to focus solely on conducting grid searches for the required hyperparameters across various ensemble sizes. This appendix presents examples of our grid search results to provide insight into the comprehensive range of our hyperparameter optimization. Complete results can be found in our GitHub repository\footnote{\url{https://github.com/wispcarey/DapperGridSearch}}.

For each of the four benchmark methods mentioned above, we perform grid search on the inflation parameter $\alpha > 1$. After the analysis step, the state $v^{(n)}$ is updated according to:
\begin{equation}
v^{(n)} \gets v^{(n)} + (\alpha - 1)\left(v^{(n)} - \frac{1}{N}\sum_{\ell=1}^Nv^{(\ell)}\right),\quad n=1,2,\ldots,N.
\end{equation}

In addition to inflation, for LETKF specifically, we consider the localization radius parameter $r$, which is used in the Gaspari--Cohn (GC) function \cite{gaspari1999construction} to map distances to localization weights. 

For different methods, datasets, and ensemble sizes, we employ an adaptive approach to determine the grid search step sizes and ranges. Our evaluation metric is the average relative RMSE (R-RMSE) \eqref{eq:avg_r_rmse} computed over 64 test trajectories. Lower values of this metric indicate superior performance. Initially, we establish a consistent range and step size across all scenarios, then refine these parameters based on preliminary performance results in each specific case. This adaptive refinement allows us to concentrate computational resources on the most promising hyperparameter regions.

Figure~\ref{fig:letkf_grid_search} illustrates our grid search results for the Local Ensemble Transform Kalman Filter (LETKF) \cite{hunt2007efficient} method applied to the Kuramoto--Sivashinsky (KS) dynamical system \eqref{eq:KS_equation} across different ensemble sizes ($N=5, 10, 20, 40, 60, 100$). Each heat map displays the R-RMSE performance for various combinations of inflation parameter $\alpha$ and localization radius $r$. 

\subsection{Additional Experiments: Linear Setting}\label{app:lin}

Our MNMEF aims to learn correction terms based on the EnKF scheme; see the mean-field version \eqref{eq:our_mf}--\eqref{eq:FNO} and their particle counterparts \eqref{eq:ourens}, \eqref{eq:full_K-sec4}. The motivation behind MNMEF is that EnKF is not an exact scheme for nonlinear or non-Gaussian settings from a mean-field perspective, so learnable corrections may close the modeling gap. When the forward dynamics and observation models are linear,
\begin{equation}
    \Psi(v)=Av,\qquad h(v)=Hv,
\end{equation}
with additive Gaussian noises as in \eqref{eq:dynamic}--\eqref{eq:observation}, the Kalman filter is exact, and the EnKF recovers it in the mean-field/infinite-ensemble limit. In this setting, the theoretically correct choice for the correction terms is
\begin{equation}
    \hat w_\theta\equiv 0,\qquad \hat z_\theta\equiv 0,
\end{equation}
so that $K_\theta$ reduces to the classical Kalman gain \eqref{eq:EnKF_gain}. Any nonzero learned corrections cannot improve optimality and therefore risk modeling observation/forecast noise rather than signal. Because the linear--Gaussian case is already optimal, training MNMEF with freely learnable $(\hat w_\theta,\hat z_\theta)$ can overfit stochastic fluctuations: with sufficiently many epochs (and without explicit regularization/early stopping), the network may drive the corrections to fit noise realizations. Although $K_\theta^{(2)}+\Gamma$ remains invertible (by positive definiteness of $\Gamma$), such noise-fitting can manifest as unstable updates and lead to filter divergence. We provide a minimal working example in our repository.\footnote{\url{https://github.com/wispcarey/DALearning}}

To make the discussion concrete, we run a controlled linear--Gaussian experiment. The state dimension is $d_v=10$ with a matrix $A\in\mathbb{R}^{10\times 10}$ whose eigenvalues satisfy $|\lambda_i(A)|=1$ for all $i$; this yields marginally stable linear dynamics. The observation dimension is $d_y=5$ with $H\in\mathbb{R}^{5\times 10}$ that observes every other coordinate. The process noise covariance is $\Sigma=\sigma_v^2 I$ with $\sigma_v=0.01$, the observation noise covariance is $\Gamma=\sigma_y^2 I$ with $\sigma_y=1$, and the initial condition is $v_0\sim\mathcal{N}(0,I)$.

\begin{figure}[t]
    \centering
    \includegraphics[width=\linewidth]{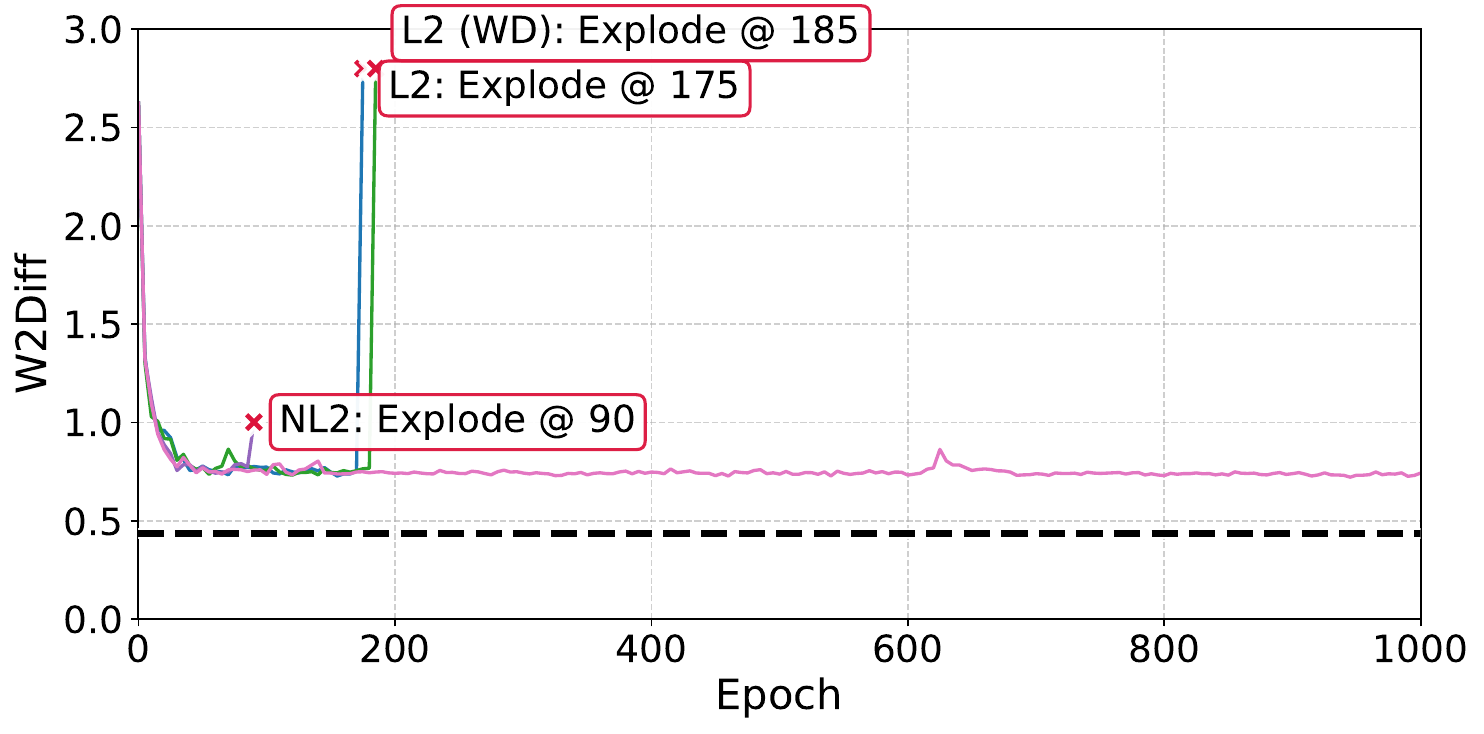}
    \vspace{2pt}
    \includegraphics[width=\linewidth]{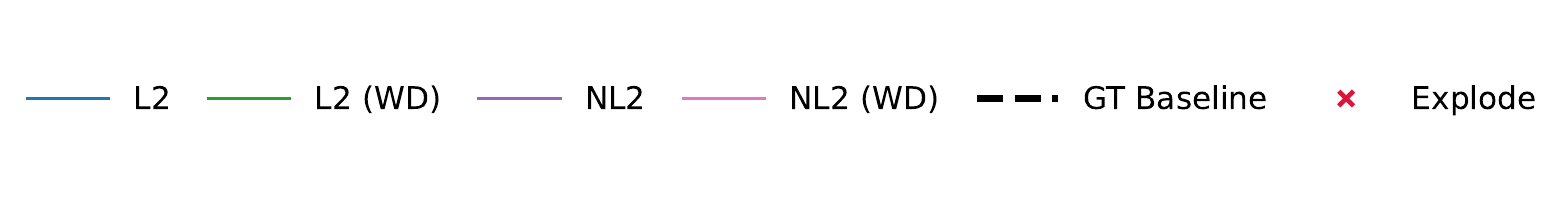}
    \caption{Linear--Gaussian experiment. Test $W_2$ (ensemble Gaussian vs.\ Kalman Gaussian) versus training epoch under the four settings in \eqref{eq:wd}, evaluated on 64 held-out trajectories (length 500 each). \emph{Legend (from the lower panel):} L2 = unnormalized loss \eqref{eq:loss_unnorm}; NL2 = normalized loss \eqref{eq:loss_app_repeat}; WD = weight decay $10^{-2}$; GT Baseline = finite-ensemble sampling baseline obtained by drawing ensembles directly from the Kalman Gaussian and computing $W_2$ via \eqref{eq:W2_gauss}. Explode marks the divergence for each curve (happening within 5 epochs of the mark). Consistent with the text, the NL2 (WD) configuration remains stable for 1000 epochs and approaches the baseline, whereas other variants are unstable.}
    \label{fig:lin_W2_vs_epoch}
\end{figure}

For training, the default loss of our MNMEF is the normalized loss used in the main text \eqref{eq:loss} and repeated here for completeness,
\begin{equation}\label{eq:loss_app_repeat}
    \mathcal{L}_m(\theta) = \frac{1}{J}\sum_{j=1}^{J}\frac{\|\bar{v}_j(\theta)- v_j^\dagger\|_2^2}{\|v_j^\dagger\|_2^2},
\end{equation}
which enforces scale invariance across trajectories. We also consider the unnormalized alternative
\begin{equation}\label{eq:loss_unnorm}
    \widetilde{\mathcal{L}}_m(\theta) = \frac{1}{J}\sum_{j=1}^{J}\bigl\|\bar{v}_j(\theta)- v_j^\dagger\bigr\|_2^2,
\end{equation}
and, orthogonally, we optionally add weight decay with coefficient $10^{-2}$, which is exactly equivalent to adding an $\ell_2$ penalty to the training objective. Combining these choices yields four settings:
\begin{equation}\label{eq:wd}
    \mathcal{L}_{\text{train}}(\theta)=
    \begin{cases}
    \mathcal{L}_m(\theta) + 10^{-2}\,\|\theta\|_2^2 & \text{(NL2 (WD): normalized loss $+$ weight decay)}\\
    \mathcal{L}_m(\theta) & \text{(NL2: normalized loss, no weight decay)}\\
    \widetilde{\mathcal{L}}_m(\theta) + 10^{-2}\,\|\theta\|_2^2 & \text{(L2 (WD): unnormalized loss $+$ weight decay)}\\
    \widetilde{\mathcal{L}}_m(\theta) & \text{(L2: unnormalized loss, no weight decay).}
    \end{cases}
\end{equation}
We train with learning rate $10^{-3}$ for $1000$ epochs on $8192$ trajectories, each of length $60$. Because the model is linear--Gaussian, the exact filtering distribution at every time step is Gaussian and can be computed by the Kalman filter, providing ground-truth means and covariances against which to evaluate.

As the metric, we compare at each step the ensemble-based Gaussian approximation (empirical mean and covariance of the ensemble) with the Kalman ground-truth Gaussian via the $2$-Wasserstein distance. For $\mathcal{N}(m_1,C_1)$ and $\mathcal{N}(m_2,C_2)$, the closed-form expression is
\begin{equation}\label{eq:W2_gauss}
    W_2^2\bigl(\mathcal{N}(m_1,C_1),\,\mathcal{N}(m_2,C_2)\bigr)
    = \|m_1-m_2\|_2^2 \;+\; \operatorname{Tr}\!\Big(C_1+C_2 - 2\bigl(C_2^{1/2} C_1 C_2^{1/2}\bigr)^{1/2}\Big).
\end{equation}
We report the test $W_2$ on 64 held-out trajectories (each of length 500 and disjoint from the training set), averaged over time and across trajectories, as a function of training epoch.

A finite-ensemble effect sets an irreducible baseline: even if an ensemble is drawn exactly from the true Kalman Gaussian at each step, the empirical mean and covariance estimated from a finite ensemble will not match the truth perfectly, hence the $W_2$ discrepancy cannot be zero. We therefore include a finite-ensemble sampling baseline by i.i.d.\ sampling from the Kalman Gaussian with the same ensemble size and computing \eqref{eq:W2_gauss}; this quantifies the best attainable error due purely to sampling.

The results are summarized in Figure~\ref{fig:lin_W2_vs_epoch}. Training with the normalized loss \eqref{eq:loss_app_repeat} without weight decay eventually exhibits exploding $W_2$, consistent with the correction pathway overfitting observation/process noise when no true signal remains to learn. Adding weight decay (normalized $+$ WD) stabilizes training throughout $1000$ epochs and closely tracks the finite-ensemble baseline. In contrast, the unnormalized loss \eqref{eq:loss_unnorm} leads to instability regardless of weight decay: both unnormalized variants eventually diverge. This confirms that, in the strictly linear--Gaussian regime where the Kalman filter is optimal and the correct corrections are $(\hat w_\theta,\hat z_\theta)\equiv (0,0)$, normalization and explicit $\ell_2$ regularization are critical to avoid learning spurious noise-driven updates.

By contrast, on nonlinear systems (Lorenz '96, Kuramoto--Sivashinsky, and Lorenz '63), where the Gaussian ansatz underlying EnKF is only approximate, the correction terms provide genuine modeling capacity. In those settings, even with very long training we did not observe noise overfitting or filter divergence; the learned corrections consistently improved or matched EnKF performance across our runs. In summary, in linear--Gaussian settings practitioners should disable the correction pathway or regularize it to recover the Kalman solution, whereas in nonlinear settings allowing trainable corrections is beneficial and did not cause instability in our experiments.

\end{document}